\documentclass[english]{article}
\usepackage{amsmath,hyperref,amssymb}
\usepackage{subfigure,graphicx,url}

\usepackage[preprint,nonatbib]{nips_2018_wider_nonotice}

\newcommand{\be}{\begin{eqnarray}}
\newcommand{\ee}{\end{eqnarray}}
\newcommand{\eeq}{\end{equation}}
\newcommand{\beq}{\begin{equation}}

\allowdisplaybreaks \numberwithin{equation}{section}
\renewcommand{\Large}{\large} 

\DeclareSymbolFont{AMSa}{U}{msa}{m}{n}
\DeclareSymbolFont{AMSb}{U}{msb}{m}{n}
\DeclareMathSymbol{\fieldR}{\mathalpha}{AMSb}{"52}

\def\beq{\begin{equation}}
\def\eeq{\end{equation}}
\def\bea{\begin{eqnarray}}
\def\eea{\end{eqnarray}}

\def\<{\langle}

\newcommand\nn{\nonumber}

\title{ A Neural Scaling Law  from  \\ the Dimension of the Data Manifold}

\author{
    Utkarsh Sharma \\
    \texttt{usharma7@jhu.edu} \\
\And 
    Jared Kaplan\\
    \texttt{jaredk@jhu.edu} \\
\AND
 {\normalfont Department of Physics and Astronomy } \\ Johns Hopkins University
}

\begin{document}
\maketitle

\begin{abstract}
When data is plentiful, the loss achieved by well-trained neural networks scales as a power-law $L \propto N^{-\alpha}$ in the number of network parameters $N$.  
This empirical scaling law holds for a wide variety of data modalities, and may persist over many orders of magnitude.  
The scaling law can be explained if neural models are effectively just performing regression on a data manifold of intrinsic dimension $d$.  This simple theory predicts that the scaling exponents $\alpha \approx 4/d$ for cross-entropy and mean-squared error losses.  We confirm the theory by independently measuring the intrinsic dimension and the scaling exponents in a teacher/student framework, where we can study a  variety of $d$ and $\alpha$ by dialing the properties of  random teacher networks.  We also test the theory with CNN image classifiers on several datasets and with GPT-type language models.

\end{abstract}

\newpage

\setcounter{tocdepth}{2}  
\tableofcontents

\section{Introduction}

\begin{figure}
\noindent \centering{} 
\includegraphics[width=0.8\textwidth]{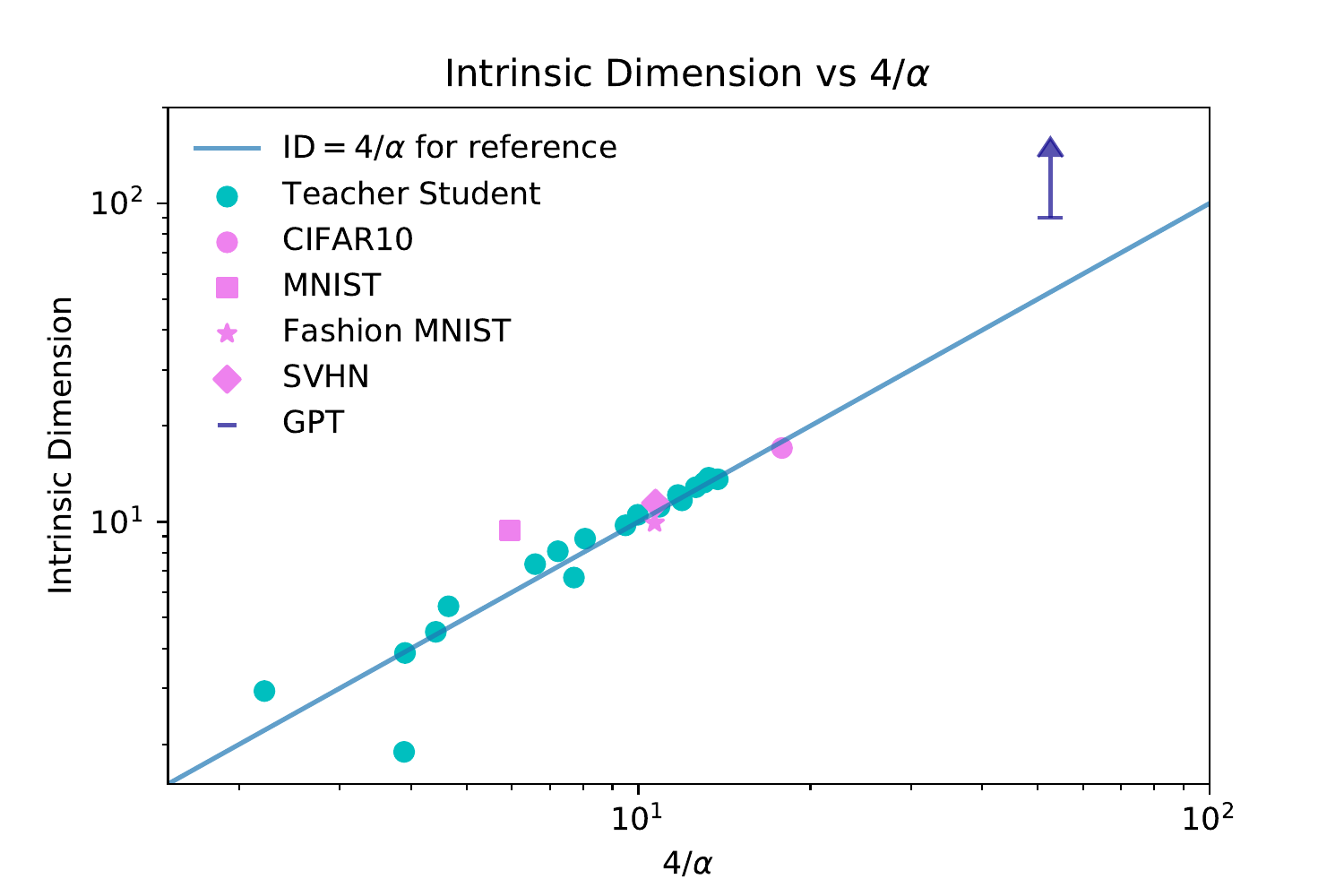}
\caption{ This figure shows the relationship between the measured intrinsic dimension (ID) of the data manifold and $\frac{4}{\alpha}$, where $\alpha$ is the model size scaling exponent.  We include data from fully-connected teacher/student experiments, simple CNNs, and GPT-type \cite{radford2018improving, radford2019language} language models  (represented as a lower-bound due to large uncertainties with large IDs).  \label{fig:AllDataDimensionvsAlpha}}
\end{figure}

Neural Network based Machine Learning has  made enormous progress in a wide variety of domains. Scale has been a key ingredient in this success: large amounts of computation, large datasets, and large models with millions or billions of parameters.

Not only is scale beneficial to performance, but the benefits from scale can be predicted precisely.  Recent works \cite{1712.00409, hestness2019beyond, rosenfeld2019constructive, kaplan2020scaling} studying a variety of data modalities and model architectures all find the same scaling relation in the underfitting regime.  In particular, the dependence of the loss on the number of model parameters $N$ has the following properties, and each suggests a corresponding question:
\begin{itemize}
\item As the number of model parameters $N$ is increased, the cross-entropy loss of well-trained and well-tuned models scales with $N$ as a power-law
\be
L(N) \propto \frac{1}{N^{\alpha}}
\ee
with observed values such as $\alpha \approx 0.076$ for language modeling \cite{kaplan2020scaling}, and much larger $\alpha \approx 0.5$ observed for image classification \cite{rosenfeld2019constructive}.  
{Why do we encounter this simple functional form, and what determines the value of the exponent $\alpha$?}
\item  Scaling holds very accurately across a wide range of $N$, sometimes spanning many orders of magnitude \cite{1712.00409, hestness2019beyond, kaplan2020scaling}. {Why does scaling persist over a large range of  model sizes, and what determines the $N_{\rm max}$ where it eventually breaks down?}
\item Empirically, the scaling exponent $\alpha$ may not depend greatly on model architecture. For example, LSTMs and Transformers scale similarly over a large range of $N$ \cite{kaplan2020scaling}, with losses differing only by an overall, $N$-independent factor.  {Why would scaling exponents be roughly independent of model architecture?}
\end{itemize}
We will argue that a simple conjectural theory can address these questions while making a number of testable predictions.  

\subsection{Main Ideas}

The key idea is that neural models map the data to a  manifold with intrinsic dimension $d$, and then use added capacity to carve up this manifold into ever smaller sub-regions.  If the underlying data varies continuously on the manifold, then the size of these sub-regions (rather than their number) determines the model's loss.  To shrink the size of the sub-regions by a factor of $2$ requires increasing the parameter count by a factor of $2^d$, and so the inverse of the scaling exponent $1/\alpha$ will be proportional to the intrinsic dimension $d$ of the data manifold.  We develop these ideas in detail in section \ref{sec:Theory}.

\begin{figure}
\noindent \centering{} 
\includegraphics[width=0.48\textwidth]{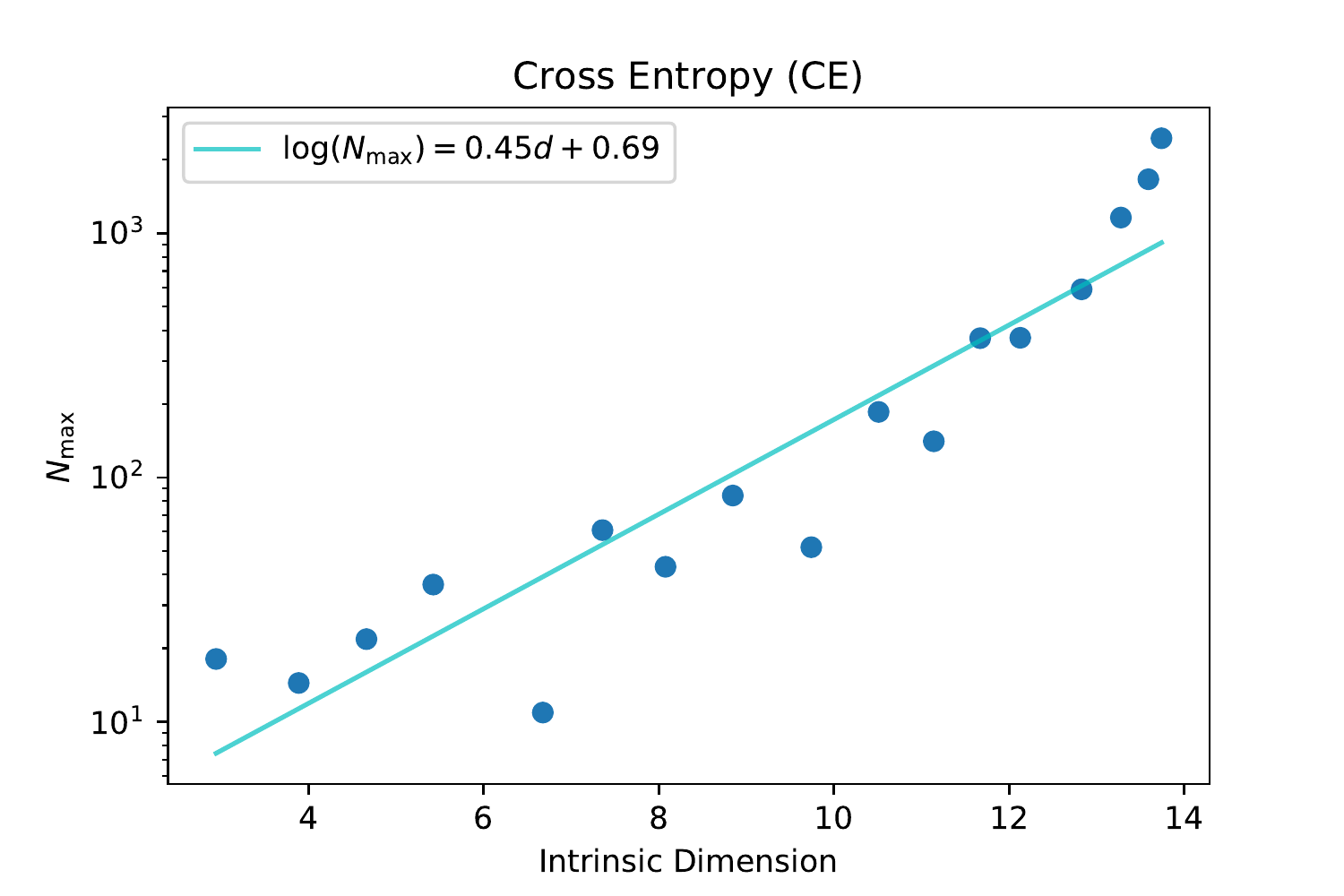}
\includegraphics[width=0.48\textwidth]{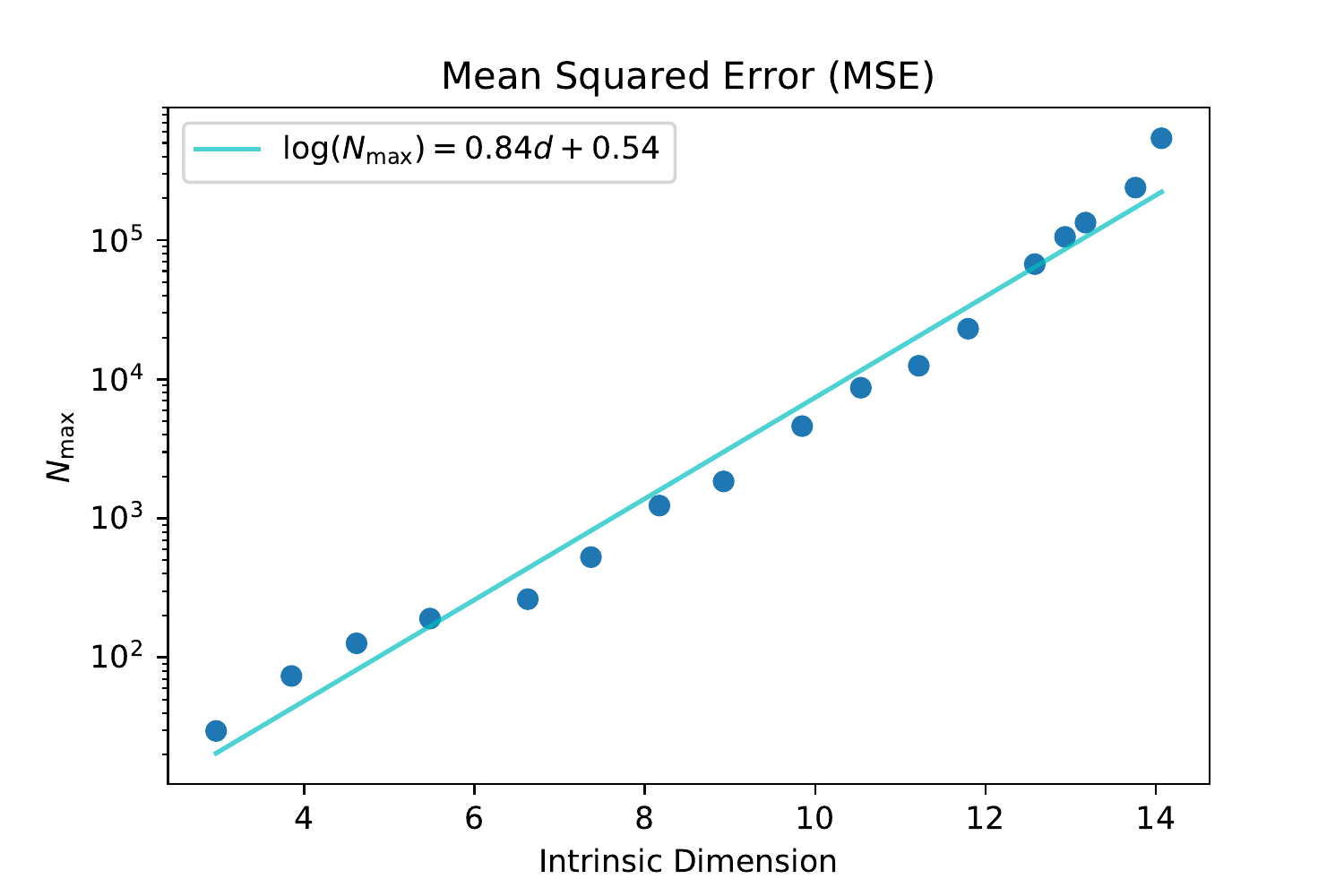}
\caption{This figure estimates the behavior of  $N_{\rm max}$, the maximum network size where we find power-law scaling, as a function of the intrinsic dimension in student/teacher experiments.  We determine $N_{\rm max}$ as the model size where the loss reaches an arbitrarily chosen small value of $0.006$, as a stand-in for the entropy of real data.  We discuss this procedure in section \ref{sec:BasicTSExperiment}.  \label{fig:PowerLawScalingRegionvsDimension}}
\end{figure}

The scaling exponent $\alpha$ can be measured by training a succession of models of varying size.  We measure the intrinsic dimension $d$ within the final layer\footnote{It was shown in  \cite{ansuini2019intrinsic} that the final hidden layer activations have the smallest intrinsic dimension in image classifiers.  Our findings are largely consistent with this.} activations of trained networks, using the distances among nearest neighbor activation vectors \cite{levina2005maximum, TwoNN}. 

We test the theory in a student/teacher framework, which makes it possible to scan over a large range of $\alpha$ and $d$ and test more idiosyncratic features of the theory (see figure \ref{fig:NetworkDiagrams}).  We also perform tests using CNNs for image classification, and by measuring the intrinsic dimension of GPT-type models \cite{radford2018improving, radford2019language}, where scaling exponent have already been documented \cite{kaplan2020scaling}.

\subsection{Contributions: Predictions and Results}

In what follows we list the concrete predictions made by our theory, and their status based on our results\footnote{Code for our experiments will be available at: \url{https://github.com/U-Sharma/NeuralScaleID}} and information in the literature.  Throughout we use $L$ to denote the loss, $N$ to denote the number of parameters in a neural network (often referred to informally as `model size'), $\alpha$ as the power-law scaling exponent, and $d$ as the intrinsic dimension of the data manifold.

\begin{figure}
\noindent \centering{} 
\includegraphics[width=0.48\textwidth]{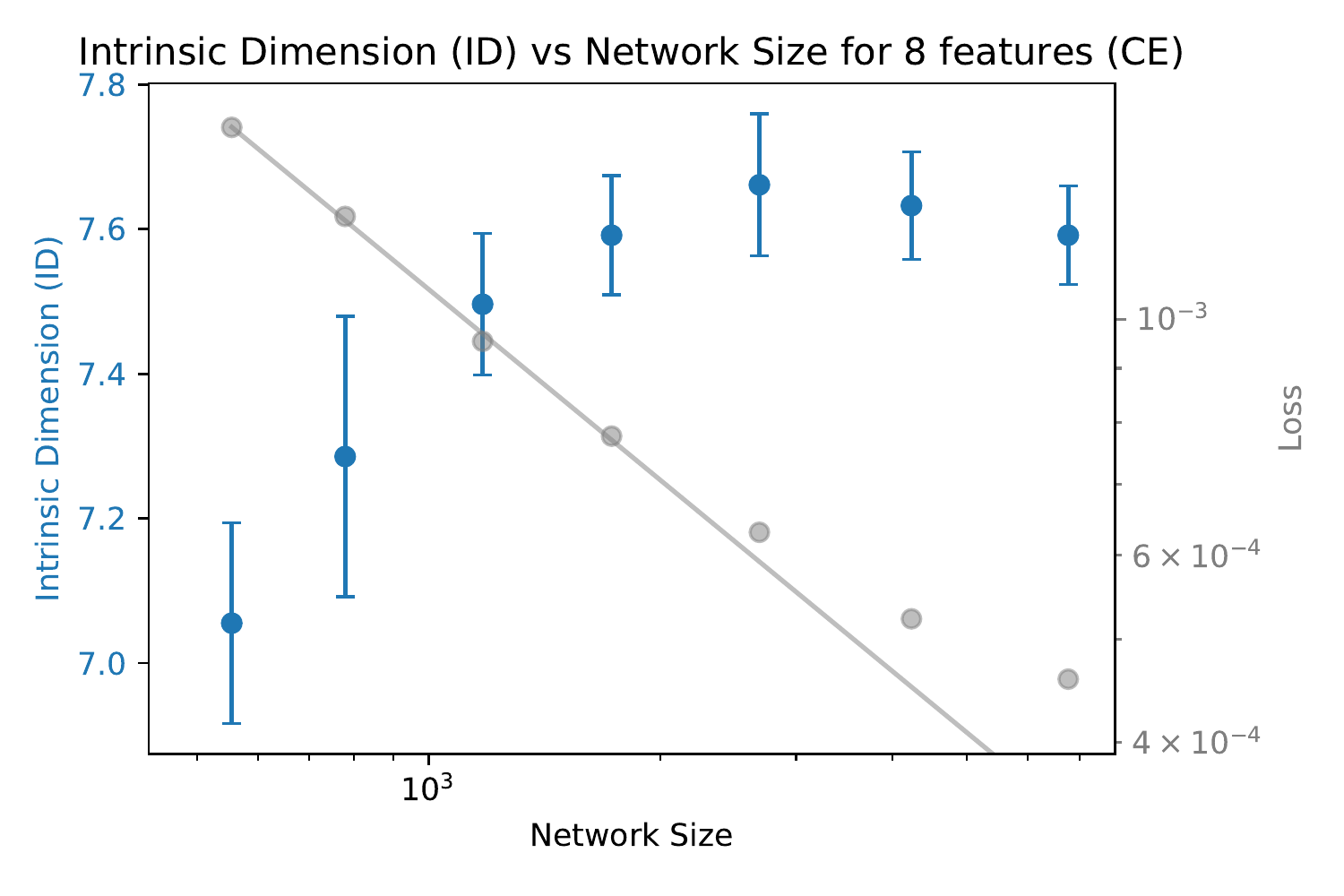}
\includegraphics[width=0.48\textwidth]{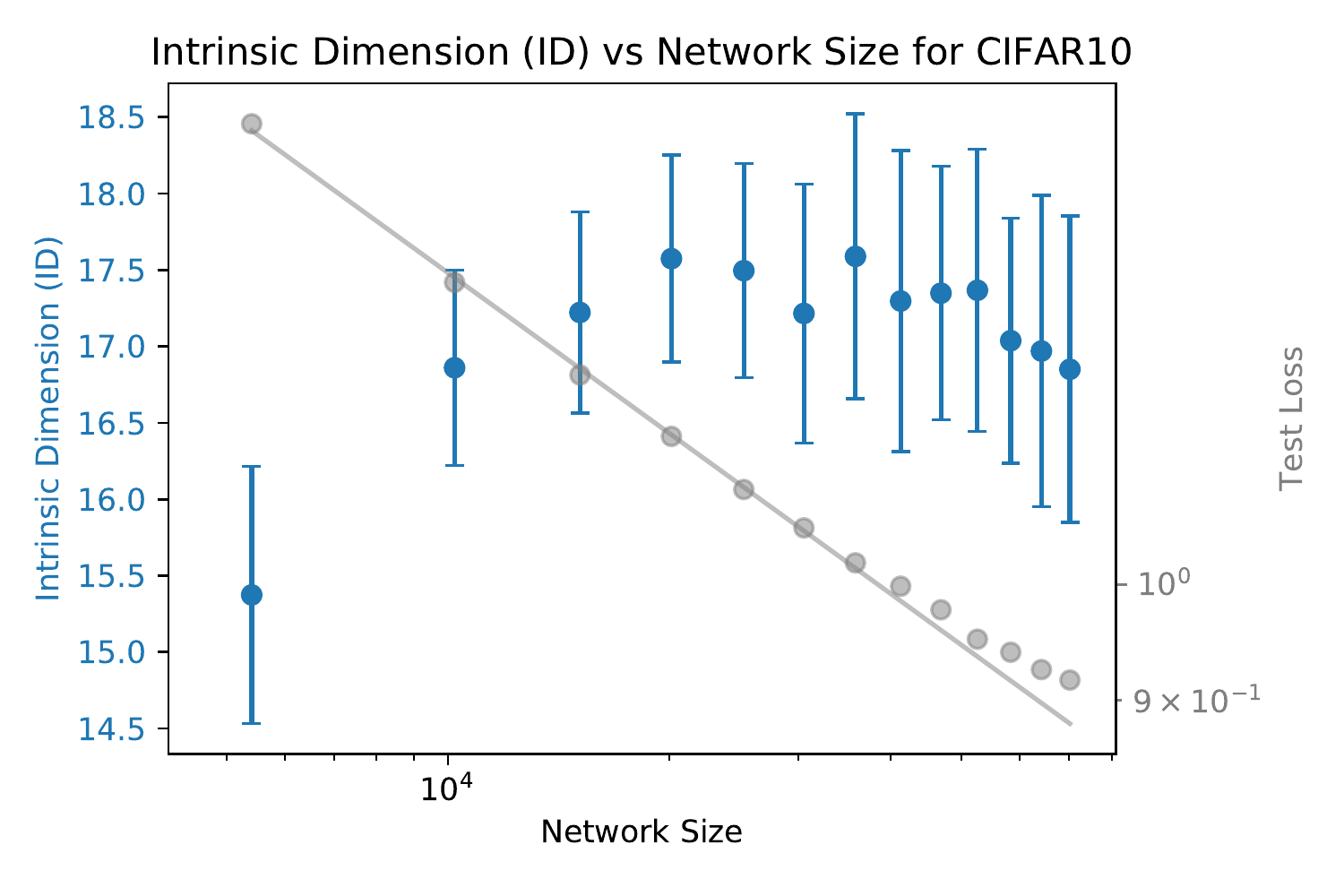}
\caption{ We show how  ID measurements vary among different student network sizes $N$ trained from the same teacher (left), and for CNNs on CIFAR10 (right).  We display the test loss $L(N)$ for reference.  The ID does not depend significantly on $N$, though it increases by about $10$\% among the various model sizes tested as $N$ increases. \label{fig:TSIDvsN}}
\end{figure}

\begin{enumerate}
\item {\bf Prediction:}  In the range of $N$ where the loss scales as $L(N) \propto \frac{1}{N^{\alpha}}$, we predict $\alpha \propto \frac{1}{d}$, where $d$ is the intrinsic dimension of the data manifold for the dataset and task in question.  If the network is composed of ReLU non-linearities and the loss is mean squared error or cross-entropy (or KL divergence), we  predict 
\be
\alpha \gtrsim \frac{4}{d}
\ee
with equality expected in the generic case.

{\bf Results:}  See figure \ref{fig:AllDataDimensionvsAlpha} for the summary combining all datasets. We find a variety of evidence  supporting this prediction, and  the factor of `4' fits quite well.  We show  in figure \ref{fig:GeneralizedLosses} that this factor can be modified if we use other loss functions. For language modeling with GPT \cite{radford2018improving,  radford2019language}, we know  $\frac{4}{\alpha} \approx 53$ while we measure the intrinsic dimension as $d \geq 90$ (figure \ref{fig:LanguageID}), in accord with the  inequality, but quite far from equality.  

\item {\bf Prediction:} The maximum network size $N_{\rm max}$ where we obtain power-law scaling grows with $d$ via $\log N_{\rm max} \propto d$.  Larger $d$ should correspond with much larger $N_{\rm max}$. 

{\bf Results:} 
We have confirmed the approximate relation $\log N_{\rm max} \propto d$ (see figure \ref{fig:PowerLawScalingRegionvsDimension}) with teacher/student experiments by identifying when $L(N_{\rm max})$ reaches a fixed  value.
\item {\bf Prediction:} The exponent $\alpha$ will not depend significantly on model architecture except through the intrinsic dimension $d$.  Since larger $\alpha$ and smaller $d$ lead to  improved performance with scale, the best architectures will tend to have the smallest  $d$. 

{\bf Results:} In \cite{ansuini2019intrinsic} it was discovered empirically that better performing image classifiers have smaller $d$, and \cite{kaplan2020scaling} showed that LSTMs and Transformers have very similar exponents.  We leave the measurement of both $\alpha$ and $d$ across distinct architectures to future work.
\item {\bf Prediction:} Models with size $N \in [N_{min}, N_{max}]$ where the loss scales as a power-law in $N$ all map the data to a manifold with the same intrinsic dimension $d$.

{\bf Results:}  We verify this for teacher/student experiments in figure \ref{fig:TSIDvsN} and for CIFAR10 in figure \ref{fig:CIFAR}.  This prediction holds to about 10\%  for these models.
\item {\bf Prediction:} If the data manifold $M = X_1 \times X_2 \cdots \times X_n$ and the loss $L(x) = \sum_i L_i(x_i)$, then we should replace the dimension of $M$ with the maximum dimension of $X_i$ when estimating $\alpha$, as the network can behave as an ensemble, modeling each $X_i$ independently (see the right of figure \ref{fig:NetworkDiagrams}).

{\bf Results:}   We confirm this prediction in section \ref{sec:ProductManifoldTests}, see figure \ref{fig:ProductManifold}.  
\end{enumerate}

\section{A Simple Theory for Scaling in the Underfitting Regime}
\label{sec:Theory}

In this section we explain our theory, beginning with a toy model in section \ref{sec:ToyTheory}.  Then in section \ref{sec:TheoryforNN} we argue\footnote{one might say conjecture; for a more sophisticated perspective in a simpler context see \cite{bickel2007local}} that the toy model can be applied to realistic neural networks with only a few small modifications.  In section \ref{sec:MeasuringID} we explain how we measure the dimension of the data manifold, a necessary step in validating the theory.  

\subsection{A Toy Model}
\label{sec:ToyTheory}

Consider one of the simplest  scenarios for multidimensional regression.   We are given a Lipschitz function $f: [0,1]^d \to \mathbb{R}$, and we would like to approximate it as  a piecewise constant function $c(x)$, by cutting $[0,1]^d$ into smaller hypercubes.  If these hypercubes have a side length $s$, then we will have
\be
N = s^{-d}
\ee
cubes, and so our approximation will depend on the $N$ constant values $c(x)$ takes within each hypercube.  If the loss is mean-squared error (MSE), then it will be bounded by 
\be
L = \int_0^1 d^d x | f(x) - c(x)|^2 \lesssim  \lambda^2 \left( s^2 d  \right)
\ee
where $\lambda$ is the Lipschitz bound $|f(x +y) - f(x)| < \lambda |y|$, and we have ignored overall numerical factors.  Translating the $s$-dependence into $N$, this means that
$L(N) \lesssim  \frac{1}{N^{2/d}}$ up to a constant factor. 

If the model is piecewise linear instead of piecewise constant and $f(x)$ is smooth with bounded derivatives, then the deviation $|f(x) - c(x)| \propto s^2$, and so the $L^2$ loss will scale\footnote{A straightforward generalization suggests that if $c(x)$ is composed of piece-wise $k$-degree polynomials, and we use a loss $|f - c|^p$, then
\be
L(s) \propto s^{(k+1) p}
\ee
in the infinite data limit.  But if $p$ is large then $c(x)$ within each hypercube will utilize many parameters.  We test the $p$-dependence of this prediction in figure \ref{fig:GeneralizedLosses}.
} as $s^4$.  We would predict 
\be
L(N) \propto \frac{1}{N^{4/d} }
\ee  
This will be important later, since networks with ReLU activations produce piecewise linear functions.

Finally, consider the case where $f_i(x)$ encode a smooth probability distribution over $i =1, \cdots, k$ possibilities, and we replace the MSE loss with the KL 
divergence.  If the $c_i(x)$ are a piecewise linear model for the logits, then 
we also find that $L \propto s^4$. So the KL and MSE losses will scale with the same exponent in $N$ at a given value of $d$.  We demonstrate this in appendix \ref{app:ScalingKL}; it is a simple consequence of the fact that the expansion of $D_{KL}(p || q)$ in $(q-p)$ begins at second order.  Note that if we use a cross-entropy instead of the KL divergence, the loss will scale in the same way towards a fixed constant value, the entropy of the true distribution.

\begin{figure}
\noindent \centering{} 
\includegraphics[width=0.35\textwidth]{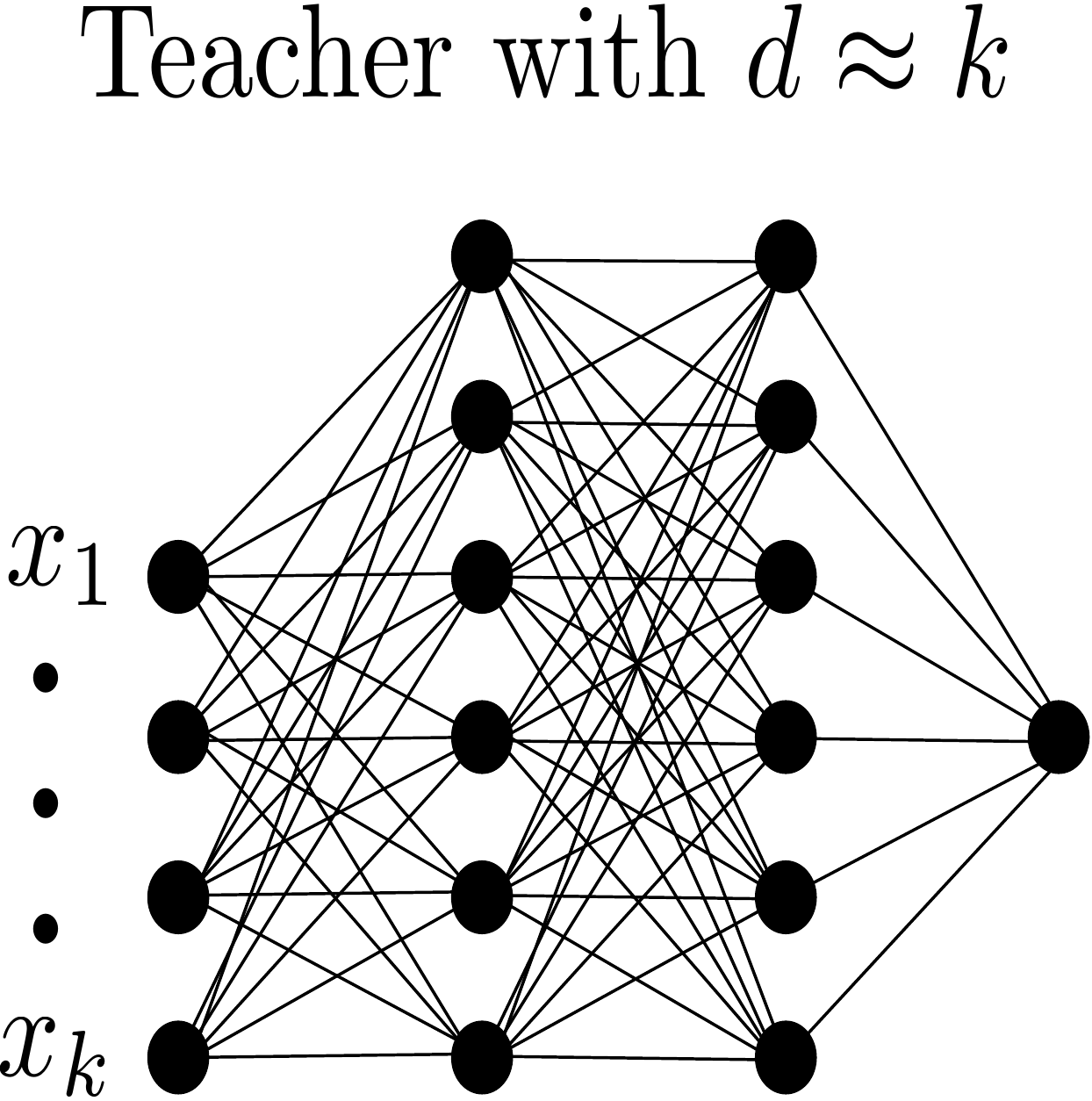}
\hspace{1.5cm}
\includegraphics[width=0.35\textwidth]{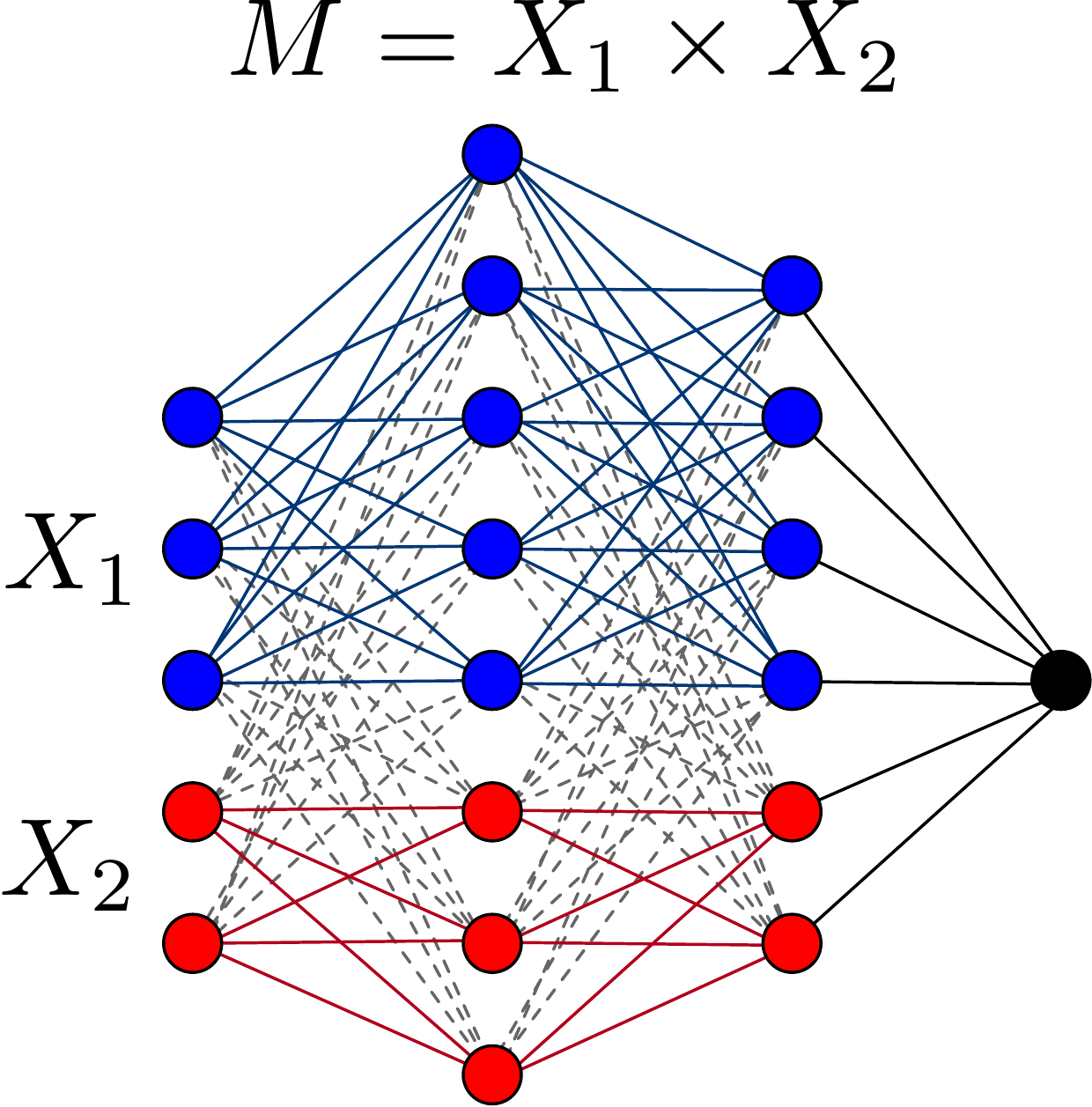}
\caption{ {\bf Left}: This shows the setup of a teacher network, emphasizing how we can control the data manifold dimension via the number of input features $k$.  {\bf Right}:  When the data manifold is a product and the teacher $T(X) = T_1(X_1) + T_2(X_2)$, then student networks can learn  $T$ by combining sub-networks and behaving, in effect, like an ensemble.  Then we predict $4/\alpha \approx d_{\rm max}$,  the maximum $d$ among the components. \label{fig:NetworkDiagrams}}
\end{figure}

\subsection{A Conjectural Theory for Neural Networks }
\label{sec:TheoryforNN}

 Neural Networks perform well on data with thousands or even millions of dimensions.  It is widely believed that this is possible because neural networks map the data into a much lower-dimensional `data manifold', preserving and focusing on the features that are relevant for the task.  

We emphasize that the data manifold is a feature of both the dataset and the task or loss function that has been optimized.  Classifiers need only attend to features relevant for classification.  Similarly, in the case of autoregressive models the data manifold would consist only of the features necessary to predict the next token in a sequence.  So the data manifold for such a model (as we are defining it) may have many fewer dimensions than the space of full sequences, such as complete images or text samples.  Properties of the data manifold may also depend on the model that is learning it, such as its architecture and activation functions.

We can explain the observed scaling relations for NNs by applying our toy theory while replacing the ambient dimension of the dataset with the intrinsic dimension of the data manifold.  If we perform regression with a neural network  with ReLU activations and a mean-squared error or KL divergence loss, the analysis of section \ref{sec:ToyTheory} implies\footnote{Depending on the network architecture and parameter values, the network could represent a piecewise linear function with $C \gg N$ piecewise components \cite{montufar2014number}.  However, these $C$ components cannot be independently configured to optimize the loss.  Since there are only $N$ independent degrees of freedom available, we expect $N$, rather than $C$, to determine the effective capacity.}
\be
\label{eq:ScalingandDimensionMain}
L(N) \propto \frac{1}{N^{\alpha} } \ \ \  \mathrm{with} \ \ \ \alpha \approx \frac{4}{d}
\ee
In the case where the function $f(x)$  depends in a generic way on $d$ independent variables, we will confirm this prediction empirically in section \ref{sec:BasicTSExperiment} (see figure \ref{fig:AllDataDimensionvsAlpha}).  We also explore some special data manifolds and other loss functions in section \ref{sec:AdditionalExperiments}.

This theory also largely explains why the scaling relation holds over such a large range of $N$.  To double the resolution with which the model differentiates different points on the data manifold, we need $2^d$ times more parameters.  It's reasonable to expect that model performance improves smoothly when we change the resolution by an order-one factor.  But this seemingly natural assumption implies that  if $d \gg 1$, we will see smooth scaling with $N$ over many orders of magnitude.  {We would predict that the range in $\Delta N$ over which smooth scaling holds satisfies $\log (\Delta N) \propto d$.}  This also  strongly suggests $\log N_{\rm max} \propto d$, where $N_{\rm max}$ is the largest network size exhibiting power-law scaling, as we do not expect $N_{\rm min}$, the beginning of the power-law region, to increase with $d$.  We discuss some reasons why power-law scaling may cease in section \ref{sec:WhyDoesScalingEnd}.

Finally, the theory suggests an interpretation for the fact that different NN architectures tend to have similar scaling exponents when applied to the same dataset.  It would appear that {a given dataset and task are associated with a data manifold of fixed dimension, and improvements in architecture do not greatly alter its properties.}  Network architectures that can achieve smaller $d$ on the same dataset can be scaled up to achieve  larger gains, and so we would expect smaller $d$ to correlate with better performance.

The interpretation of $4/\alpha$ as the dimension of the data manifold has a  close connection with the notion of fractal dimensions.  Typically fractal dimensions measure how the number of components needed to approximate a fractal scales as the components shrink.  But we can reinterpret this definition by asking how many components are needed to obtain a certain quality of approximation to the underlying fractal.  When we use the loss itself to measure the quality of the approximation, then $4/\alpha$ is proportional to the corresponding fractal dimension.

Before moving on, let us  discuss a few subtleties.

\subsubsection{A Bound, Not an Equality}
\label{sec:BoundNotEquality}

The classic analysis we reviewed in section \ref{sec:ToyTheory} provides an upper bound on the loss for function approximation (regression in the infinite data limit) using piecewise constant or piecewise linear approximators.  This bound becomes an estimate when the function being approximated is a generic Lipschitz function in $d$-dimensions.  However, if the function has a simple, non-generic structure then the loss may decrease much more quickly with  increasing model size.  So we should expect that
\be
\alpha \gtrsim \frac{4}{d}
\ee
In special cases where the true underlying function or distribution is non-generically simple, we may find that this inequality is far from saturation.

As a concrete example, consider a data manifold $M = X_1 \times X_2 \times \cdots \times X_n$ with loss $L(x) = \sum_i L_i(x_i)$, as suggested on the right of figure \ref{fig:NetworkDiagrams}.  In this case a fully connected neural network may learn\footnote{If the total loss does not decompose as a sum, it is less clear that the network can learn an effective decomposition, but it may still be possible.} this decomposition, computing each $L_i(X_i)$ using a separate path through the network, and only combining these paths in the last layer.  This would result in a scaling exponent determined by the maximum of the dimensions $d_i$ of the manifolds $X_i$.  We test $L(N)$ for product data manifolds in section \ref{sec:ProductManifoldTests} and verify these predictions.

We may end up finding $d > \frac{4}{\alpha}$ for other reasons.  We will attempt to measure $d$ among neural activations, but there may not be any single layer where the model compresses all of the data onto the data manifold.  For example, one might imagine a scenario where different components of the manifold are processed or compressed in different layers of the network.  And networks with non-ReLU activations (eg Transformers and ResNets) may mix and superimpose different data manifolds upon each other, obscuring the manifold structure and causing the measured dimension to exceed the true dimension.

\subsubsection{Why Does Power-Law Scaling Break Down?}
\label{sec:WhyDoesScalingEnd}

If the dataset size is finite, then power-law scaling with model size $N$ will cease when we begin to overfit the data.  Overfitting dominates performance on many real-world datasets, obscuring potentially clean scalings with $N$.  We encounter it with CIFAR10 in figure \ref{fig:CIFAR} and on other datasets in appendix \ref{app:CNNs}.
 
Even in the infinite data limit, if the data contains any entropy or noise then the power-law scaling must eventually end with the loss reaching a final plateau.  Scaling could also end for other, more interesting reasons.  For example, perhaps beyond a certain point the loss can only improve by exploring a higher dimensional data manifold.  This is possible if the data manifold has a pancake-like structure, with a small width that can only be dissected by models with very large capacity.  We will explore the simplest possibility, where the data has entropy, with mock teacher/student experiments; see figure \ref{fig:PowerLawScalingRegionvsDimension} for the  result.

\begin{figure}
\noindent \centering{} 
\includegraphics[width=0.95\textwidth]{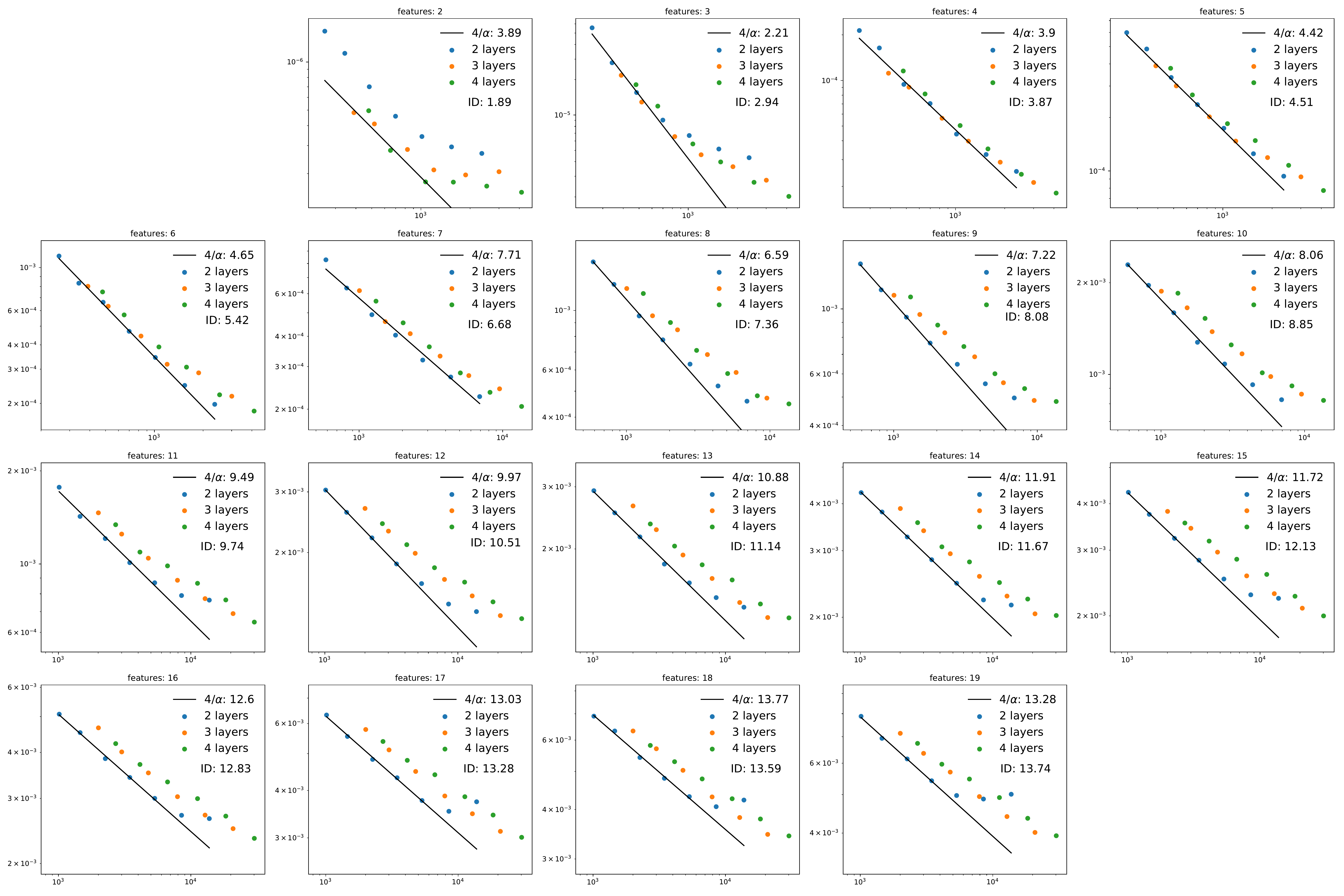}
\caption{This figure shows $L(N)$ along with power-law fits for teacher/student experiments.  The  students learn from a randomly initialized 2-layer teacher with $2$-$19$ features and use a cross-entropy loss.  The students have 2,3, or 4 layers, but for $k>5$ input features the 2-layer students perform best and determine the model-size scaling.   The measured $4/\alpha$  increases linearly with the number of features, as shown in figure \ref{fig:TeacherStudentAlphavsDim}.  \label{fig:AllTSNetworkLvsN}}
\end{figure}

\subsection{Measuring the Intrinsic Dimension of the Data Manifold}
\label{sec:MeasuringID}

In section \ref{sec:TheoryforNN} we extended the toy model in order to make a variety of predictions relating the scaling of the loss with model size to $d$, the intrinsic dimension (ID) of the data manifold.    In some of our experiments, we will control $d$ by constructing generic functions of $d$ inputs and then measuring $\alpha$.  But the theory would be tautological for real-world data if we could not independently measure the data manifold's ID. 

We will define $d$ by measuring the ID of  neural activations as the network processes data from the distribution on which it was trained.  There is an extensive literature on intrinsic dimension estimation (for a review see \cite{camastra2016intrinsic}).  In most cases we  use the  simple  two-nearest neighbors (TwoNN) method \cite{TwoNN}, though we also compare to the MLE estimation \cite{levina2005maximum} method on which TwoNN was based. 

To summarize the method, let $r_k$ be the distance from a given datapoint to its $k$th nearest neighbor, and define $\mu_k \equiv r_k / r_1$.  Then the cumulative distribution $C(\mu_k)$ takes the form
\be
C(\mu_k) = \left(1 - \frac{1}{\mu_k^{d}} \right)^{k-1}
\ee
and so we can measure the intrinsic dimension $d$ by using the relation
\be
d = \frac{\log \left( 1 - C(\mu_k)^{\frac{1}{k-1}} \right) }{\log \mu_k}
\ee
Practically speaking, we evaluate $\mu_k$ for every point on the manifold, and then apply linear regression to measure the slope $d$.
We  measure $d$ using various $k$ and verify that different values of $k$ give consistent results.  We also verify that the MLE method \cite{levina2005maximum} agrees with the TwoNN method.  Fortunately, nearest neighbors can be efficiently identified  \cite{sklearn_api}.

The TwoNN method (the case $k=2$) has already been applied to neural networks \cite{ansuini2019intrinsic}.  There it was found that the dimension is smallest when measured using the activations of the final hidden layer of the network (immediately before the logits or output, so sometimes we refer to this as `prefinal').  We will use these activations to measure $d$ and compare to $1/\alpha$.  For the GPT-type models (and for some others as a test in appendix \ref{app:TestingID}) we show ID measurements for every layer.

 For convenience we provide a self-contained derivation of these ID measurement algorithms and  a minor extension ($k > 2)$ in appendix \ref{app:IDfromNeighbors}.  We also provide several tests of the method in appendix \ref{app:TestingID}, using both synthetic and neural activation data.  We find that the method is fairly accurate for $d \lesssim 20$, while for larger dimensions it's less reliable, and typically (but not always) underestimates the true dimension.  Statistical errors from these methods are often fairly small (particularly from TwoNN), but we expect there may be larger systematic errors, as discussed in the appendices.

\section{Experiments and Results}
\label{sec:Experiments}

\begin{figure}
\noindent \centering{} 
\includegraphics[width=0.48\textwidth]{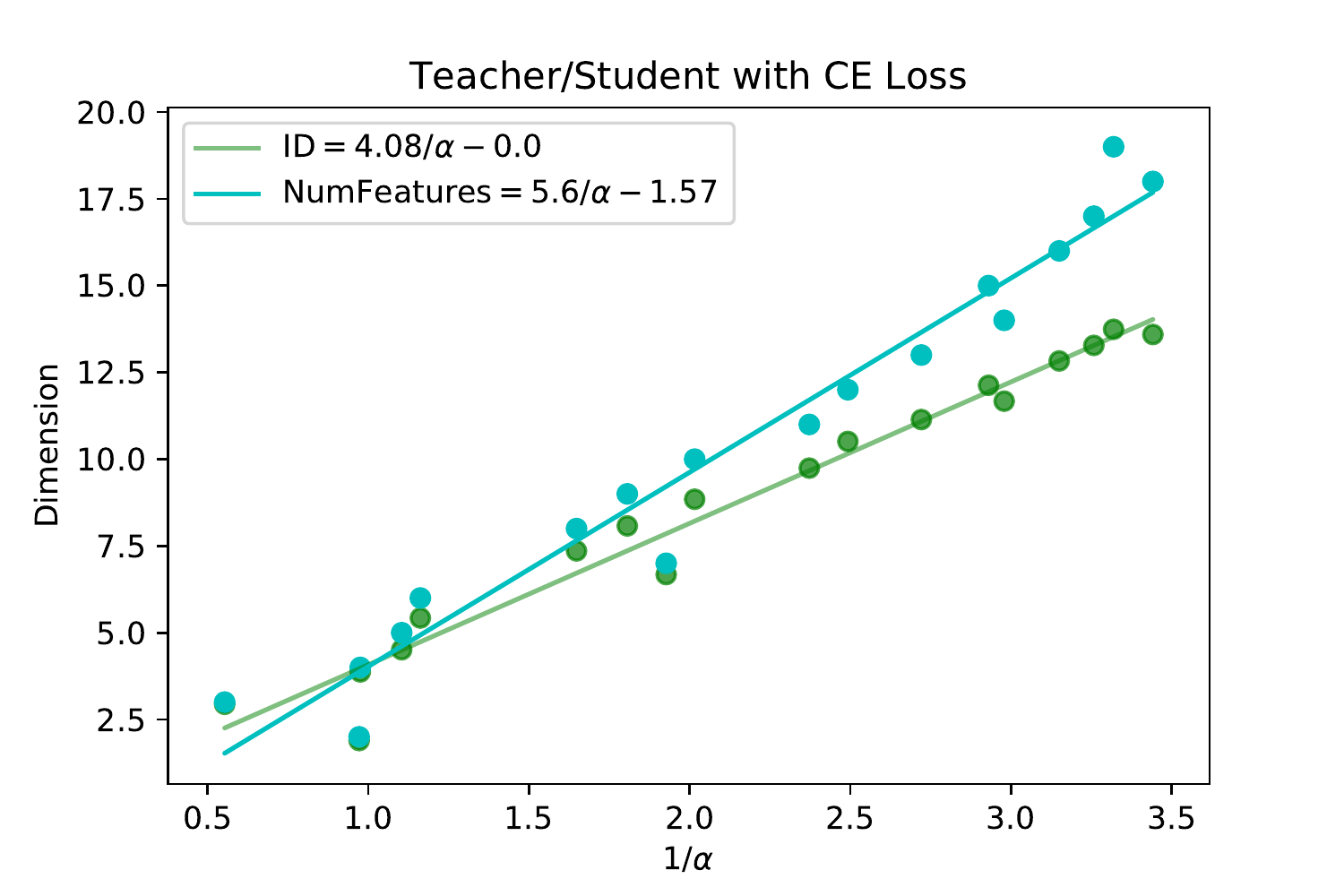}
\includegraphics[width=0.48\textwidth]{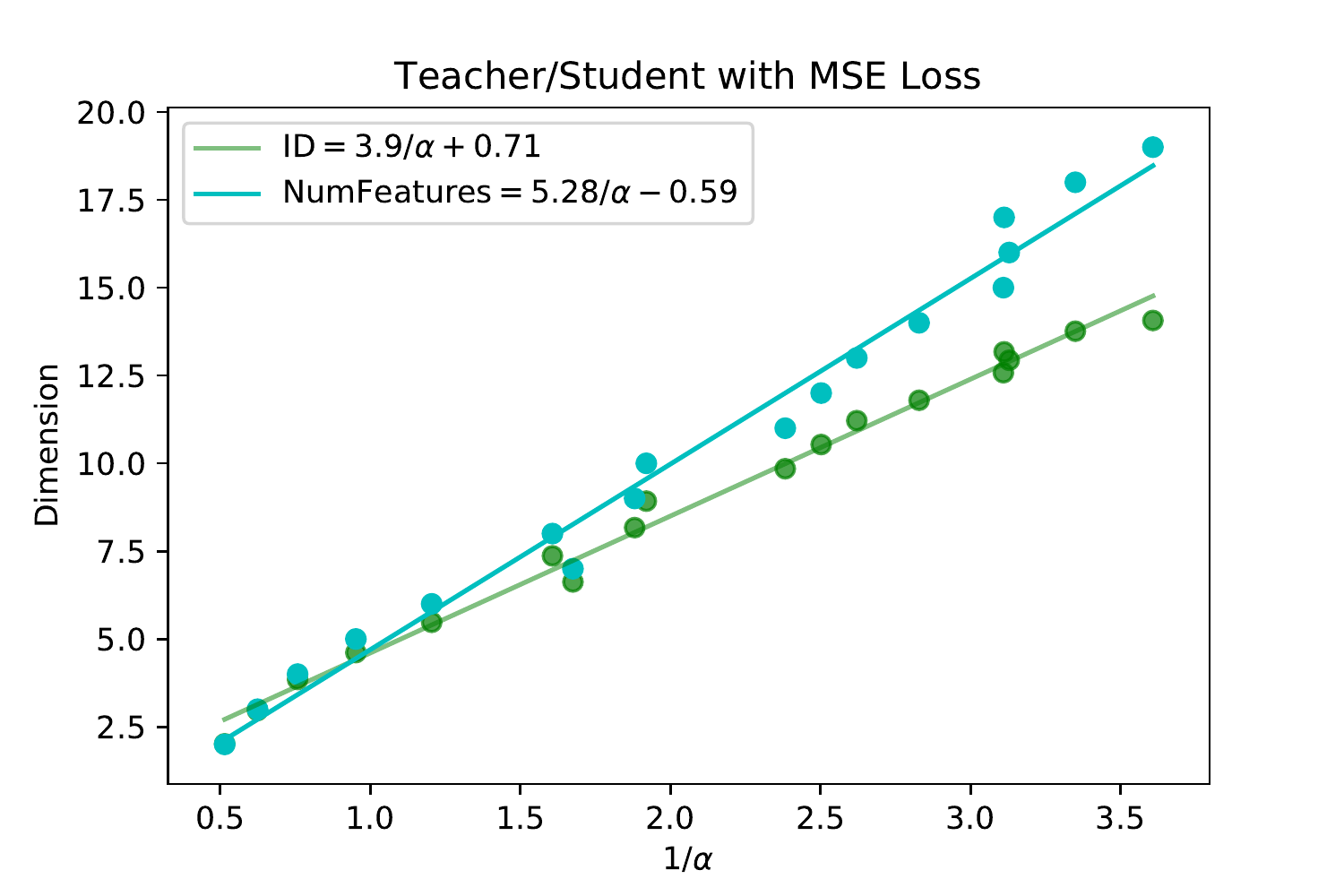}
\caption{These figures show the correlation between the inverse scaling exponent $4/\alpha$ and both the measured intrinsic dimension and the number of input features (dimensions) in the teacher network.  Both notions of dimension are linearly correlated with $1/\alpha$, and the intrinsic dimension scales almost exactly as $4/\alpha$, as predicted in section \ref{sec:TheoryforNN}.  \label{fig:TeacherStudentAlphavsDim}}
\end{figure}

In this section we discuss results from teacher/student experiments and various extensions, and also some tests using image classification and language modeling.  We relegate a variety of technical details and a few minor observations to appendix \ref{app:TechnicalDetailsMinorResults}.  We discuss potential errors in the ID measurement, along with several examples, in appendix \ref{app:TestingID}.

\subsection{Teacher/Student with Random Teachers}
\label{sec:BasicTSExperiment}

We generate functions of $k = 2,3, \cdots, 19$ input features using a randomly initialized, fully connected `teacher' neural network with a 20-dimensional input space.  To achieve $k < 20$ we simply zero out all other inputs to this single teacher.  We refer to $k$ as the number of features, and distinguish it from $d$, the intrinsic dimension, which we measure using the activations of trained student networks.  

For each value of $k$, we train fully connected student networks of various widths and depths to imitate the outputs of the teacher.  We work in the online setting, generating random inputs in $[-\frac{1}{2}, \frac{1}{2}]^k$ so  the dataset size is effectively infinite.  Details of the network topologies, training procedure, fits, errors, and ID measurements are documented in appendix \ref{app:TSDetails}.

After training the students, we evaluate the loss $L_k(N)$ for each number of features $k$.  Then we fit
\be
L_k(N) = \frac{c}{N^{\alpha} }
\ee
to measure $c, \alpha$ for each $k$.  The results of this process (with cross-entropy  loss) are shown in figure \ref{fig:AllTSNetworkLvsN}.  

Next we  measure the intrinsic dimension  from the activations of the final hidden layer of each trained student.  We use $12,000$ activation vectors for each ID measurement.  In all cases we find that using more nearest neighbors, as discussed in section \ref{sec:MeasuringID}, does not change the result significantly.  
In figure \ref{fig:TSIDvsN} we show the measured ID of the final layer of a student network with various sizes $N$, along with a plot of the loss $L(N)$.  We see that the ID is approximately constant for these networks, though it does slowly grow by about $10$\% from the smallest to the largest student network.

We plot the relationship between $4/\alpha$ and either the number of features or the measured ID $d$.  The result, along with linear fits, are shown in figure \ref{fig:TeacherStudentAlphavsDim}.  For both the cross-entropy and MSE loss functions, $\frac{4}{\alpha} \approx d$.  The inverse exponent $1/\alpha$ is linearly related to the number of input features $k$, but the multiplier is larger than $4$.

\begin{figure}
\noindent \centering{} 
\includegraphics[width=0.32\textwidth]{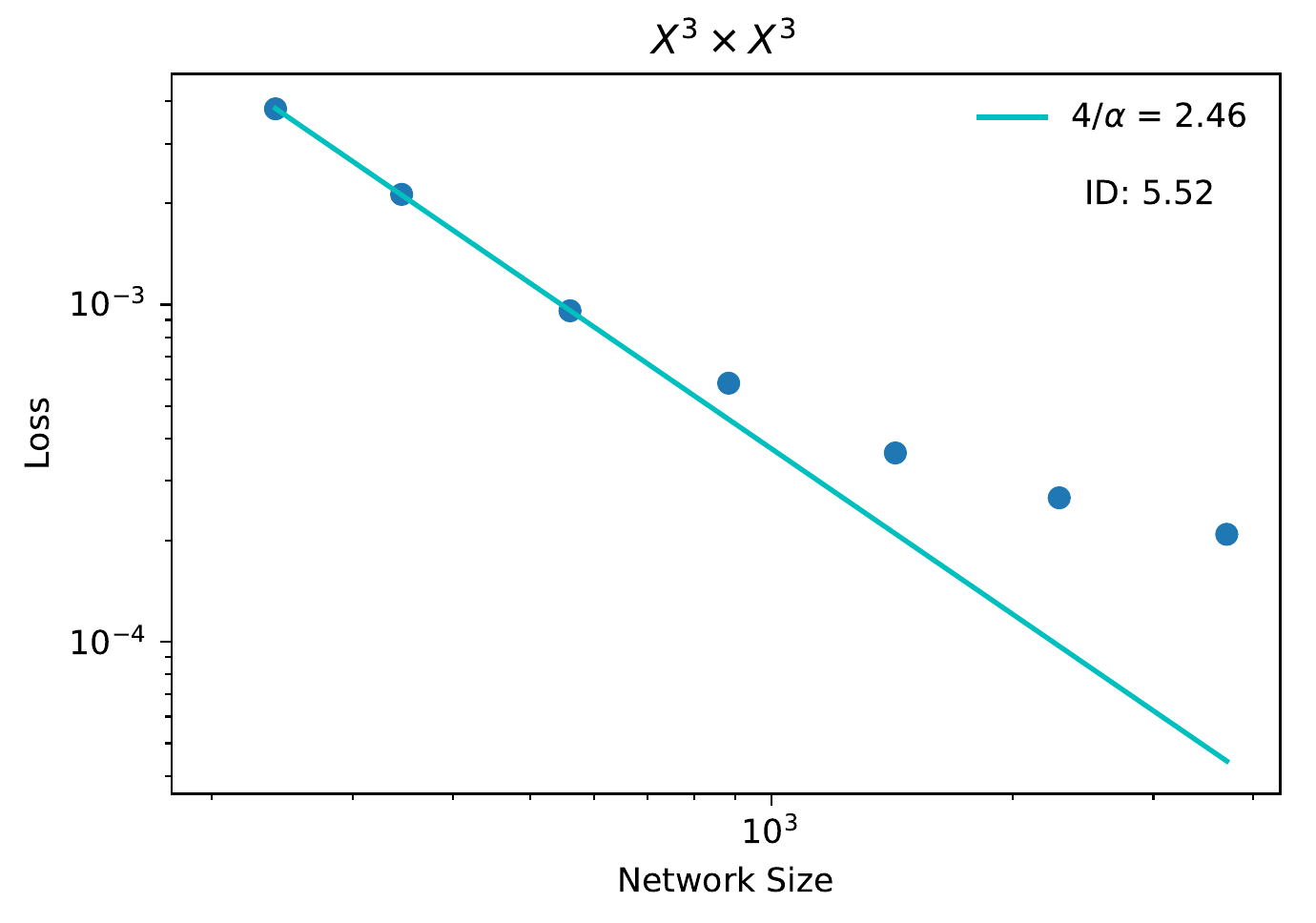}
\includegraphics[width=0.32\textwidth]{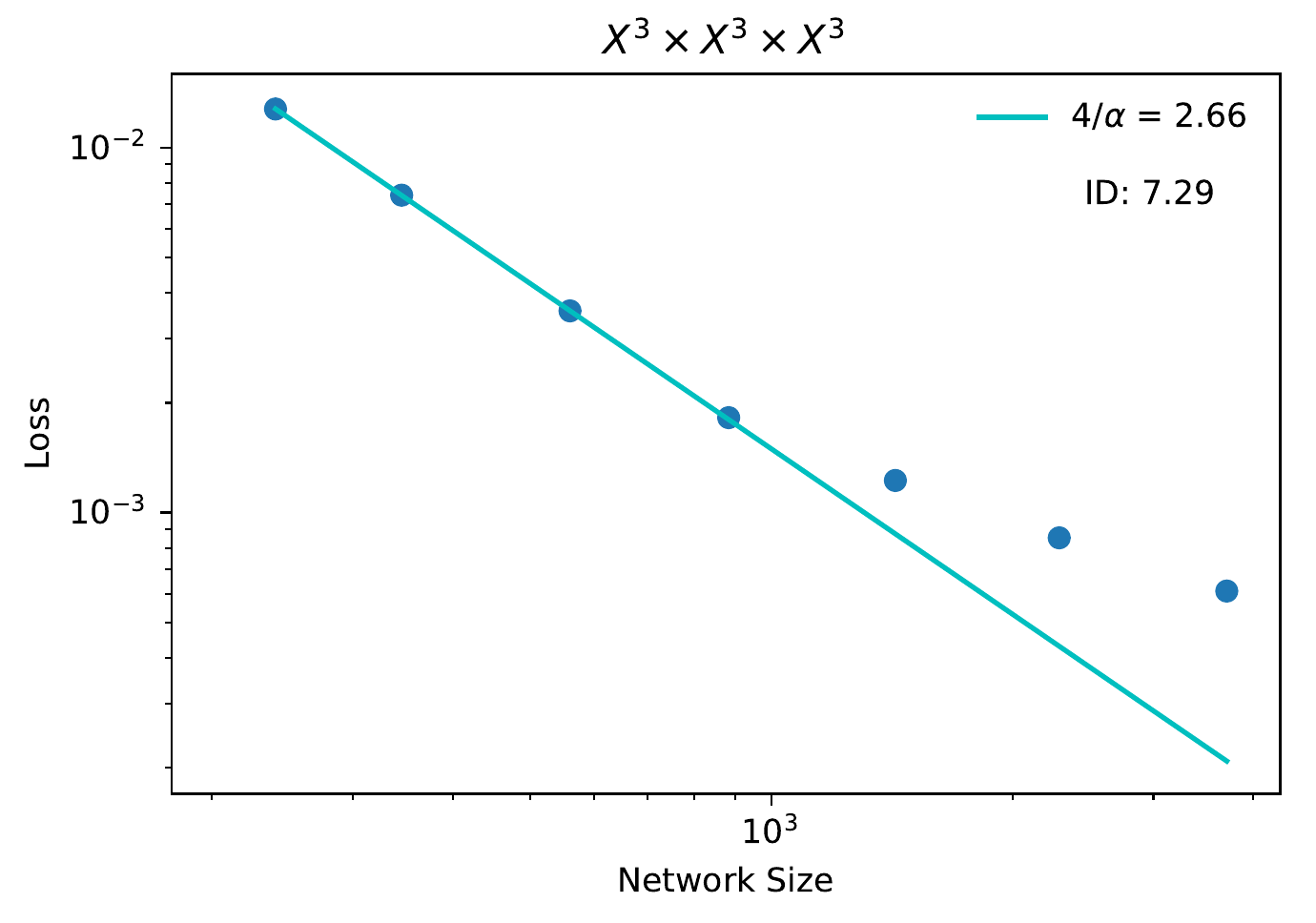}
\includegraphics[width=0.335\textwidth]{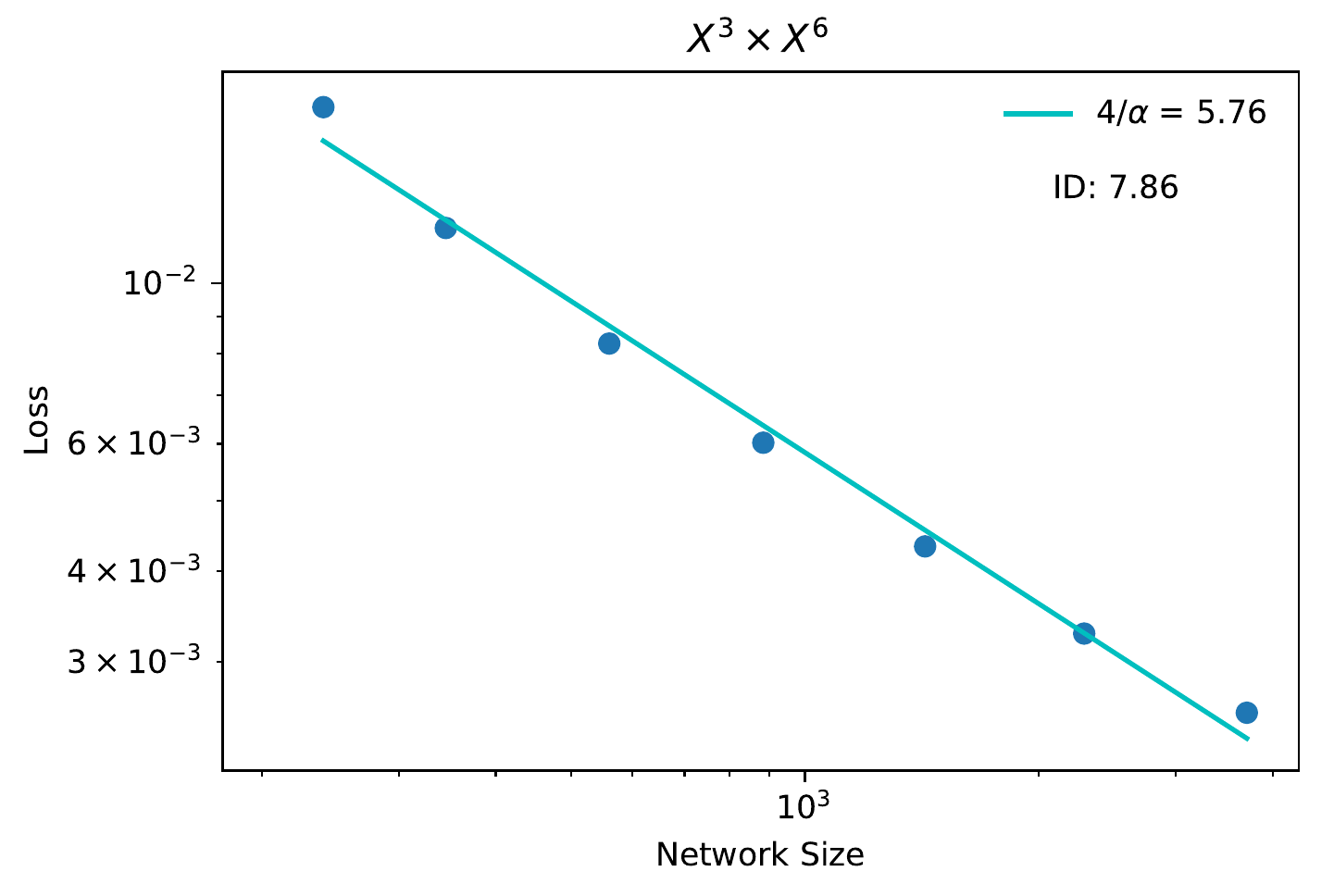}
\caption{This figure shows results for $\alpha$ and $d$ for product data manifolds with teachers $T_{3+3}$ (left),  $T_{3+3+3}$ (middle), and $T_{3+6}$ (right).  We see that in all cases $\frac{4}{\alpha} \approx {\rm max}(d_i)$ among the product factor manifolds.  The total measured IDs are approximately equal to the sum of the dimensions of the product factors, as expected.   \label{fig:ProductManifold}}
\end{figure}

In section \ref{sec:WhyDoesScalingEnd} we argued that scaling should end at an $N_{\rm max}$ that grows as $\log N_{\rm max} \propto d$.  We would like to  test this prediction with teacher/student experiments, but in this case the data has no entropy.  So instead we will introduce an artificial threshold for the loss, as a fictitious stand-in for the entropy of real data.  Then we simply ask at what $N_{\rm max}$ the loss $L(N)$ reaches this fixed, arbitrary value.  

We chose $L = 6 \times 10^{-3}$ as an arbitrary threshold in figure \ref{fig:PowerLawScalingRegionvsDimension}. Note that for the teacher networks with fewer features we used the power-law fit for $L(N)$ to estimate $N_{\rm max}$, as it was smaller than any network tested.  This means we had to extrapolate $L(N)$, so these results are not purely empirical.  We also compare $\log N_{\rm max}$ and $d$ by defining $N_{\rm max}$ as the end of the purely empirical power-law scaling region for 2-layer students (due to a failure of optimization or numerical precision issues); these results are relegated to figure \ref{fig:PowerLawScalingRegionvsDimensionEndofRegion} in the appendix. 

The ID is typically a bit smaller than the number of input features.  This may arise from a combination of two factors:  the ID measurement may be underestimating the data manifold dimension, and randomly initialized networks may  not provide sufficiently generic or non-linear functions of their inputs.  We explore the second hypothesis  in appendix \ref{sec:VettingTeachers}, where we show that by vetting the teacher networks we can improve agreement between ID and the number of input features.  Figure \ref{fig:SyntheticData} provides some idea of the potential errors in the ID measurements. Since the inputs themselves are drawn from a uniform distribution it is plausible that the ID is somewhat of an underestimate due to boundary effects.

\subsection{Product Data Manifolds and Other Loss Functions}
\label{sec:AdditionalExperiments}

\subsubsection{Product Data Manifolds $M = X_1 \times \cdots \times X_n$}
\label{sec:ProductManifoldTests}

If the data manifold takes the form $M = X_1 \times X_2 \times \cdots \times X_n$, with the underlying function of $x \in M$ decomposing as $F(x) = \sum_i f_i(x_i)$, then we expect that a neural network should be capable of separately modeling each $f_i$  within separate blocks of activations, and then combining them in the final layer to compute the full $F$.  This means that although the ID of $M$ will be measured as $d_M = \sum_i d_{X_i}$, we should expect
\be
\alpha &=& \frac{4}{{\rm max}(d_{X_i})}
\ee
as we discussed briefly in section \ref{sec:BoundNotEquality}, and demonstrate diagrammatically on the right of figure \ref{fig:NetworkDiagrams}.

\begin{figure}
\noindent \centering{} 
\includegraphics[width=0.5\textwidth]{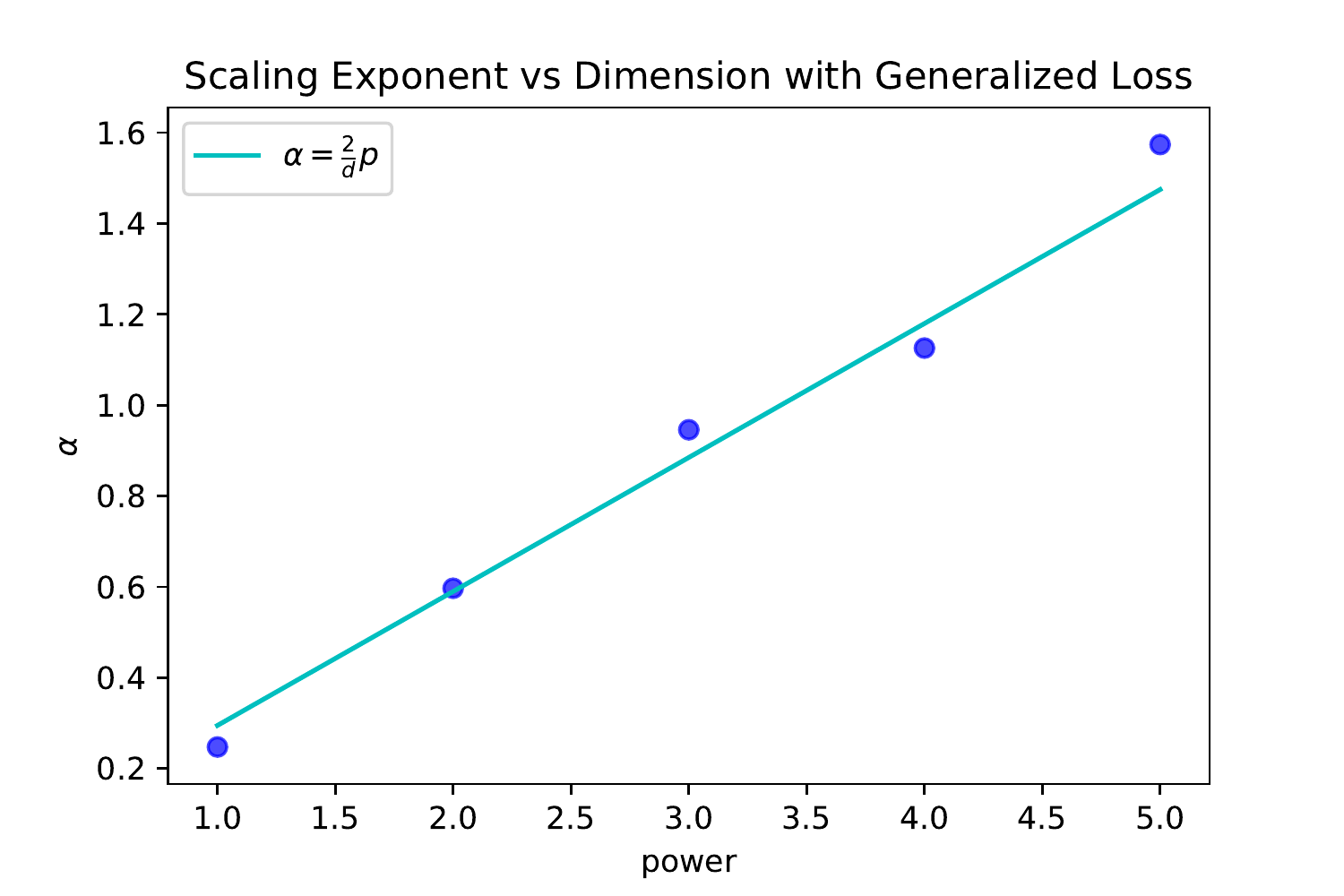}
\caption{This figure shows the relationship between $\alpha$ and the power $p$ when we use the generalized loss $|y - y^* |^p$.  As expected from section \ref{sec:ToyTheory}, we find $\alpha = \frac{2}{d} p$.  This is a student/teacher experiment with $d \approx 7$.   \label{fig:GeneralizedLosses}}
\end{figure}

To test this prediction we use a vetted teacher network with 3 real inputs $T_3(x_1, x_2, x_3)$ and another vetted teacher taking 6 real inputs $T_6(x_1, \cdots x_6)$.  Individually, these had ID $d_3 = 2.98$ and $d_6 = 5.31$ and their $L(N)$ exponents satisfied $\frac{4}{\alpha_3} = 3.3$ and $\frac{4}{\alpha_6} = 4.9$. These teachers each produce a pair of logits. We then constructed the new teacher functions with logits
\be
T_{3+3}(x) &=& T_3(x_1, x_2, x_3) + T_3(x_4, x_5, x_6)
\nn \\
T_{3+3+3}(x) &=& T_3(x_1, x_2, x_3) + T_3(x_4, x_5, x_6) + T_3(x_7, x_8, x_9)
\nn \\
T_{3+6}(x) &=& T_3(x_1, x_2, x_3) + T_6(x_4, x_5, \cdots , x_9)
\ee
and trained students to imitate these teachers using the cross-entropy loss.  We then measured the resulting ID and $\alpha$ for these three  product-manifold teachers.  For the $T_{3+3}$ and $T_{3+3+3}$ cases we used two or three different teachers to make sure the network could not take advantage of the exact repetition of a single teacher.

As shown in figure \ref{fig:ProductManifold}, the results confirm our predictions.  This provides a concrete example where we may find that $\alpha > \frac{4}{d}$ for reasons that the theory precisely anticipates.  More importantly, it provides a very detailed test of our theoretical picture relating scaling exponents to properties of the data manifold.

\subsubsection{Other Loss Functions}

The factor of `4' in the relation $d \approx \frac{4}{\alpha}$ is derived from the behavior of the loss function and the expectation that networks with ReLU activations form piecewise linear functions.  If we use a loss function such as $L(y, y^*) = |y - y^* |^p$ for regression, from the argument of section \ref{sec:ToyTheory} we would  expect
\be
\alpha \approx \frac{2p}{d}
\ee
where the MSE case corresponds to $p=2$.  We verify this  in figure \ref{fig:GeneralizedLosses} using a fixed teacher with intrinsic dimension $d \approx 7$, as measured in the usual student/teacher context.

\begin{figure}
\centering
\includegraphics[scale=0.5]{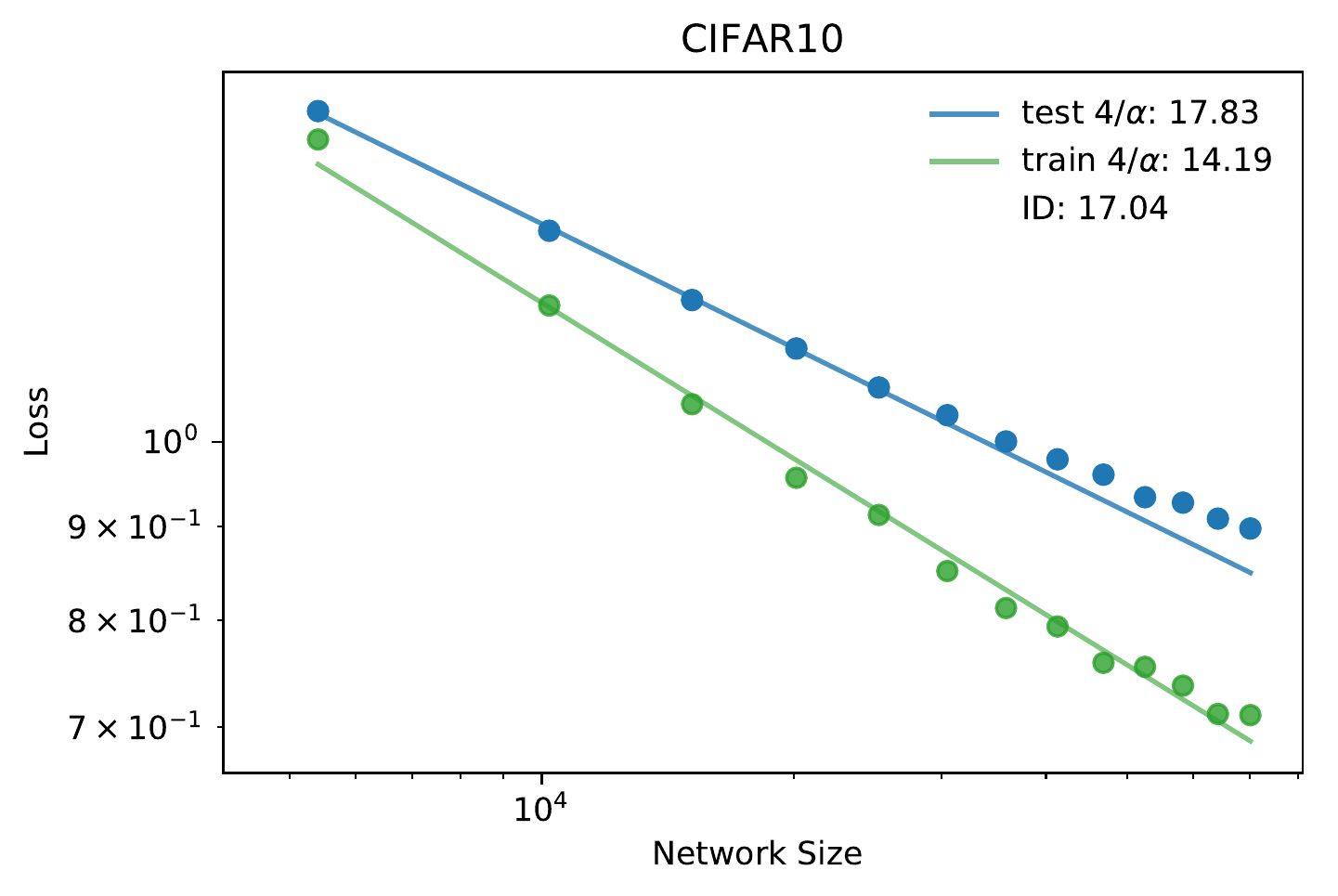} 
\includegraphics[scale=0.5]{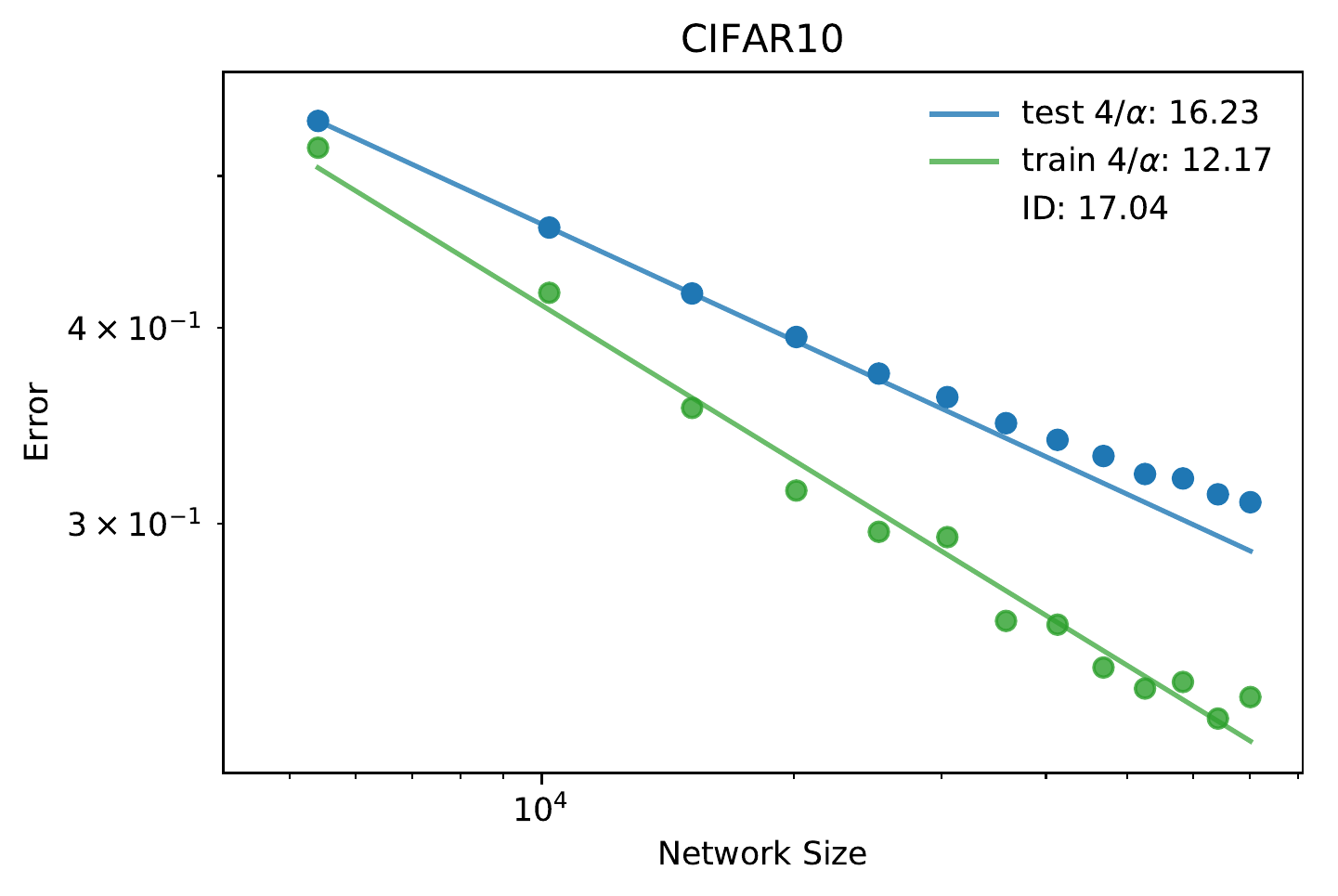} 
\caption{The left figure shows the test and training loss $L(N)$ for various sizes of CNN trained on CIFAR10, while the right figure shows error ($1 - $ accuracy).  All results are evaluated at the early stopping step, where the test loss is minimized. We report test loss results in figure \ref{fig:AllDataDimensionvsAlpha}, but note that the exponents for accuracy are very close to those for loss.  \label{fig:CIFAR}}
\end{figure}

\subsection{Image Classification with Simple CNNs}
\label{sec:CNNs}

Our goal with these experiments was to study a simple, all ReLU architecture that could scale down to a small enough size to avoid overfitting CIFAR10 \cite{Krizhevsky09learningmultiple}.
So we used a  version of the default tutorial CNN in tensorflow \cite{tensorflow2015-whitepaper}, which we modified only by scaling the number of channels (ie the width).  
Figure \ref{fig:CIFAR} shows the scaling of the  test loss with number of parameters $N$.  Our only regularization was early stopping.  The results match $4/\alpha = d$ quite well.

In an ideal test of the theory, we would measure $\alpha$ fully in the underfitting regime, with no distinction between train and test performance.  But there is a train/test gap even for the smallest network sizes, so its unclear how to model the error in the  $\alpha$ measurement.  In addition to the test loss, we also measured the scaling of the training loss for these models, recording it at the early-stopping step, and found that it also scales similarly.    Furthermore, note that on the right of figure \ref{fig:CIFAR} we record the error rate ($\equiv 1 - $ accuracy), and find that it scales very similarly to the loss.

We performed a very similar analysis on the MNIST  \cite{lecun-mnisthandwrittendigit-2010}, fashion MNIST \cite{fmnist}, and SVHN \cite{netzer2011reading} datasets using slightly smaller networks (see section \ref{app:CNNs}).  We plot $L(N)$ in figure \ref{fig:FMNIST}, which we have relegated to the appendix, as the power-law trends on these datasets are less clear than on CIFAR10.

Power-law exponents and IDs for CIFAR10 have been measured elsewhere using more powerful architectures, finding both a larger value of $\alpha \approx 0.5$ (for the error rate) \cite{rosenfeld2019constructive} and a smaller ID $\approx 8$  \cite{ansuini2019intrinsic}.  We cannot make a  clean comparison, but given that we find that the exponent for error-rate and loss scaling seem to be similar, these results  appear to match our predictions.

\subsection{Language Modeling with GPT-type Models}
\label{sec:Language}

\begin{figure}
\centering
\includegraphics[scale=0.48]{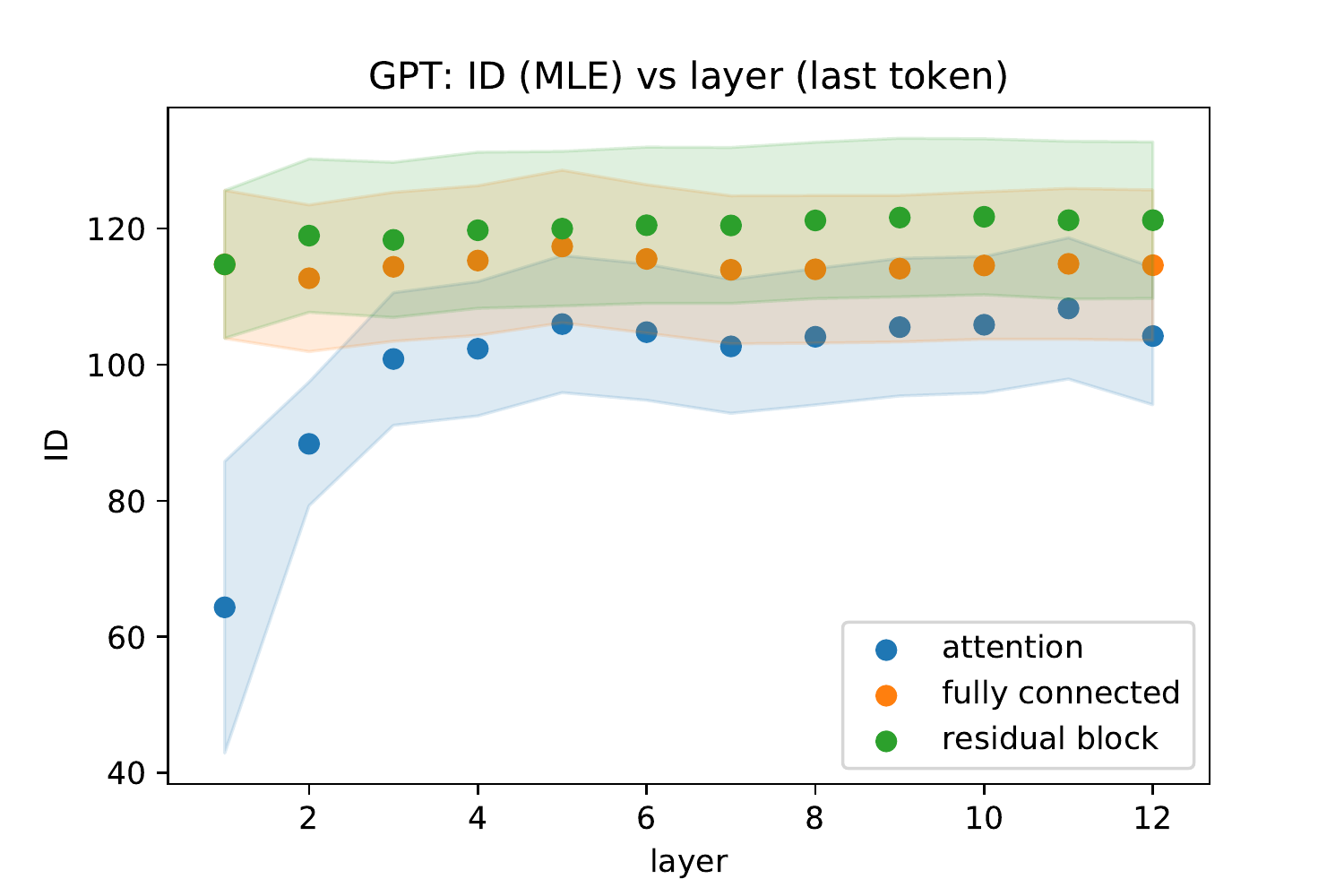} 
\includegraphics[scale=0.48]{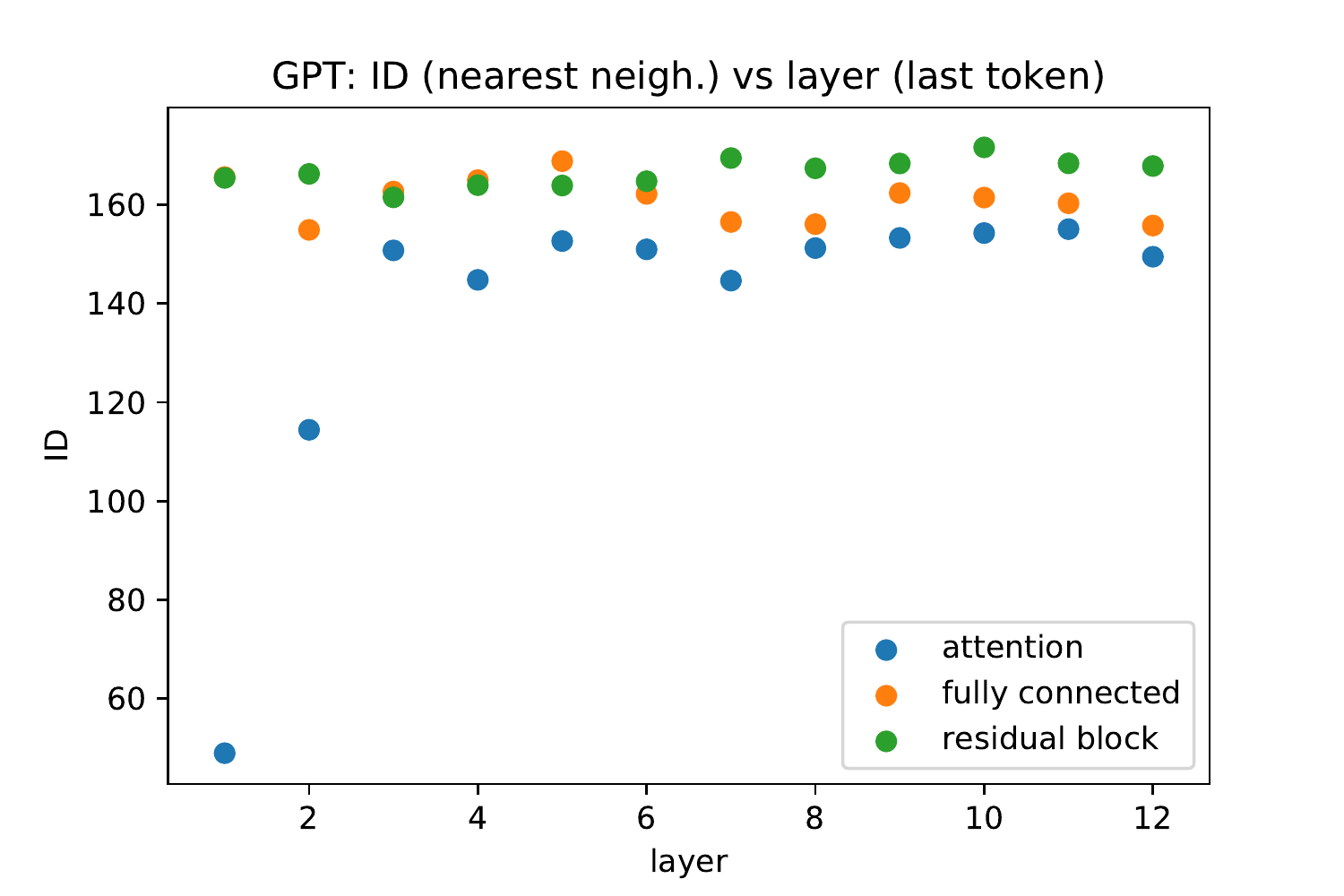} 
\caption{These figures show the ID estimates for the attention and fully-connected outputs of a 117M parameter GPT-type model, where $4/\alpha \approx 53$.  The left figure shows results from the nearest neighbor method, with 2,3, and 4 neighbors, while the right plot shows results from the MLE method.  The results roughly agree for the first layer, but the MLE method gives smaller IDs for later layers, and is likely an under-estimate. \label{fig:LanguageID}}
\end{figure}

The GPT-type  language models display power-law scaling of $L(N)$ over at least five orders of magnitude in $N$, with exponent $\alpha \approx 0.076$ \cite{kaplan2020scaling}.  This value of $\alpha$ is much smaller than those observed for many other datasets \cite{rosenfeld2019constructive}, meaning that it allows us to probe a rather different regime, where we predict the quite large value $d \gtrsim 53$.

We generated activation vectors from the `small' 117M parameter GPT-2 model using test data drawn from the same distribution as the training data \cite{radford2018improving,  radford2019language}, and measured the IDs.  Decoder-only  \cite{liu2018generating} Transformers \cite{OriginalTransformer} have a residual structure with blocks including an attention mechanism and a fully-connected component.  For each layer of blocks, one can measure the ID from the output of the attention mechanism, the fully-connected layer, or from the output of the residual re-combination.

\begin{figure}
	\centering
	\includegraphics[scale=0.48]{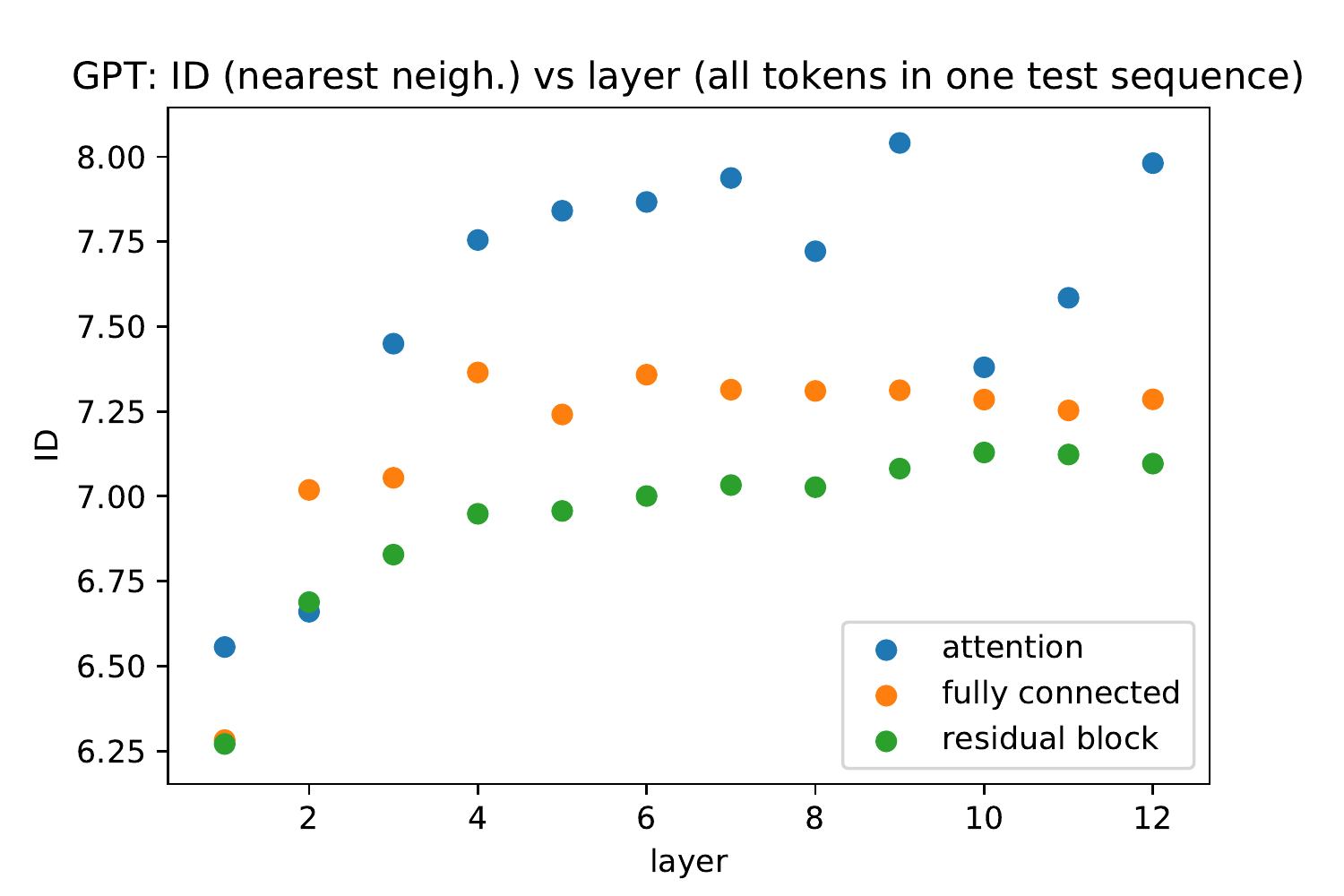} 
	\includegraphics[scale=0.48]{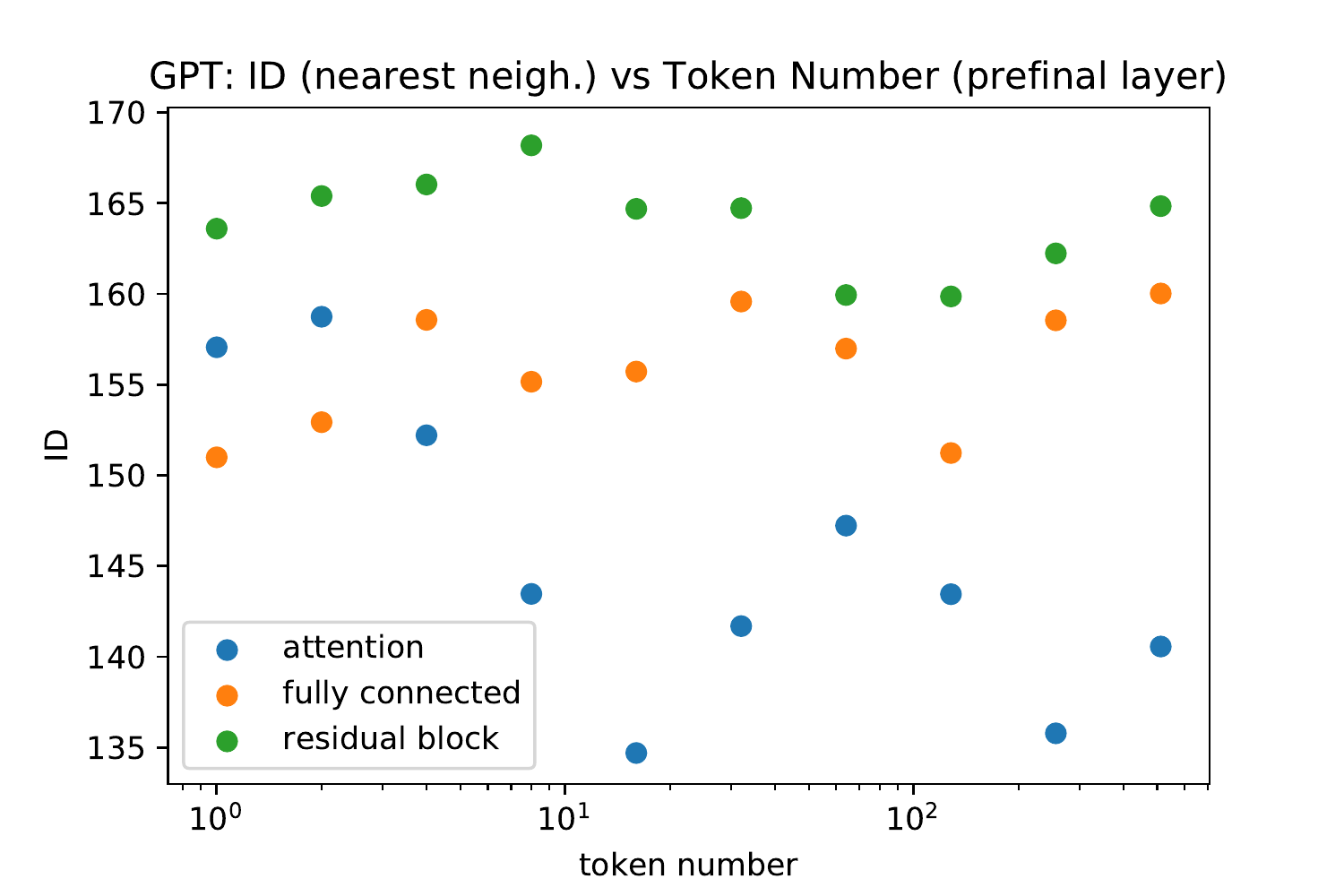} 
	\caption{ID estimates from a single 1024-token text sequence (left) and the final layer ID as measured using tokens with  fixed positions within distinct sequences (right).  The data manifold associated with a single sequence has a much, much smaller dimension than the full manifold.  \label{fig:GPTIDOneSequenceAndPerToken}}
\end{figure}

The activations that contribute to the Transformer's outputs at any given token-position depend on all activations from earlier in the sequence, except for the case of the final layer (before multiplying by the unembedding matrix).  {Thus it is only the final layer activations that can be said to capture the data manifold associated with the model's prediction for a single token.}  The mean loss over tokens has scaling exponent $\alpha \approx 0.076$, and from figure 21 of \cite{kaplan2020scaling} we see that $\alpha$ is roughly constant for tokens that occur late in any text sequence.  So we use the activations from the last token in each sequence to measure the ID, though the ID does not vary significantly across token positions (see figure \ref{fig:GPTIDOneSequenceAndPerToken}).

In figure \ref{fig:LanguageID} we plot the measured ID for the attention output, the fully connected output, and the combined output of the residual blocks for all layers.  For these measurements we used 10,000 activation vectors, each from the last token in a different text sequence (for more details see appendix \ref{app:Language}).  We see that unlike the case of image classifiers \cite{ansuini2019intrinsic}, the ID is roughly constant across layers, with the exception of the first layer, where it is significantly smaller.   If instead we measure the ID from the 1024 tokens in a single contiguous passage of text, we instead find an ID $ \approx 7$.  This strongly suggests that the data manifold has a scale-dependent structure, and may not be well-characterized by a single intrinsic dimension.

It is tempting to observe that the intrinsic dimension of activations from the first attention layer is of order $50$-$80$, which matches well with  $4/\alpha$ for these models. One might argue that this bounds the total data manifold dimensionality entering the model through its input tokens.  But as discussed above, this reasoning seems untrustworthy as an estimate of the data manifold dimensionality relevant for next-token predictions. So we take a  conservative attitude and do not use early layer IDs as an estimate of the relevant ID for scaling.

We conclude that since $d > 90$, we have that $d \geq 4/\alpha \approx 53$, which accords with our expectations (see \ref{sec:BoundNotEquality}).  Given the very small value of $\alpha$ in language modeling, it is satisfying to observe that the corresponding ID is very large.  But it would have been more exciting to discover $\alpha \approx 4/d$ for language modeling.  We do not know if the discepancy is due to added complexities from the structure of the Transformer,  special structure on the data manifold itself, a scrambling of data manifolds due to the residual structure and attention mechanism, or some other oversimplification in our theory.

\section{Related Work}

The theory of scaling we have advocated applies basic, `textbook' \cite{wasserman2006all} ideas from regression and density estimation.  Our work was also partly inspired by similar scaling relations in random forest models;  with some added assumptions, it is possible to prove them \cite{biau2012analysis}.  As one passes from classical techniques, to random forests, and then to neural networks, the models become increasingly powerful but less and less amenable to a direct analysis.  Nevertheless, we argue that similar principles apply and underly their scaling behavior.  A  similar overall perspective has been discussed by Bickel and collaborators \cite{bickel2007local}.

There is a large literature on dimensionality estimation; for a nice overview see \cite{camastra2016intrinsic}.  We have primarily used the two nearest neighbor method \cite{TwoNN}, which was based on the MLE method \cite{levina2005maximum} for distances among points in a local neighborhood.  In neural image classifiers, the intrinsic dimension of the data manifold was  studied \cite{ansuini2019intrinsic} using the TwoNN method.  They demonstrated that the ID is much smaller than the dimension estimated via linear methods such as PCA, among other interesting results.  Other authors have established a connection between ID and noisy labels \cite{ma2018dimensionalitydriven}, and demonstrated that neural models can effectively identify a low-dimensional manifold in a larger ambient space \cite{basri2016efficient}.  It would be interesting to understand the relationship between the data manifold and neural circuits \cite{olah2020zoom}, and how the manifold changes when non-robust features are eliminated \cite{notbugsfeatures}. Recent work \cite{spigler2019asymptotic} relates data dimensionality and dataset size scaling exponents for kernel methods.  The intrinsic dimension of the neural network parameter space has also been discussed \cite{li2018measuring}.

Neural scaling laws have been studied in a number of papers.  Perhaps the first work on the subject was \cite{1712.00409}.  The more recent work \cite{rosenfeld2019constructive} studies scaling with model size and dataset size, both independently and simultaneously.  Language models were studied in \cite{kaplan2020scaling}, where scaling relations with model size, dataset size, training compute, and training steps were identified.  EfficientNet \cite{DBLP:journals/corr/abs-1905-11946} displays near power-law scaling with model size, though these models are not in the underfitting regime.  

\section{Discussion}

We have proposed a theory connecting the model-size scaling exponent with the intrinsic dimension of the data manifold.  Many other neural scaling laws have been identified \cite{1712.00409, rosenfeld2019constructive, kaplan2020scaling}, including scalings with dataset size and compute budget, and fairly accurate power-law fits to learning curves.  We have focused on scaling with model size in the infinite data limit because we expect it to be the simplest and most theoretically tractable scaling relation.  Scaling with dataset size may involve issues of regularization,  requiring a balance between bias and variance, while understanding the scaling with compute would require that we contend with  optimization.  

Nevertheless, neural scaling exponents with dataset size are often very similar\footnote{Though in almost all cases \cite{rosenfeld2019constructive, kaplan2020scaling} dataset exponents are slightly larger.  This runs somewhat counter to classical expectations \cite{wasserman2006all}, where the number of parameters determines a tradeoff between bias and variance, and dataset size exponents are smaller than the bias-scaling exponents that depend on model size.} to model size exponents.  One might argue that dataset size scaling can be understood as a consequence of interpolation between points on the data manifold, and so should have a similar relationship to the data manifold dimension.  Recent works have made this case \cite{spigler2019asymptotic}.  Compute scaling exponents \cite{kaplan2020scaling} are also not far from model-size exponents, but combine optimization and model scaling.  It seems most natural to interpret them by modeling learning curves, but perhaps optimization can be re-interpreted as the identification and dissection of the data manifold.  Something like this will be necessary in order to explain the fact that larger models are much more sample efficient \cite{kaplan2020scaling}  than small models.  This may be the most impactful direction for future work.

It will be interesting to test this theory with a wider variety of models and datasets.  Generative modeling may be the ideal setting, since the abundance of unlabeled text, image, and video data provides many opportunities to train large models on nearly unlimited datasets.  In this context, it may be interesting to explore what the theory suggests for finetuning pre-trained generative models on downstream tasks.  We would expect that these tasks benefit from the pre-established existence of the data manifold; perhaps finetuning can be understood as a process of zooming-in and refining performance in a small region of this manifold.  It would also be interesting to understand how scaling relations for the loss compare to those for quantities that are not directly optimized, such as prediction accuracies.  In the case of CIFAR10 we saw that accuracy and loss exhibit similar exponents.  Finally, it's worth thinking about the extent to which larger models  perform better in reinforcement learning \cite{cobbe2019leveraging}.  Due to the non-stationary distribution in RL it may be difficult to understand model-size scaling quantitatively, and it's less clear how to apply our theory in that context.  A theory of sample efficiency scaling would be more likely to be relevant to RL.

\section*{Acknowledgments}

We thank Yasaman Bahri, Ethan Dyer, Tom Henighan, Danny Hernandez, Jaehoon Lee, and Sam McCandlish for interesting discussions and feedback.  We especially thank Ethan for sharing his notes on linear models and Yasaman for emphasizing that our theory of model size scaling might be re-purposed as a theory of dataset size scaling.  JK  has  been supported in part by NSF grant PHY-1454083.  This work was also supported in part by Open Philanthropy.  
\appendix

\addtocontents{toc}{\protect\setcounter{tocdepth}{1}}

\section{Technical Details and Minor Results}
\label{app:TechnicalDetailsMinorResults}

\subsection{Fitting}
\label{app:Fitting}

To extract the scaling exponent $\alpha$ we need to fit power-laws to the empirical $L(N)$ for trained models with $N$ parameters.  For this purpose we simply fit straight lines to $\log L$ vs $\log N$, assuming that the error in $\log L$ was independent of $N$ (ie we assumed Gaussian errors in $\log L$).  We fit from the smallest value of $N$ tested until the power-law behavior breaks down.  This point is quite clear visually in most cases, as seen in figures \ref{fig:AllTSNetworkLvsN}, \ref{fig:MSEAllTSNetworkLvsN}, and \ref{fig:CIFAR}.  For the case where we had networks with both different widths and different depths \ref{fig:AllTSNetworkLvsN} we only used the networks that performed among the best at each model size (ie we used points on the `convex hull' in the $L$ vs $N$ plane).

However, to avoid bias we determined the last point to include in the fit in the following way.  We fit a circle (parameterized by its center and radius) to the first $n \geq 3$ points in the $\log L$ vs $\log N$ plane (starting at $N = N_{\rm min}$), and evaluated $r(n)$, the radius of the best-fit circle for each $n$.  We then chose the value of $n$ that achieved the maximal radius $r$, as this is the `most linear' set of points.  Finally, we fit a straight line $\log L = -\alpha \log N + b$ to this collection of points to determine $\alpha$.

\begin{figure}
\noindent \centering{} 
\includegraphics[width=0.48\textwidth]{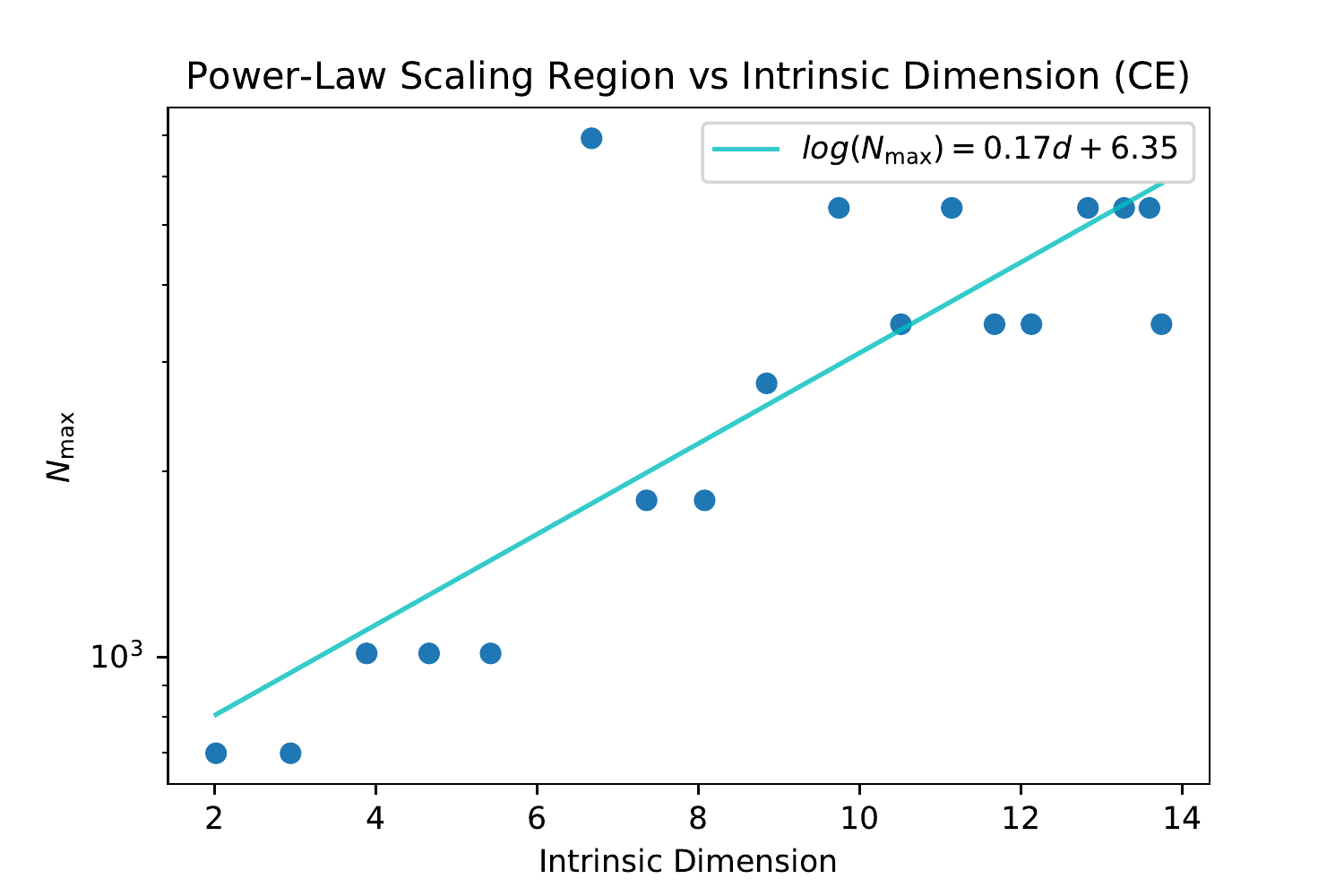}
\includegraphics[width=0.48\textwidth]{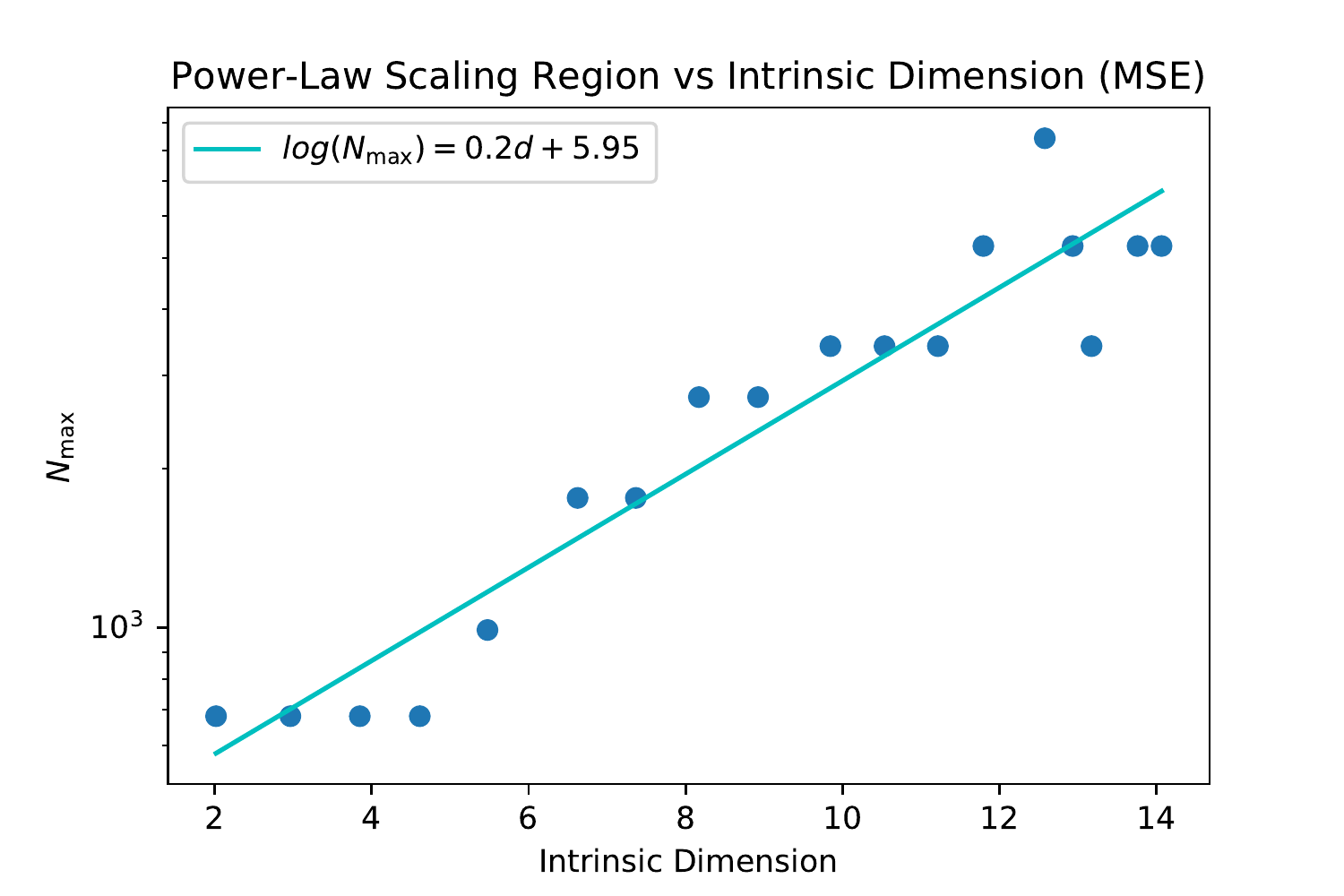}
\caption{This figure shows the maximum number of parameters $N_{\rm max}$ at which we observe  power-law scaling of $L(N)$, as a function of the intrinsic dimension, for teacher/student experiments.  This $N_{\rm max}$ is determined as described in appendix \ref{app:Fitting}. The left plot uses cross-entropy loss, while the right uses MSE loss.  This plot should be viewed as a more empirical (but less well understood) alternative to figure \ref{fig:PowerLawScalingRegionvsDimension}. \label{fig:PowerLawScalingRegionvsDimensionEndofRegion}}
\end{figure}

Note that this provides an alternative way to determine $N_{\rm max}$, the largest network in the power-law scaling region.  This was the  input for figure \ref{fig:PowerLawScalingRegionvsDimensionEndofRegion}, where we show $N_{\rm max}$ as a function of $d$ for teacher/student experiments.  

The power-law scaling breaks down in CIFAR10 and other small image datasets due to overfitting.  We do not have a complete understanding of why it breaks down for the teacher/student experiments, but it seems to be due to a failure of optimization, perhaps related to  numerical precision.  We note that the power-law behavior persists to larger model size and smaller loss with the deeper networks in figure \ref{fig:AllTSNetworkLvsN}.

\subsection{Teacher/Student Experiments}
\label{app:TSDetails}

\subsubsection{Network Architectures}

\begin{figure}
\noindent \centering{} 
\includegraphics[width=0.7\textwidth]{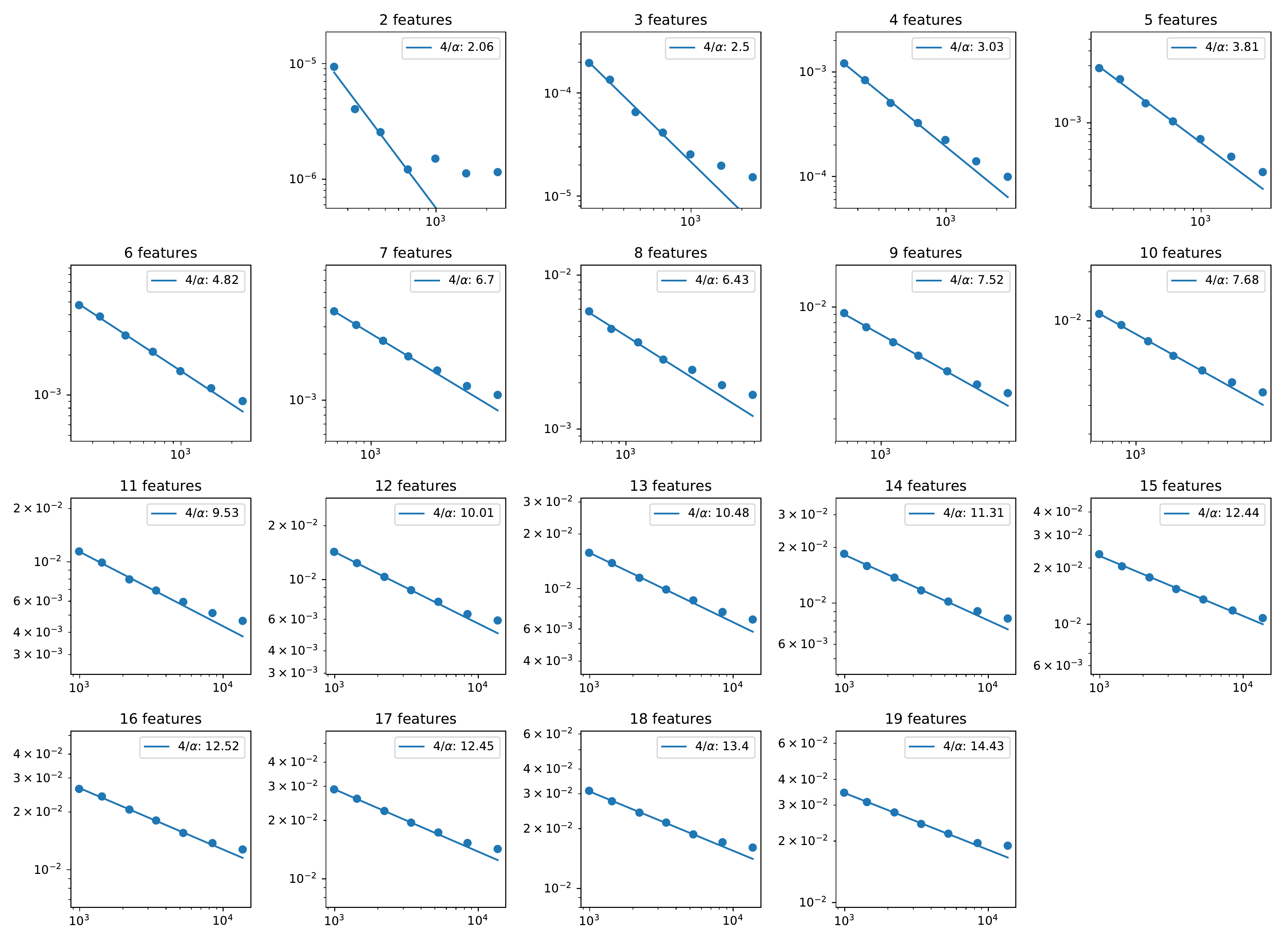}
\caption{This figure shows $L(N)$ with a MSE loss for students (all with 2 hidden layers) learning from a randomly initialized teacher with $2$-$19$ features.  Figure \ref{fig:AllTSNetworkLvsN} shows the results for cross-entropy loss.  \label{fig:MSEAllTSNetworkLvsN}}
\end{figure}

Our teacher networks had shape $[20,600,600,2] $ (i.e. $20$ dimensional input, two hidden layers of output dimension $600$, and final layer ouput of dimension $2$) for experiments with cross entropy loss (figures \ref{fig:AllTSNetworkLvsN}, \ref{fig:ProductManifold} and \ref{fig:GeneralizedLosses}), $[20,600,600,1] $ for MSE loss (figure \ref{fig:MSEAllTSNetworkLvsN}) and $[9,240,240,2]$ for cross entropy loss with vetted teacher (figure \ref{fig:VettedTeacherStudentAlphavsDim}). The teachers are randomly initialized, with biases set to zero, and weights picked from a gaussian distribution of mean zero and standard deviation $1/\sqrt{N}$, where $N$ is the input size of the layer.  We experimented with including random non-zero biases, but did not find that they significantly alter the behavior of teachers.

For experiments with mean-squared error loss, the teacher and student networks each outputted a single real value.  For experiments using a cross-entropy loss, networks output two logits, and we computed the cross entropy directly from these teacher outputs (ie we did not sample discrete values from the teacher, but used its exact output distribution).  For cross-entropy experiments we used students with 2, 3, and 4 hidden layers, and let the best performing models define the $L(N)$ fits, while for MSE loss we simply used students with 2 hidden layers.

We ran $10$ trials each for cross-entropy and MSE losses, and in each case selected the ones with the $9$ lowest losses. Intrinsic dimension calculations were done using the same $9$ networks.  For vetted teacher experiments, we took $90$ trials and computed the mean of the loss excluding the $10$ worst performing students.

\subsubsection{Optimization and LR Schedule}

We use the ADAM optimizer \cite{kingma2014adam} with default settings except for the learning rate.  In order to optimize effectively, we scanned over a grid of learning rates, and experimented with cosine, linear, and step-function learning rate schedules. We ended up using step function schedules for teacher/student experiments, and a constant learning rate for CIFAR10 and other image datasets, as these performed roughly as well or better than other choices.  We did not find it necessary to vary the overall learning rate among different network sizes, but the schedules themselves were important for optimization.  Our learning rate schedules for the various teacher/student experiments in the paper (labeled by  associated figures) are summarized in table \ref{table:training_schedules}.  

\begin{table} 
	\centering
	\begin{tabular}{ |c|c|c|c|c| } 
		\hline
		Experiment  & student & training steps & batch size & learning rate\\
		(T/S) & architecture & & &  (ADAM)\\
		\hline
		    (random) &MSE: [20,n,n,1]&  0-200k  &200 & 0.01\\ 
		 figures \ref{fig:TeacherStudentAlphavsDim}, \ref{fig:ProductManifold}, \ref{fig:GeneralizedLosses}  & CE: [20,n,n,2]   
		    &  200-220k & 1000& 0.01\\
		    && 220-240k &4000& 0.001\\
		\hline
		 (vetted)   &&0-100k  &200&0.01\\ 
		 figure \ref{fig:VettedTeacherStudentAlphavsDim} & [9,n,n,2]
		 & 100-150k& 200& 0.001\\
		 &  &150-170k  & 200& 0.0001\\ 
		\hline
			\end{tabular}
	\caption{Architectures and training schedules for Teacher/Student experiments in the paper, referenced by the figures in which the results are described. }
\label{table:training_schedules}
\end{table}

\subsection{Vetting  Teachers to Increase Intrinsic Dimension}
\label{sec:VettingTeachers}

In figure \ref{fig:TeacherStudentAlphavsDim}, the ID is typically smaller than the number of features, especially when the latter is large.  One might worry that this indicates ID measurements are inaccurate.
In fact, we believe that this occurs partly because randomly initialized teacher networks do not typically  produce fully generic functions of their inputs.

\begin{figure}
\noindent \centering{} 
\includegraphics[width=0.48\textwidth]{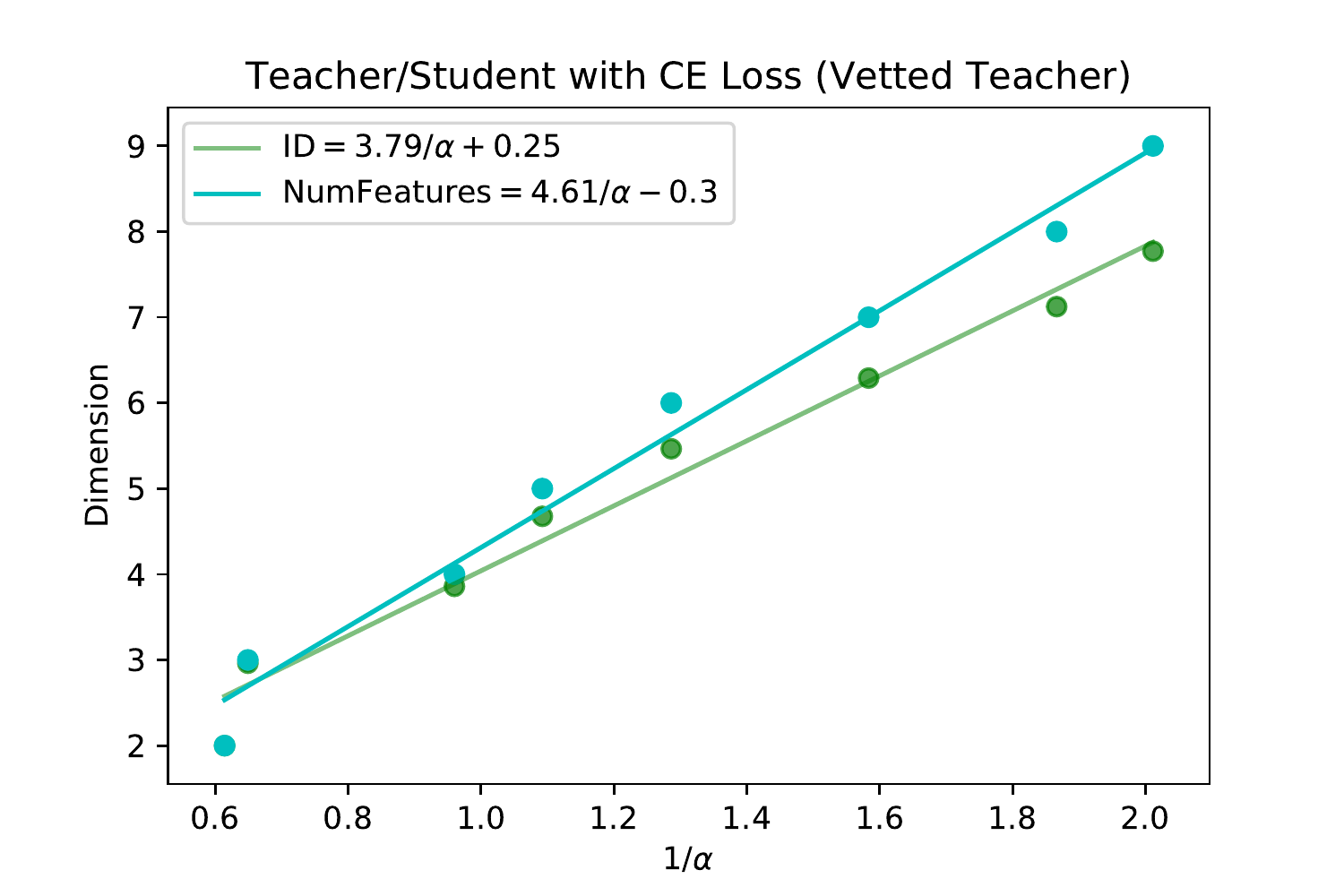}
\caption{This figure shows the number of features and  ID vs $1/\alpha$ for vetted teachers.   ID is still  smaller than the number of input features, but vetting partially closes the gap.  Compare the  slope of $4.61$ for number of features vs $1/\alpha$ here to the left of figure \ref{fig:TeacherStudentAlphavsDim}, where the slope was $5.48$.  Slopes for ID vs $1/\alpha$ are very similar with or without vetting. \label{fig:VettedTeacherStudentAlphavsDim}}
\end{figure}

We can partially remedy this problem by generating a large number of teachers and vetting them, keeping only those  that produce the most complicated and non-linear functions of their inputs.  The result is pictured in figure \ref{fig:VettedTeacherStudentAlphavsDim}, where we repeat the experiment of section \ref{sec:BasicTSExperiment} with up to $9$ features.  We see that sufficiently vetted teachers have ID nearly equal to their feature count, and that the relationship $\alpha \approx \frac{4}{d}$ continues to hold.  

Presumably many vetting procedures could be successfully applied to filter the teacher networks.  
To increase the complexity and non-linearity of teachers so that ID would better match the number of input features, we followed this ad-hoc approach:
\begin{enumerate}
	\item For a given teacher, we took a random slice along each input coordinate axis (i.e. the values of the other coordinates are chosen uniformly at random from $[-1/2,1/2)$). We performed linear regression on this slice and computed the score($R^2$, the coefficient of determination), and took the mean of the scores across coordinate axes. A low score implies more non-linearity.
	\item We repeated this procedure $200$ times  and computed the mean score of all the trials. This is the score for the teacher.
	\item We iterated over $5000$ randomly generated teachers and selected the one with the minimum score. 
\end{enumerate}

\subsection{CNNs on CIFAR10, MNIST, FMNIST, and SVHN}
\label{app:CNNs}

\begin{figure}
	\centering
	\includegraphics[scale=0.32]{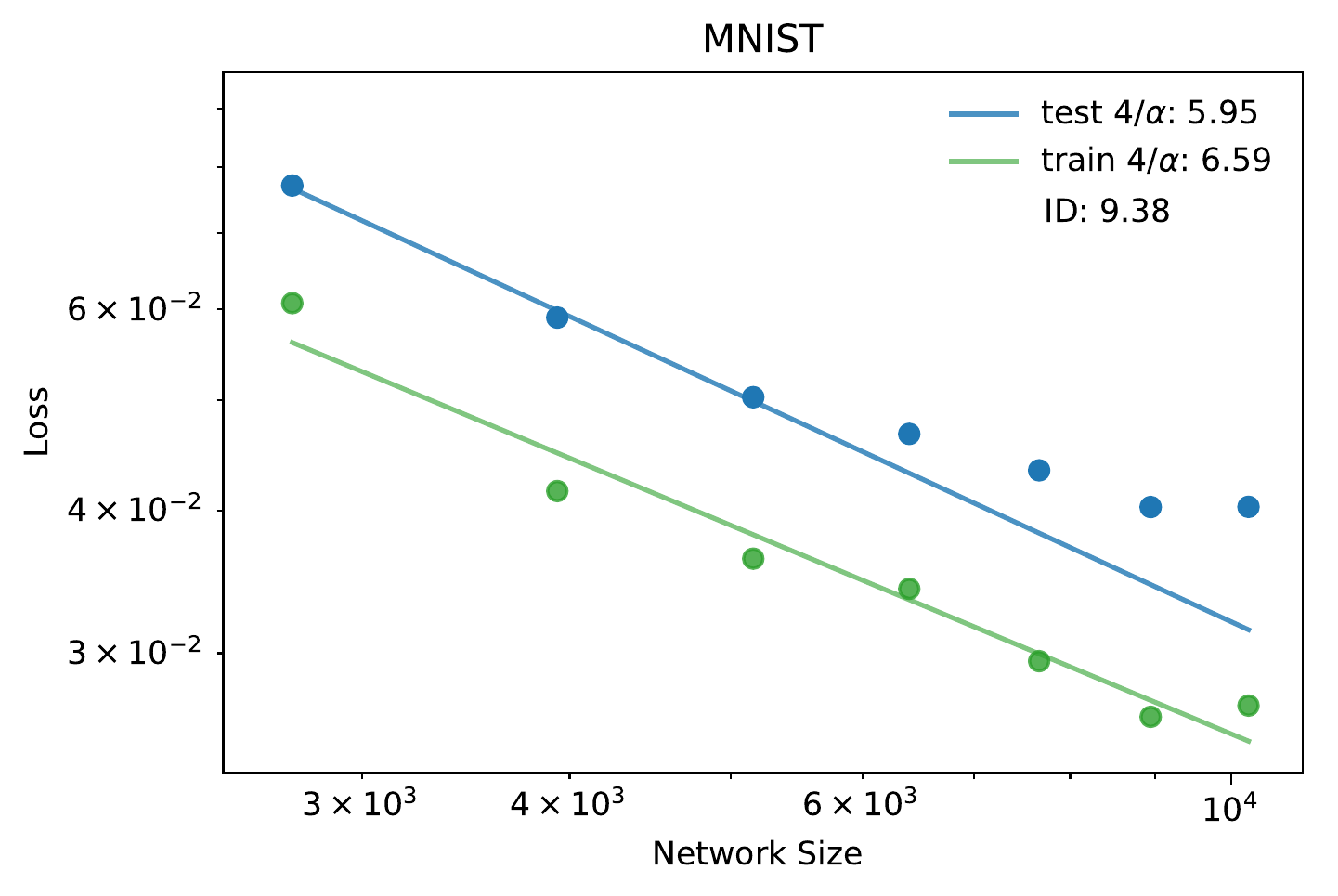} 
	\includegraphics[scale=0.32]{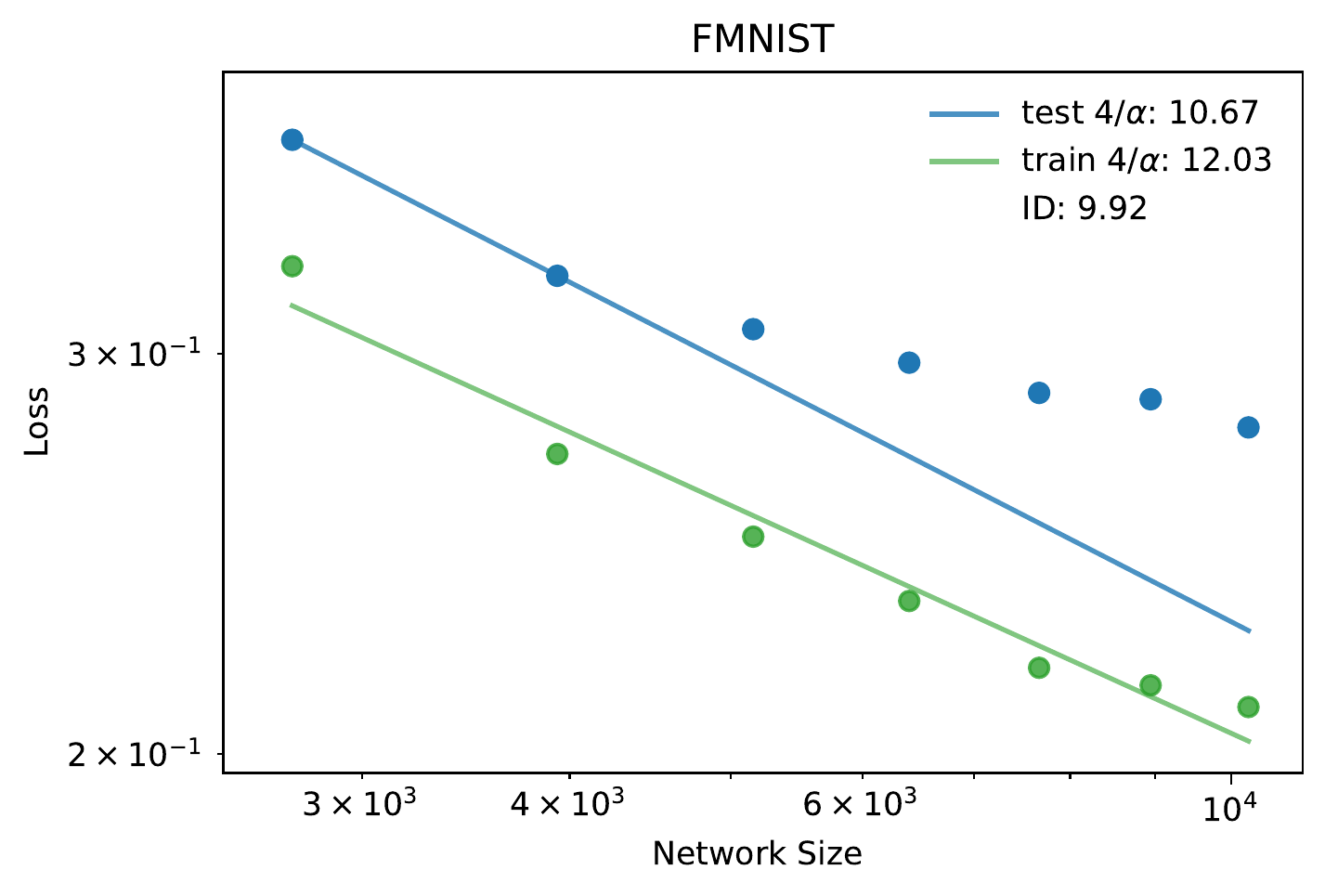} 
	\includegraphics[scale=0.32]{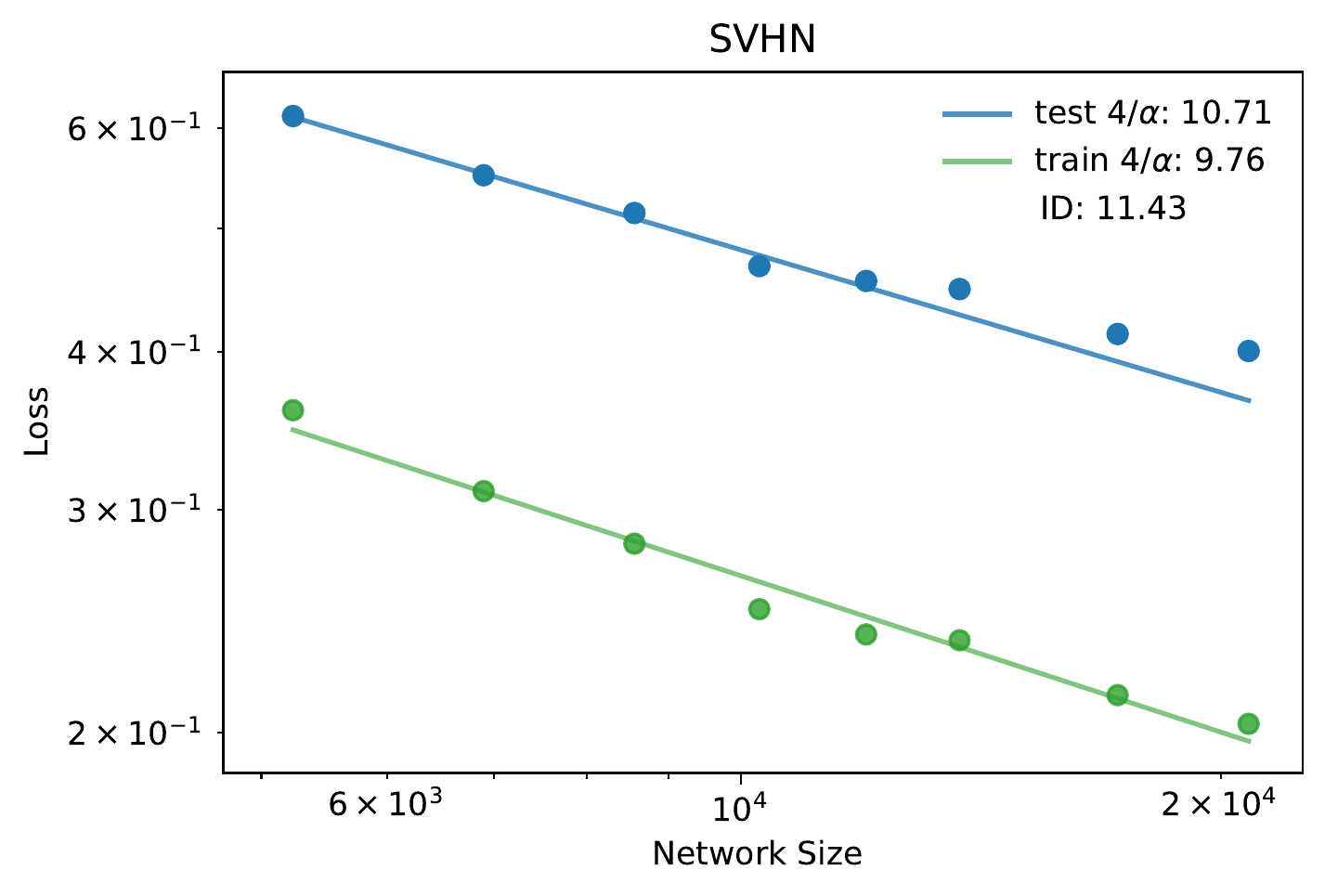} 
	\caption{This shows train and test loss on MNIST, Fashion MNIST, and test loss on SVHN, along with the exponents and ID measurement. \label{fig:FMNIST}}
\end{figure}

For CIFAR10 we used the architecture from the tensorflow CNN tutorial \cite{tensorflow2015-whitepaper}, and modified the channel width. The architecture is recorded in table \ref{table:CIFAR_architecture}.

The networks were trained for $50$ epochs with the ADAM optimizer with default hyperparameters. We use $40$ iterations of each network and average the loss (on log scale) over the iterations.
Note that we record the test and training loss at the early stopping point where the test loss reaches its minimum value.  These are the results in figure \ref{fig:CIFAR}.

For MNIST  \cite{lecun-mnisthandwrittendigit-2010}, fashion MNIST \cite{fmnist}, and SVHN \cite{netzer2011reading}, we use a slightly smaller network (3 instead of 4 hidden layers) with architecture shown in table \ref{table:FMNIST_architecture}. We used a smaller network in the hopes of identifying a power-law scaling region without  significant overfitting.

For MNIST and fashion MNIST, we ran each network for $20$ trials and took the mean loss (on log scale). The networks were trained for $50$ epochs with the ADAM optimizer with default hyperparameters. As with CIFAR10, we take the minimum test loss during training (i.e. early stopping), and also report training loss at this point.

For SVHN, the networks were trained for $5$ epochs with both training and additional datasets used for training (total $604$k images), and test dataset ($26$k images) for testing. We used default hyperparameters.

\begin{table} 
	\centering
	\begin{tabular}{ |c|c|c| } 
		\hline
		Layer  &Output shape \\
		\hline
		Conv2D &$(32, 32, n)$\\
		\hline
		MaxPooling2D&$(16, 16, n)$\\
		\hline
		Conv2D &$(16, 16, 2n)$\\
		\hline
		MaxPooling2D&$(8, 8, 2n)$\\
		\hline
		Conv2D  &$(6, 6, 2n)$\\
		\hline
		Dense&$(64)$ \\
		\hline
		Output &$ (10)$\\ \hline
		
	\end{tabular}
	\caption{Architecture of the CNN network used for CIFAR10. We chose $n$  in the range $1\leq n\leq 13$ to minimize overfitting.  All convolutions were $3 \times 3$ with unit stride, and the images have 3 colors, so the network has a total of $N = 714 + 4640 n + 54 n^2$ parameters.}
	\label{table:CIFAR_architecture}
\end{table}

\begin{table} 
	\centering
	\begin{tabular}{ |c|c|c| } 
		\hline
		Layer  &Output shape \\
		\hline
		Conv2D &$(28, 28, n)$\\
		\hline
		MaxPooling2D&$(14, 14, n)$\\
		\hline
		Conv2D &$(12, 12, n)$\\
		\hline
		MaxPooling2D&$(6, 6, n)$\\
		\hline
		Dense&$(32)$ \\
		\hline
		Output &$ (10)$\\ \hline
		
	\end{tabular}
	\quad
	\begin{tabular}{ |c|c|c| } 
		\hline
		Layer  &Output shape \\
		\hline
		Conv2D &$(32, 32, n)$\\
		\hline
		MaxPooling2D&$(16, 16, n)$\\
		\hline
		Conv2D &$(14, 14, n)$\\
		\hline
		MaxPooling2D&$(7, 7, n)$\\
		\hline
		Dense&$(32)$ \\
		\hline
		Output &$ (10)$\\ \hline
		
	\end{tabular}

	\caption{Architecture of the CNN network used for MNIST and fashion MNIST (left) and SVHN (right). All convolutions were $3 \times 3$ with unit stride.}
	\label{table:FMNIST_architecture}
\end{table}

\subsection{Scaling of KL Divergence with Piecewise Linear Logits}
\label{app:ScalingKL}

We assume the logits $c_i(x)$ are linear in a small region of volume $s^d$ we take to surround the origin, and that the underlying probability distribution $f_i(x)$ over $k$ discrete choices is smooth.  The loss in this region is
\be
L &=& \sum_{i=1}^k \int d^d x f_i(x) \log \frac{f_i(x)}{q_i(x)}  
\ee 
where $\log q_i(x) = c_i(x) + \log \left( \sum_{j=1}^k e^{c_j(x)} \right)$.  If we write $q_i(x) = f_i(x) + \delta_i(x)$ then as is well known
\be
L &=& \sum_{i=1}^k \int d^d x f_i(x) \log \frac{f_i(x)}{f_i(x) + \delta_i(x)} 
\nn \\
&=&  \int d^d x \sum_{i=1}^k f_i(x) \left(0 - \frac{\delta_i(x)}{f_i(x)} + \frac{1}{2} \left( \frac{\delta_i(x)}{f_i(x)} \right)^2 + \cdots \right)
\nn \\
& \approx &  \int d^d x \sum_{i=1}^k \frac{1}{2}  \frac{\delta_i(x)^2}{f_i(x)} 
\ee 
After optimization the linear $c_i(x)$ will determine a $\delta_i(x)$ that is quadratic in $x$, and so the loss per unit volume will scale as $s^{4}$, as claimed.

\section{Review of Intrinsic Dimension Estimation Methods}
\label{app:IDfromNeighbors}

In this section we review the two nearest neighbor method \cite{ansuini2019intrinsic} and explain that it can be extended to $k$-nearest neighbors.  Then we note that the same analysis derives the maximum likelihood method \cite{levina2005maximum}.

\subsection{The Two Nearest Neighbor Method}

Assume that points are drawn from a distribution with density $\rho(x)$ with support on a $d$-dimensional manifold in a potentially much higher dimensional ambient space.    We will see that $\rho(x)$ drops out of our results, assuming that it is constant across the first few nearest neighbors, so we will drop its explicit $x$-dependence in what follows.

The probability of finding $n$ points from the dataset in a region with $d$-dimensional volume $V$ is Poisson:
\be
P_n(V) = \frac{(\rho V)^n}{n!} e^{- \rho V}
\ee
To see this, note that in an infinitesimal volume $\delta V$, $P_0 = 1 - \rho \delta V$ and $P_1 = \rho \delta V$, with all $P_{n > 1} = 0$.  Thus the generating function for $P_n$ in a finite volume $V$ can be found by taking the product of binomial distributions over all $\delta V$ in $V$, giving
\be
G(x; V) =  \lim_{\delta V \to 0}  \left(  (1 - \rho \delta V) + x \rho \delta V \right)^{\frac{V}{\delta V}} = \sum_{n=0}^\infty \frac{(x \rho V)^n}{n!} e^{- \rho V}
\ee
The coefficients of $x^n$ are the $P_n$ above.

With this result in hand, we can consider the distribution of nearest-neighbor distances.  Consider some point in the dataset.  The probability for its nearest neighbor to be in $[r_1, r_1 + dr]$ is given by the product of the probability that there are no points in $r < r_1$ times the probability of finding a point in the shell $r_1 < r < r_1 + dr$, which is
\be
P(r_1) dr_1 = \left( d \rho \omega_d r_1^{d-1} dr_1 \right) e^{-\rho \omega_d r_1^d} 
\ee
where $\omega_d$ is the volume of a unit $d$-ball.  This result easily generalizes to the case where there are many $r_i$ corresponding to the first $k$ nearest neighbors.  For example for two nearest neighbors we find
\be
P(r_1, r_2) dr_1 dr_2 = (\rho \omega_d d)^2 e^{-\rho \omega_d r_2^d} r_1^{d-1} r_2^{d-1} dr_1 dr_2
\ee
since we are demanding that there are two points on two infinitesimal shells at radii $r_1, r_2$ and no points otherwise.

\begin{figure}
	\centering
	\includegraphics[scale=0.31]{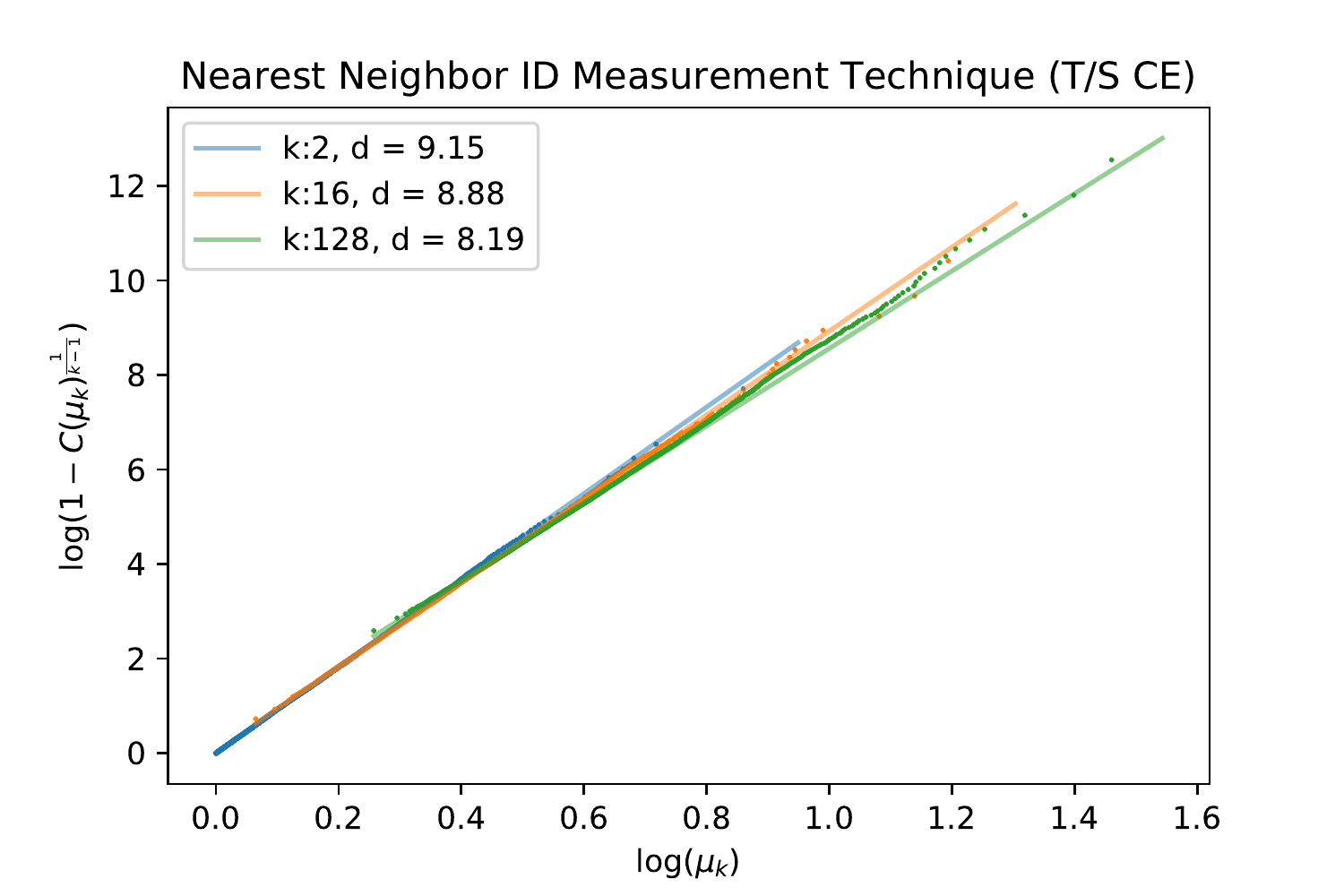} 
	\includegraphics[scale=0.31]{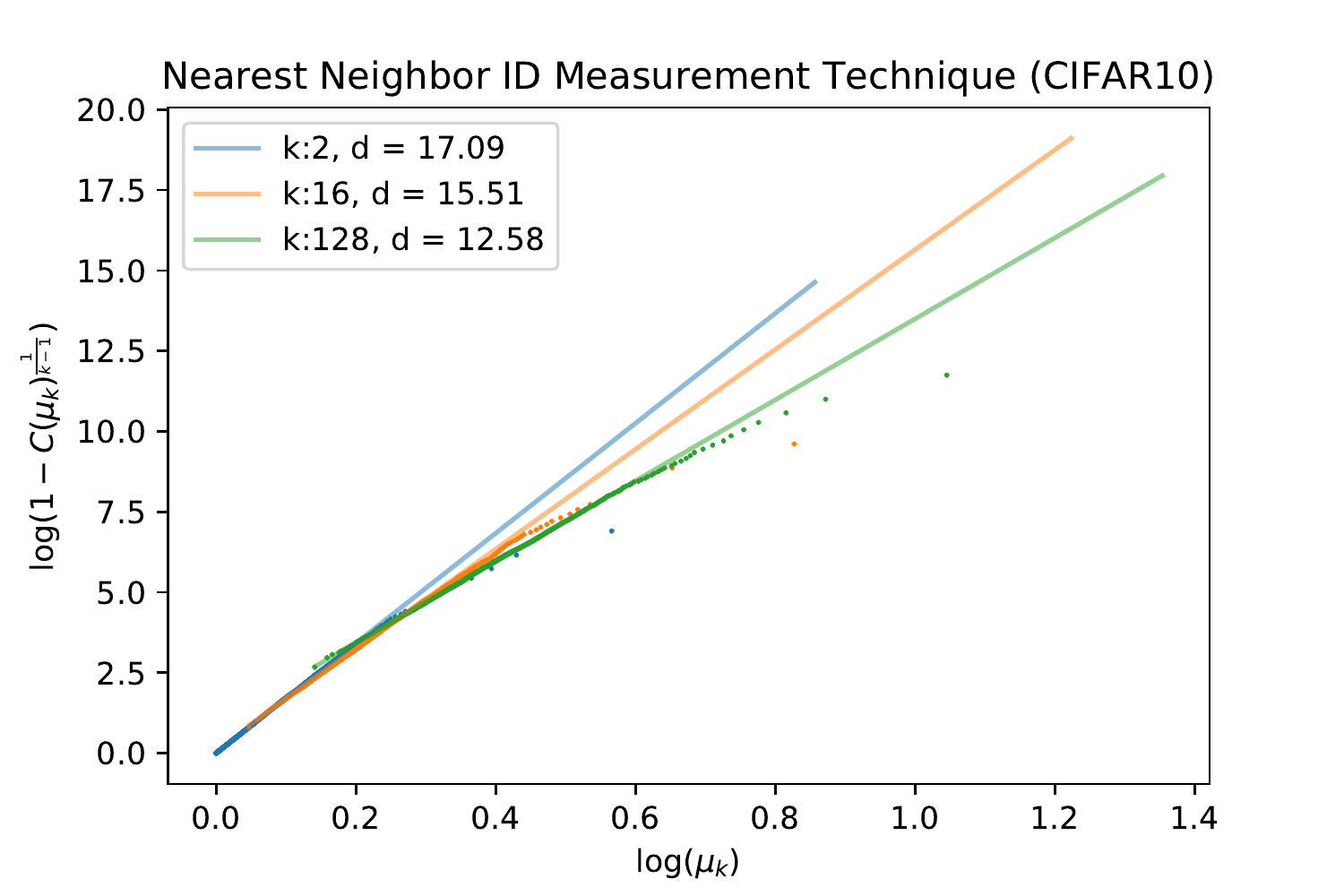} 
		\includegraphics[scale=0.31]{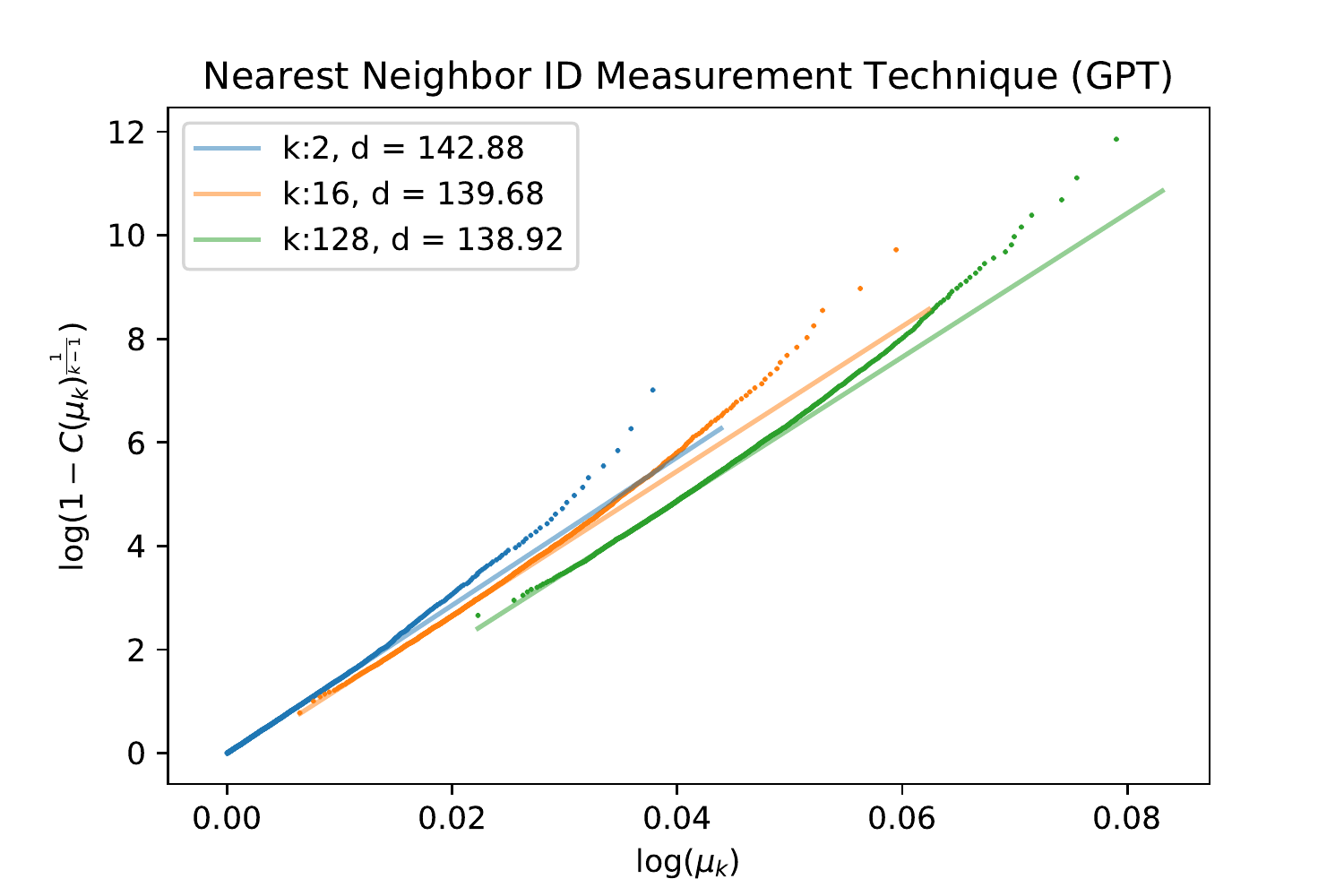} 
	\caption{This figure shows the relationship in equation \ref{eq:Cvsmu}, which we use to determine the ID using the nearest neighbor method.  We display examples using teacher/student data, CIFAR10, and GPT.  \label{fig:logFvslogmu} \label{fig:GPTIDvsNumvct}}
\end{figure}

Now we can compute the distribution over nearest neighbor distances, and their ratios.  The TwoNN method \cite{ansuini2019intrinsic} is based on the distribution of the ratio $\mu_2 = r_2/r_1$, which we can compute by integrating over $r_1, r_2$ while fixing their ratio:
\be
P(\mu_2) &=&  \int dr_1 dr_2  \delta \left(\mu_2 - \frac{r_2}{r_1} \right) (\rho \omega_d d)^2 e^{-\rho \omega_d r_2^d} r_1^{d-1} r_2^{d-1} 
\nn \\
&=&  \int dr_1 (\rho \omega_d d)^2 e^{-\rho \omega_d \mu^d r_1^d} r_1^{2d-1}  \mu_2^{d-1}
\nn \\
&=& \frac{d}{\mu_2^{d+1}}
\ee
This means that the  cumulative distribution for $\mu_2$ is
\be
C(\mu) = \int_1^\mu \frac{d}{\mu_2^{d+1}} d\mu = 1 - \frac{1}{\mu_2^d}
\ee
This means that \emph{we can identify the dimension $d$ by measuring the slope of a linear fit of $\log \mu_2$ vs $\log(1 - C(\mu_2))$}.  That's the TwoNN method, as seen in figure \ref{fig:logFvslogmu}.

\subsection{Extension to $k$-Neighbors and MLE}

The beauty of the TwoNN method \cite{ansuini2019intrinsic}  is that it uses very short-distance information, and so it's plausible that the density $\rho(x)$ can be well-approximated as a constant.  A down-side of this method is that it primarily measures the dimension on short scales.  This can be  mitigated by applying the method while sampling different numbers of points from the data distribution, but it's also easy to validate the TwoNN method by simply using more neighbors.  

Let's see what happens with three neighbors, and then we will  generalize.  We can  compute the distribution of $\mu_2 = r_2/r_1, \mu_3 = r_3 / r_1$, and  use it for validation.  We have
\be
P(\mu_2, \mu_3 ) &=&  \int dr_i \delta \left(\mu_2 - \frac{r_2}{r_1} \right) \delta \left(\mu_3 - \frac{r_3}{r_1} \right) (\rho \omega_d d)^3 e^{-\rho \omega_d r_3^d} (r_1 r_2 r_3)^{d-1}
\nn \\
&=&  \int dr_1  (\rho \omega_d d)^3 e^{-\rho \omega_d \mu_3 r_3^d} r_1^{3d-1} \mu_2^{d-1} \mu_3^{d-1}
\nn \\
&=&  \frac{2 d^2 \mu_2^{d-1} }{\mu_3^{2d+1} }
\ee
Intuitively, large $\mu_3$ becomes unlikely because it implies that there are few points inside a large radius, but with fixed $\mu_3$, a  larger value of $\mu_2$ is more probable due to the larger volume at large radius.

We find a nice simplification when we study $P(\mu_3)$ and its cumulative distribution after marginalizing over $\mu_2$.  The probability distribution is
\be
P(\mu_3) = \int_1^{\mu_3} d\mu_2  \frac{2 d^2 \mu_2^{d-1} }{\mu_3^{2d+1} } =  \frac{2 d }{\mu_3^{2d+1} } \left( \mu_3^d - 1\right)
\ee
The cumulative distribution is then
\be
C(\mu_3) = \left(1 - \frac{1}{\mu_3^{d}} \right)^2
\ee
Thus we also find a simple method for identifying $d$ based on $\mu_3$ alone, namely
\be
d = \frac{\log \left( 1 - \sqrt{C(\mu_3)} \right) }{\log \mu_3}
\ee
This directly generalizes the TwoNN; in practice we measure $d$ via a linear fit to the numerator as a function of the denominator in this expression.

\begin{figure}
	\centering
	\includegraphics[scale=0.32]{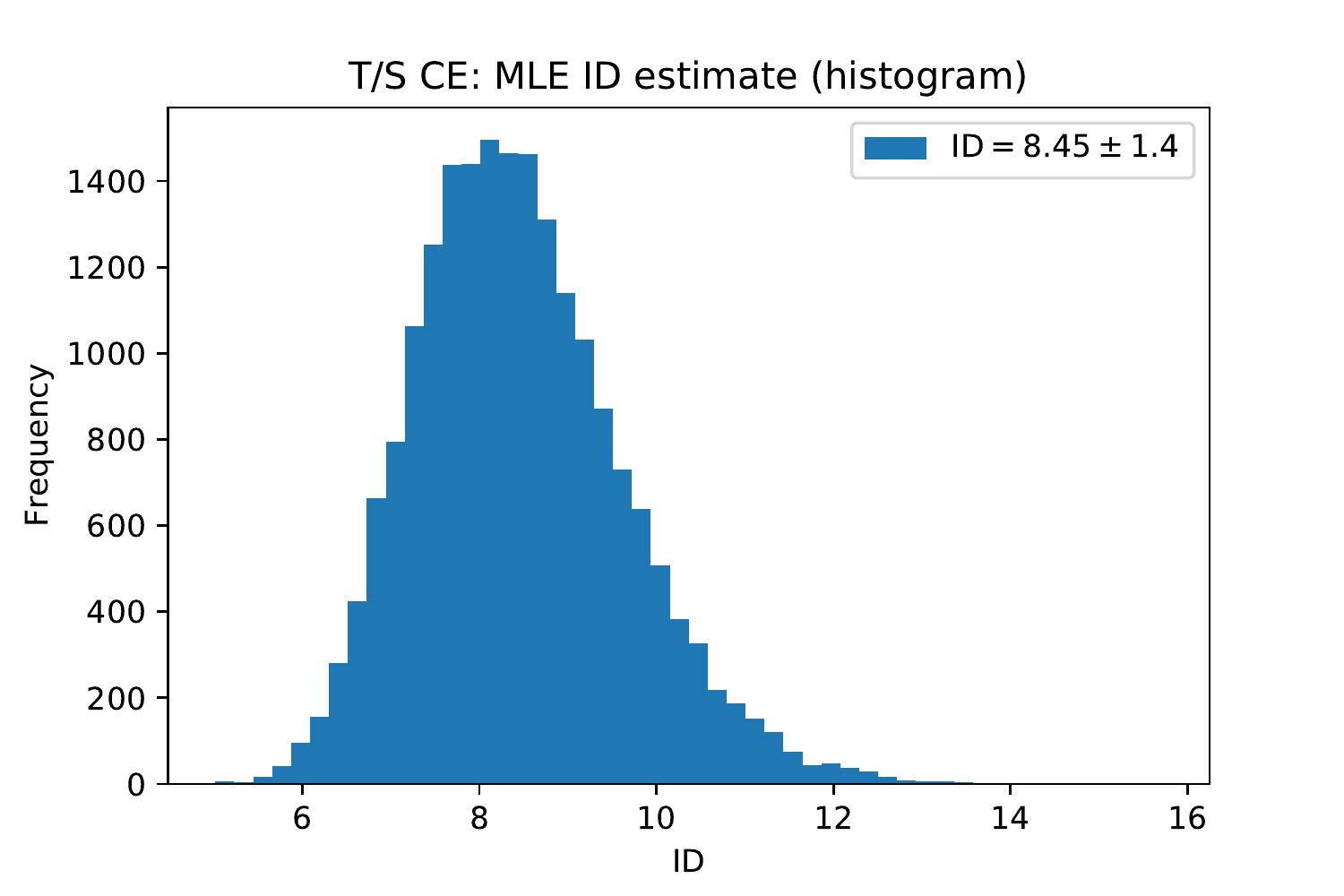}
	\includegraphics[scale=0.32]{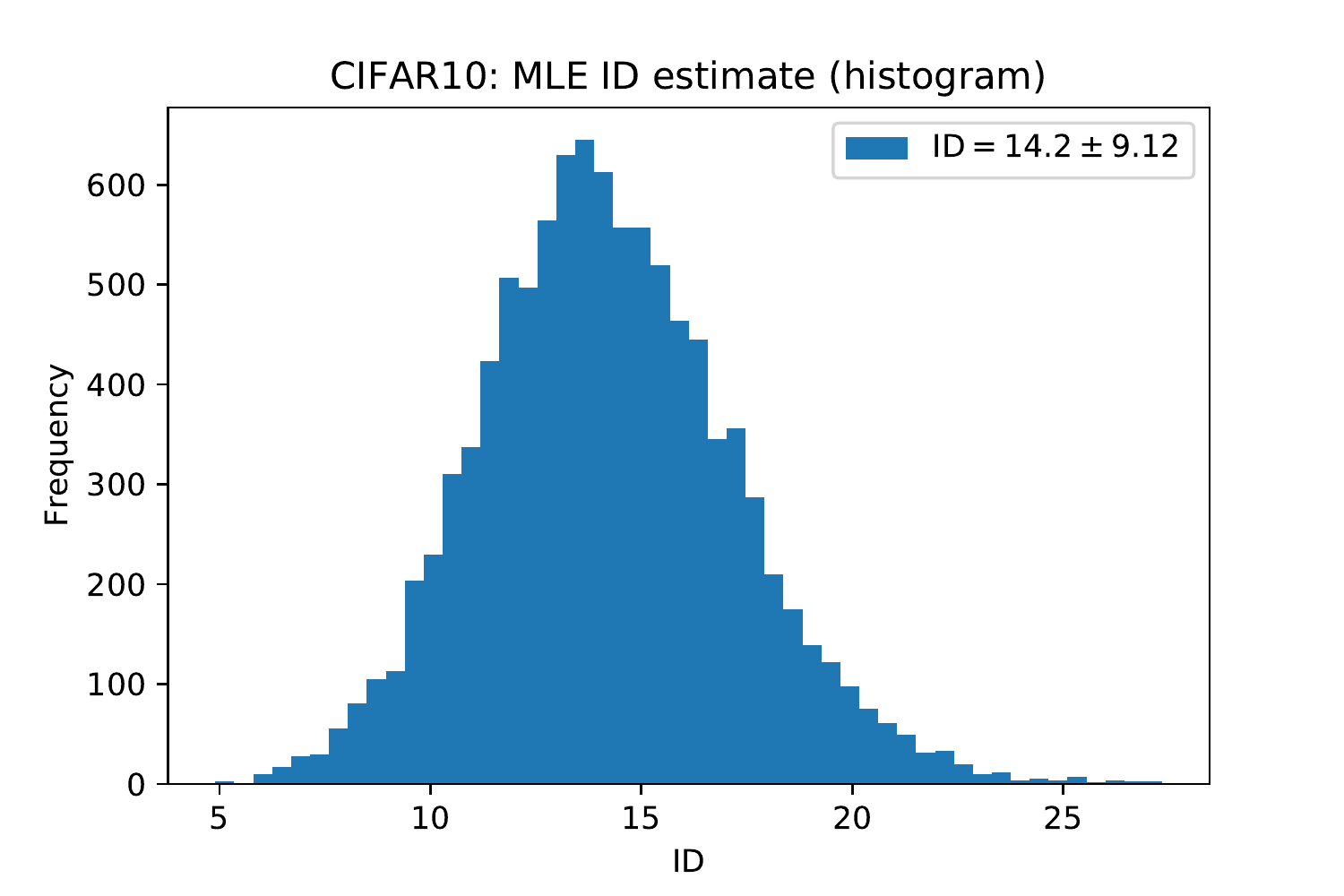}
		\includegraphics[scale=0.32]{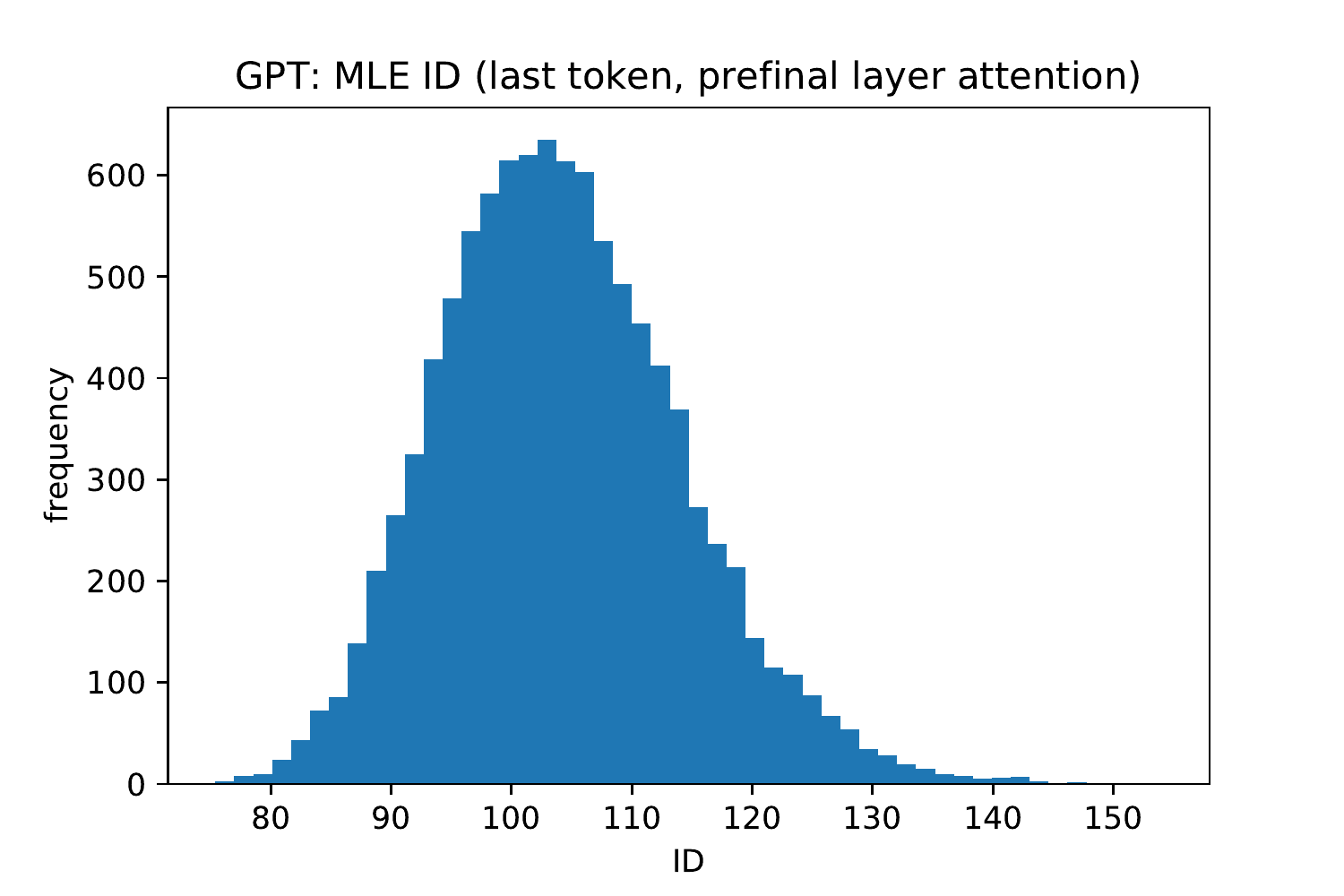} 
	\caption{ These figures show a histogram of the results for $d$ from MLE (with $100$ neighbors) among all of the points used for measurement.  On the left we have a teacher with 10 features, in the middle we have the $n=5$ CNN trained on CIFAR10, while on the right we have the GPT model's prefinal attention output for the last token in the text sequence.  Smaller numbers of neighbors typically give larger IDs. \label{fig:CIFARMLE}}
\end{figure}

Generalizing to $k$ neighbors, the probability distribution for $\mu_2, \cdots, \mu_k$ is
\be
P(\mu_i) = d^{k-1} (k-1)!  \frac{\prod_{i=2}^{k-1} \mu_i^{d-1} }{\mu_k^{1+ d(k-1)}}
\ee
for $\mu_i = r_i/r_1$.  This can be used directly for maximum likelihood estimation \cite{levina2005maximum}. If we maximize $\log P$ with respect to $d$ we find
\be
d = \frac{k-1}{(k-1) \log \mu_k - \sum_{j=2}^{k-1} \log \mu_j} 
\ee
In fact, this MLE estimator is biased;  the unbiased estimator is \cite{levina2005maximum}
\be
d = {\mathbb E} \left[\frac{k-2}{(k-1) \log \mu_k - \sum_{j=2}^{k-1} \log \mu_j} \right]
\ee
In practice, we can compute the RHS for all points in the manifold (after fixing some value for the number of neighbors $k$) and compute the mean.  We display a histogram of the MLE estimates over many points in the data manifold for two examples in figure \ref{fig:CIFARMLE}.  The variance provides some measure of the errors.  Alternatively, we could directly measure $\log P$ and evaluate the likelihood as a function of $d$.  The variance of this estimator was studied in \cite{levina2005maximum}.  They also found numerically that it can be useful to tune of the value of $k$, as very small $k$ overestimates ID while large $k$ underestimates ID.

We can use these results to extend the TwoNN method in a simple way to general $k$.
Marginalizing over all but $\mu_k$, we find that 
\be
P(\mu_k) = \frac{(k-1) d }{\mu_k^{(k-1)d+1} } \left( \mu_k^d - 1\right)^{k-1}
\ee
which leads to the cumulative distribution
\be
C(\mu_k) = \left(1 - \frac{1}{\mu_k^{d}} \right)^{k-1}
\ee
and the formula
\be
\label{eq:Cvsmu}
d = \frac{\log \left( 1 - C(\mu_k)^{\frac{1}{k-1}} \right) }{\log \mu_k}
\ee
for the $k$th nearest neighbor.  This can be used as a cross-check for TwoNN.  For examples of the relationship between the numerator and denominator with various $k$, and the relevant fits, see figure \ref{fig:logFvslogmu}.  Just as with MLE, we find empirically that larger $k$ leads to smaller estimates of ID (see figure \ref{fig:NumberNNUsedID}).

\section{Examples and Tests of Intrinsic Dimension Estimation}
\label{app:TestingID}

The MLE and TwoNN methods have been tested and demonstrated by their authors \cite{levina2005maximum, ansuini2019intrinsic}. We conduct a few tests with synthetic data.  Then we provide some  other examples of the ID measurement process, including errors, using our student/teacher, CIFAR10, and language data.   

\subsection{Tests on Synthetic Data}

\begin{figure}
\centering
\includegraphics[width=0.38\textwidth]{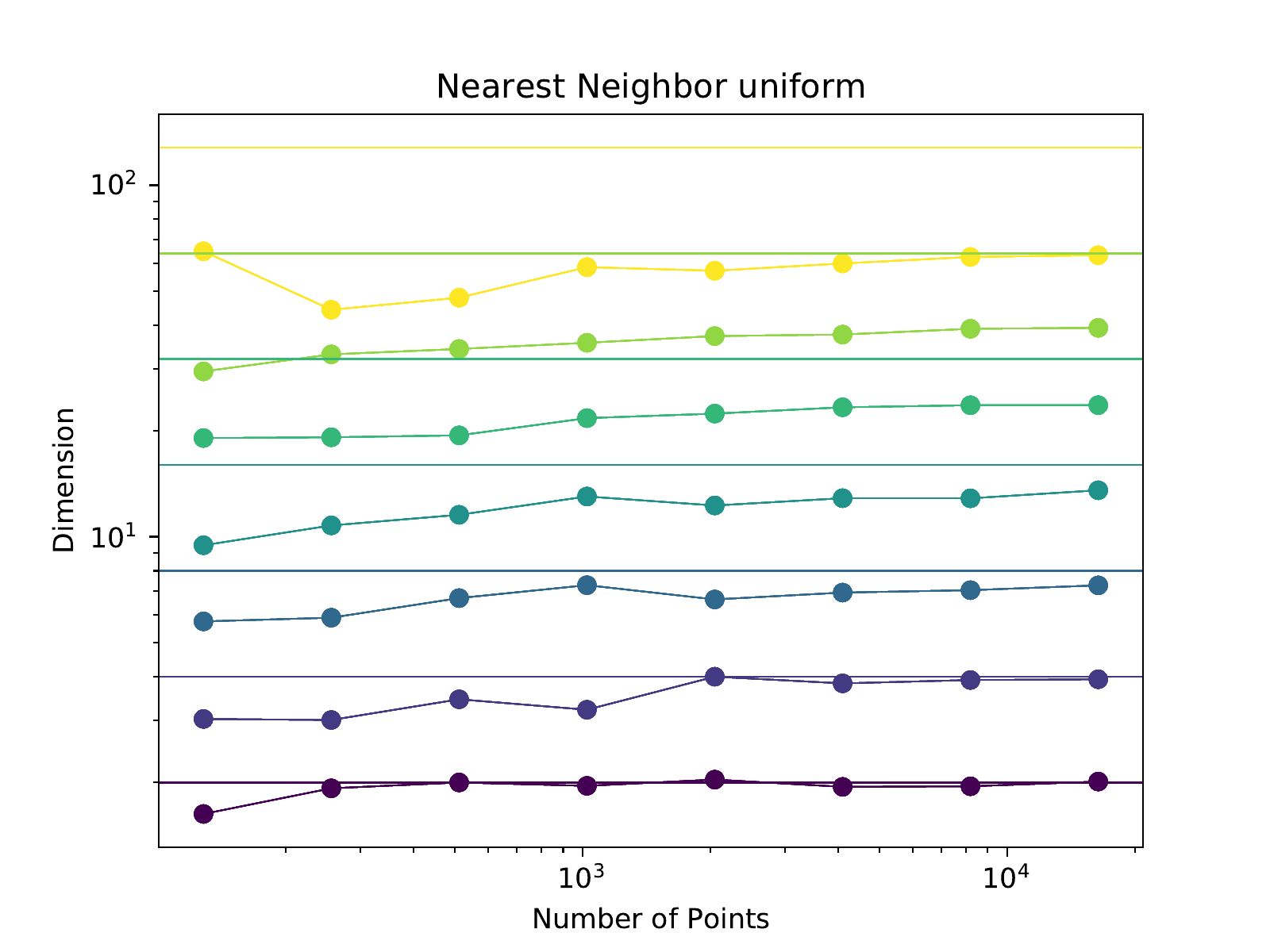} 
\includegraphics[width=0.38\textwidth]{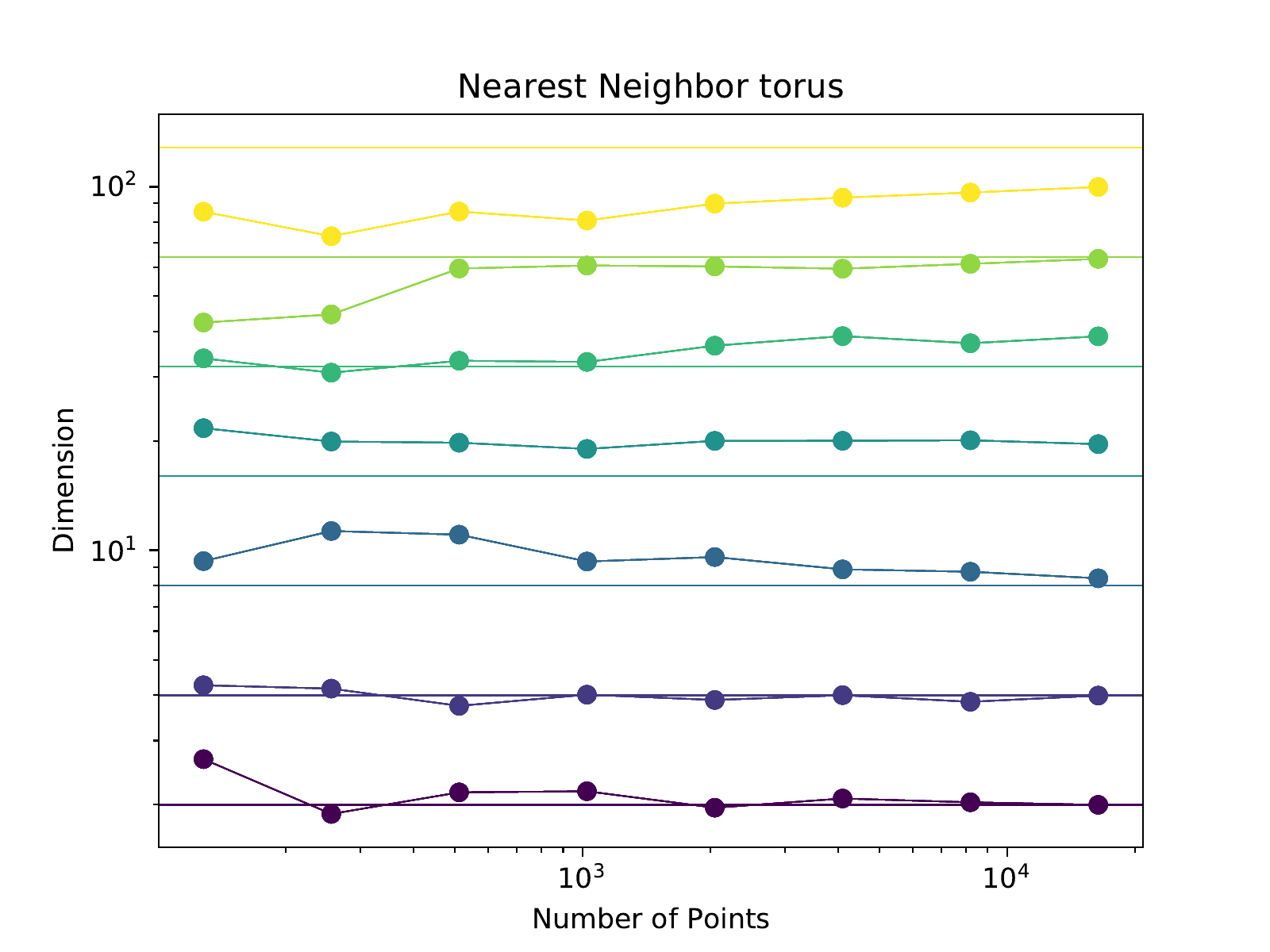}
\includegraphics[width=0.38\textwidth]{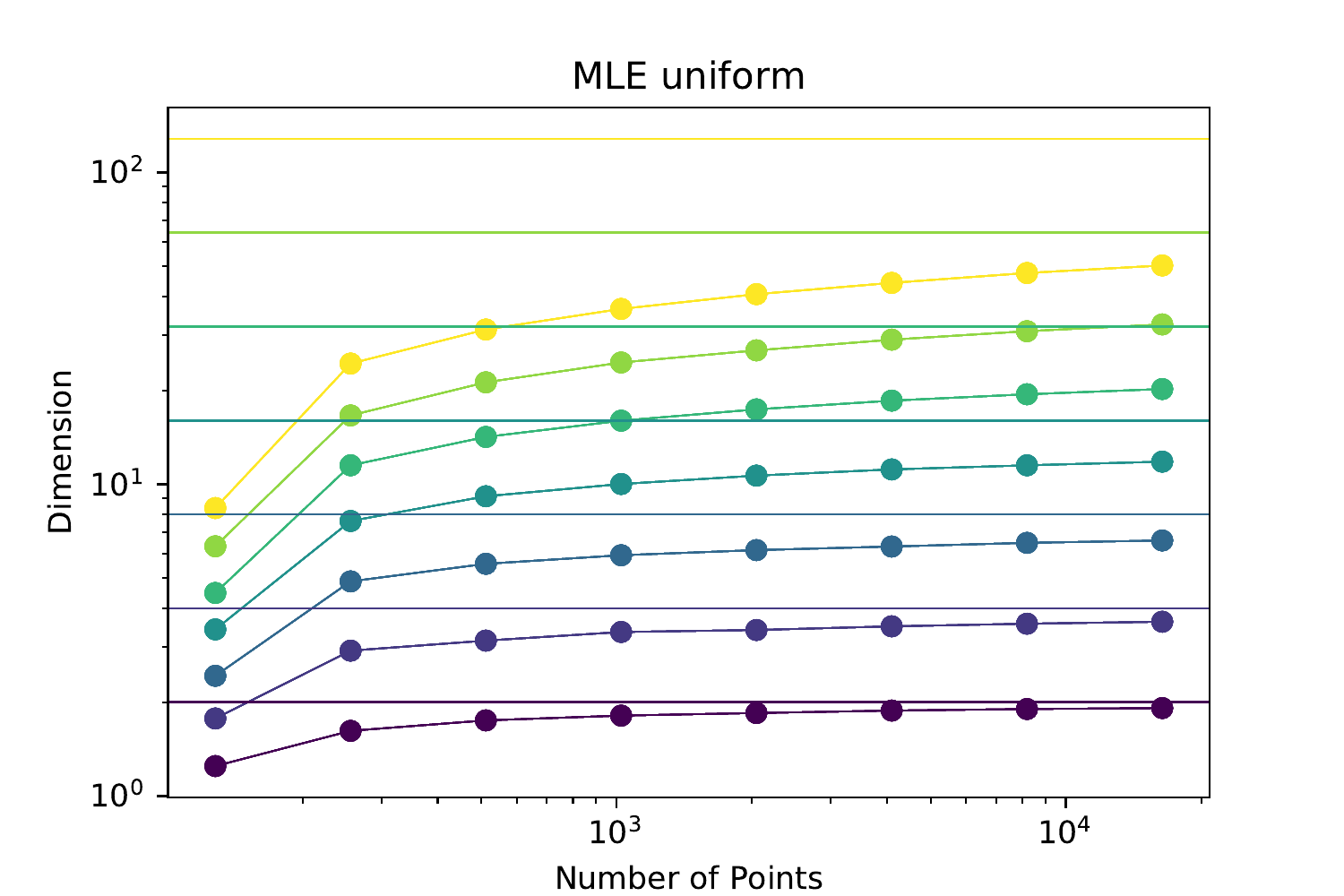} 
\includegraphics[width=0.38\textwidth]{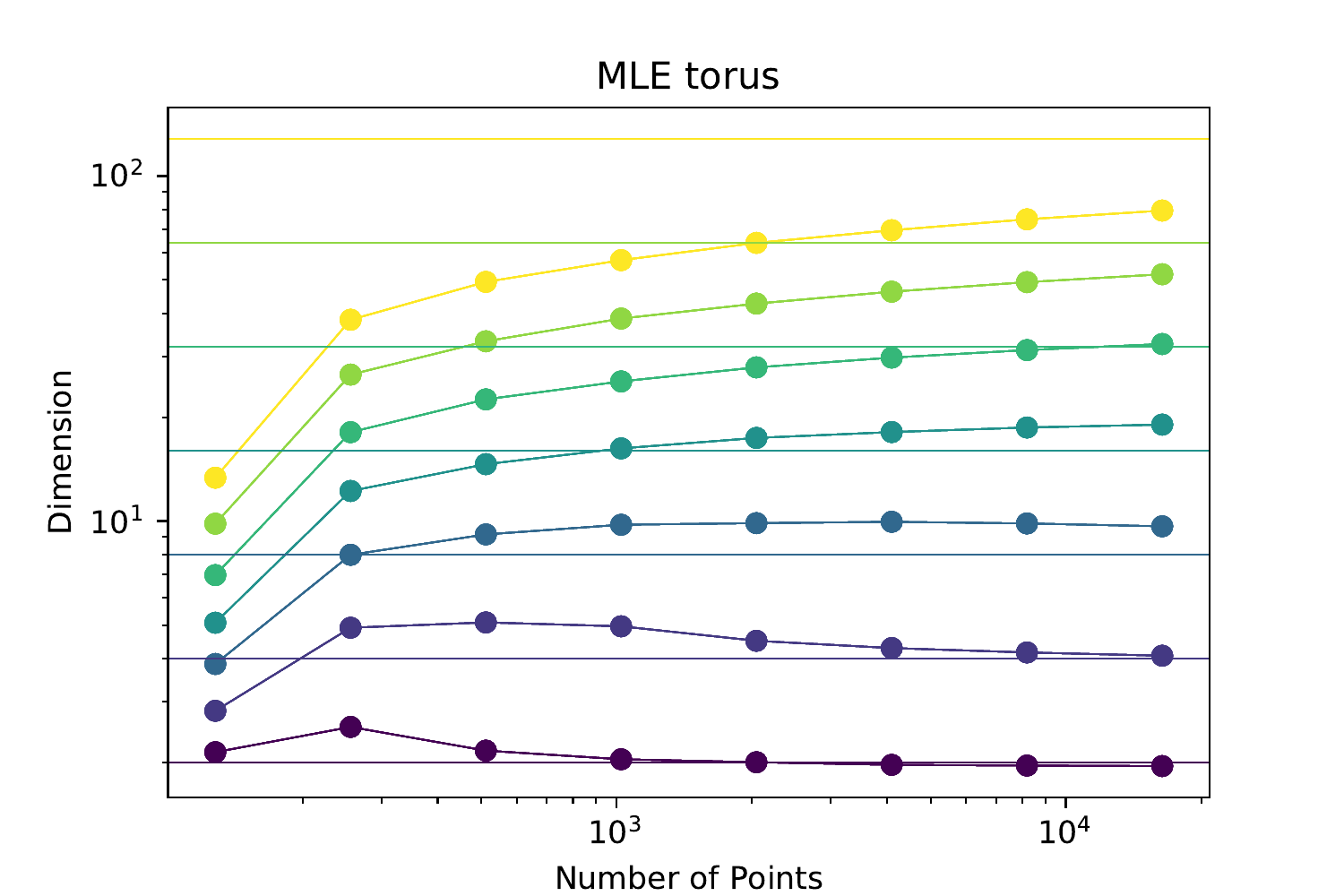}
\caption{ Here we show measured ID as a function of the number of points in the dataset used for the measurement, for both the TwoNN (top) and MLE (bottom) methods (with $k=100$).  The left plots show a uniform distribution in the hypercube $[0,1]^d$, while the plot on the right show a $d$-torus embedded in $2d$ dimensions.  \label{fig:SyntheticData} }
\end{figure}

As a baseline test, we evaluate the TwoNN and MLE methods on synthetic datasets with dimensions ranging from $2$ to $128$, with results in figure \ref{fig:SyntheticData}.  We display synthetic data on the hypercube $[0,1]^d$ as well as a $d$-torus $S^1 \times S^1 \times \cdots \times S^1$ embedded in $2d$ dimensions (in the simplest way, by embedding each circle factor in 2 Euclidean dimensions).  

We notice that 1) results are more accurate for smaller $d$, with quite reliable results for the TwoNN method for $d \lesssim 20$, 2) at large $d$ all methods tend to underestimate the true ID, but 3) its certainly possible to both under and over-estimate the true ID, and measurements are not necessarily even monotonic with the number of points used for the measurement.  We also see that for the torus the  ID estimates are reasonably accurate even for dimensions $\sim 100$, though there's certainly no guarantee that this will hold for unknown data manifolds.  

As other authors have noted \cite{camastra2016intrinsic}, the ID is under-estimated on the hypercube, likely because cubes have sharp boundaries and corners which reduce the number of neighbors.  Similarly, we believe that the ID is often over-estimated for the torus because (due to the curvature of the circles in the embedding space) points are often closer together than they would be in flat Euclidean space.  We have also seen as shown in \cite{levina2005maximum} that for small $k$ the MLE method typically overestimates ID.  The NN method seems a bit less sensitive to $k$ as compared to MLE.

\subsection{Tests on Neural Network Activations}
\label{app:Language}

\begin{figure}
	\centering
	\includegraphics[scale=0.32]{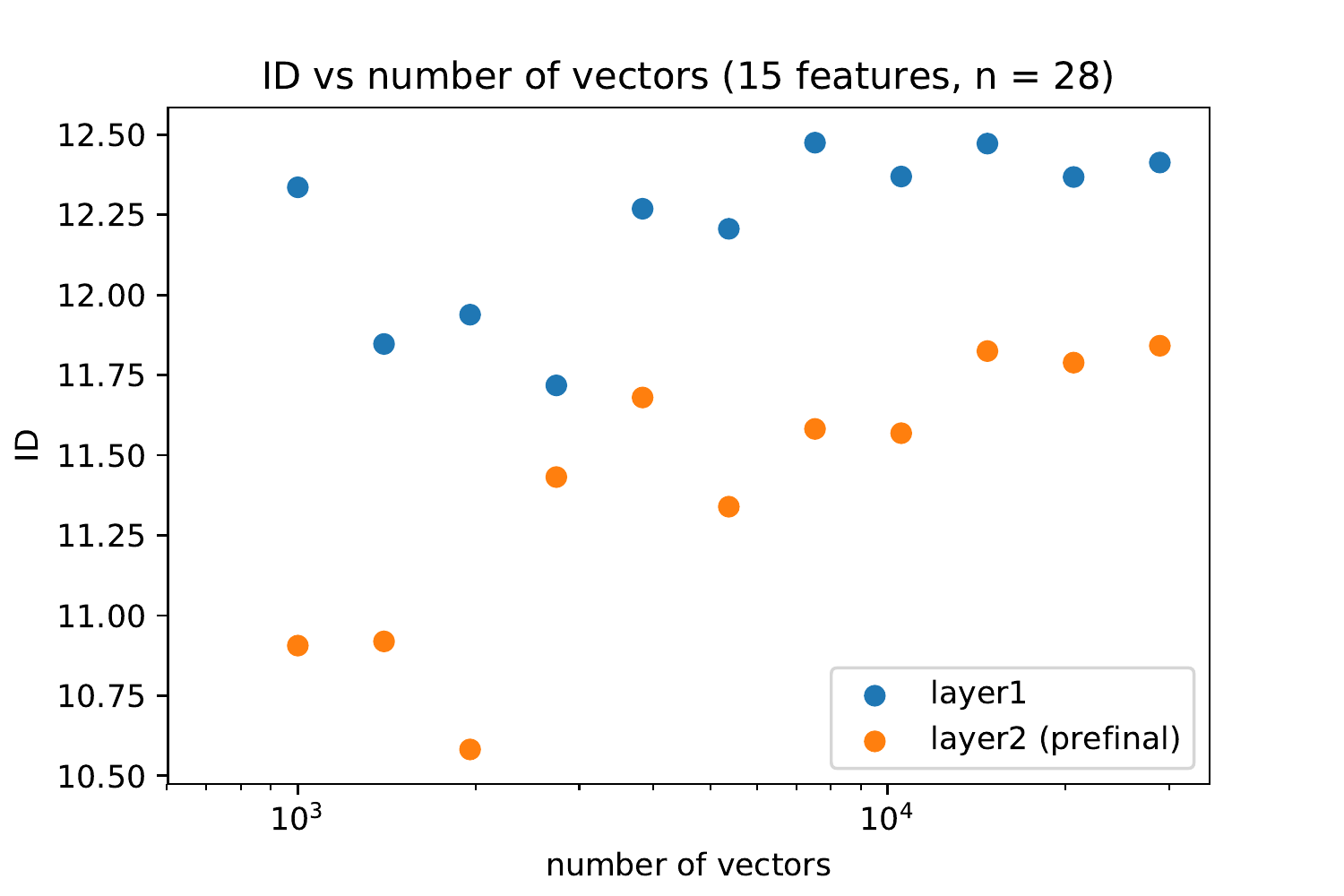} 
	\includegraphics[scale=0.32]{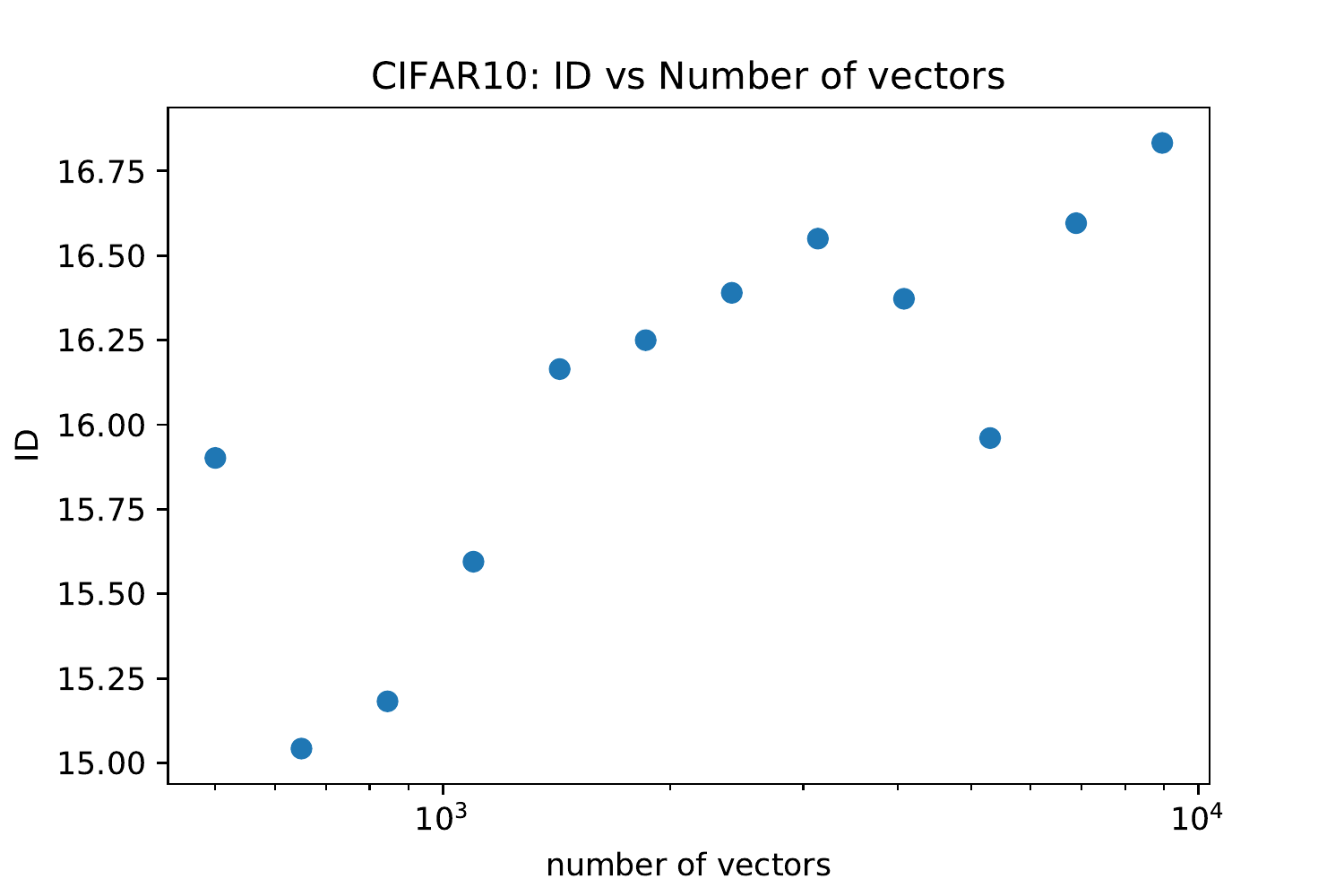}
	\includegraphics[scale=0.32]{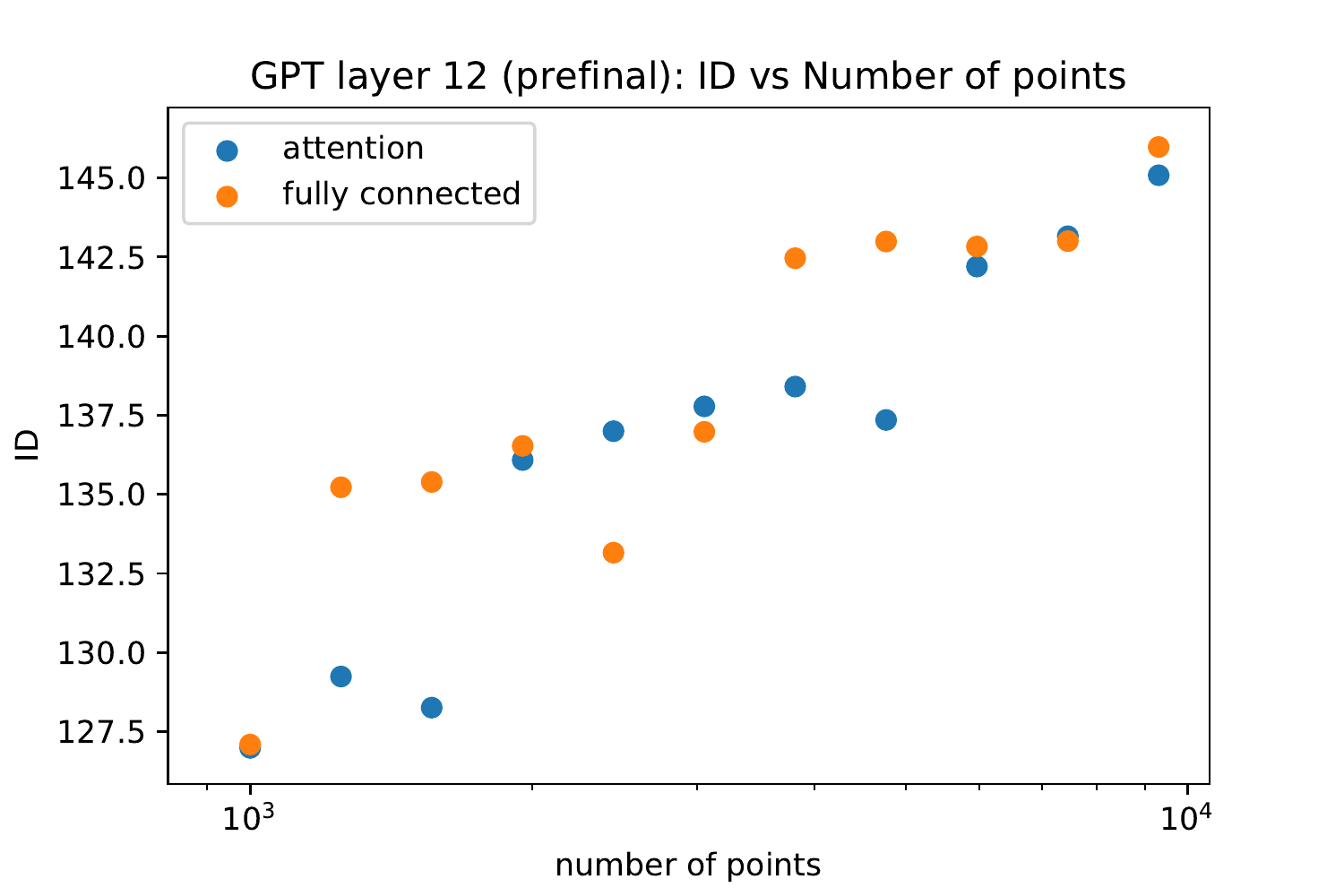} 
	\caption{Variation of Intrinsic Dimension(ID) with number of vectors for a single student network (left), for the last layer of an $n=5$ CNN trained on CIFAR10  (middle), and also for the last layer and last token of GPT (right). The student is of size $[15,28,28,2]$ and was trained on teacher with $15$ features. \label{fig:LayerVariationIDTS} \label{fig:NumberVctUsedID}}
\end{figure}

\begin{figure}
\centering
\includegraphics[scale=0.48]{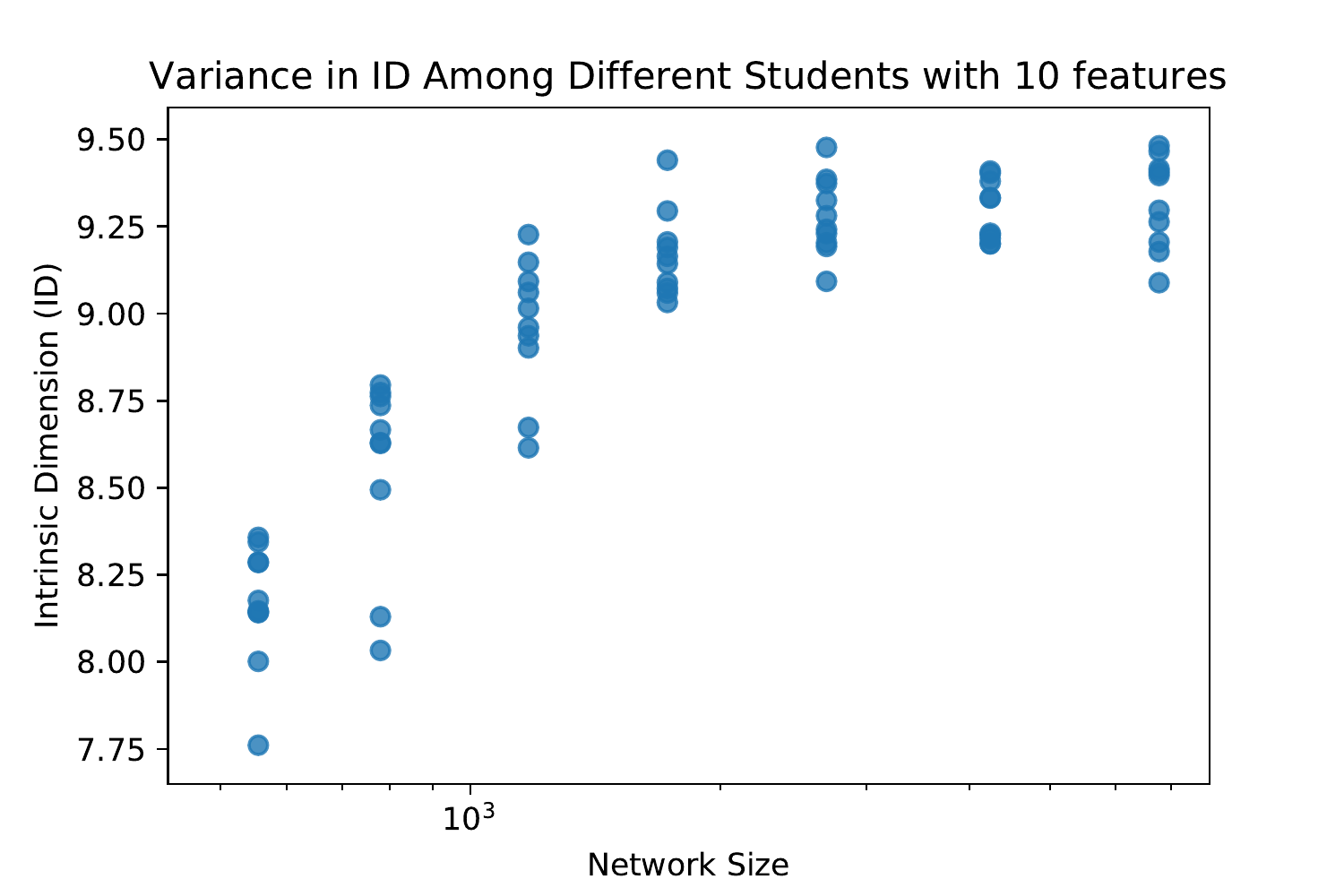} 
\includegraphics[scale=0.48]{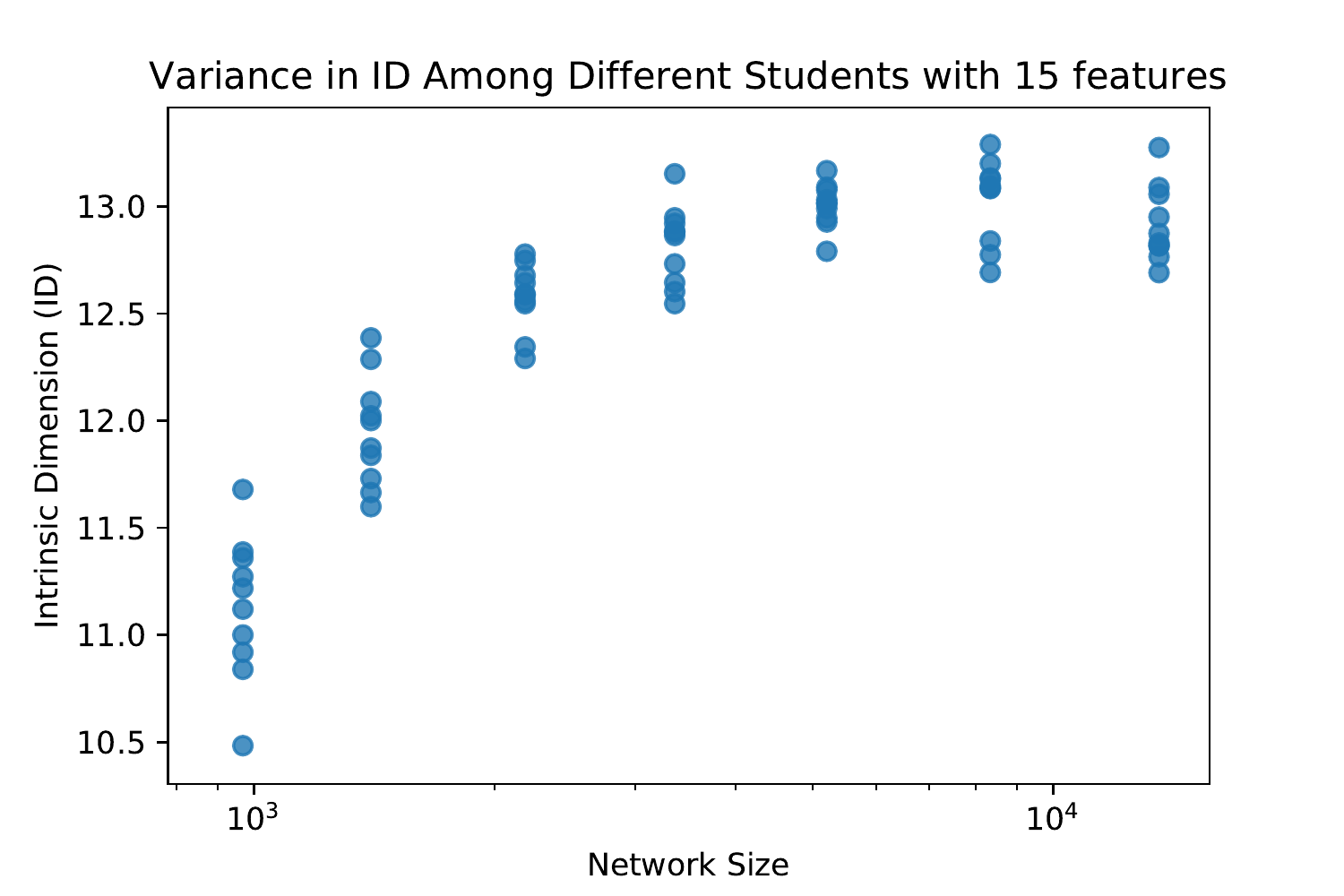} 
\caption{Variation of Intrinsic Dimension (ID) across network sizes for a single teacher. The figure on the left shows number of inputs features $ = 10$ and the one on the right has $15$. Each point on either figure is one student. All students on each figure are trained on the same teacher, but the teacher for the left and right figures are different. \label{fig:VariationAmongStudentsofID} }
\end{figure}

\begin{figure}
	\centering
	\includegraphics[scale=0.48]{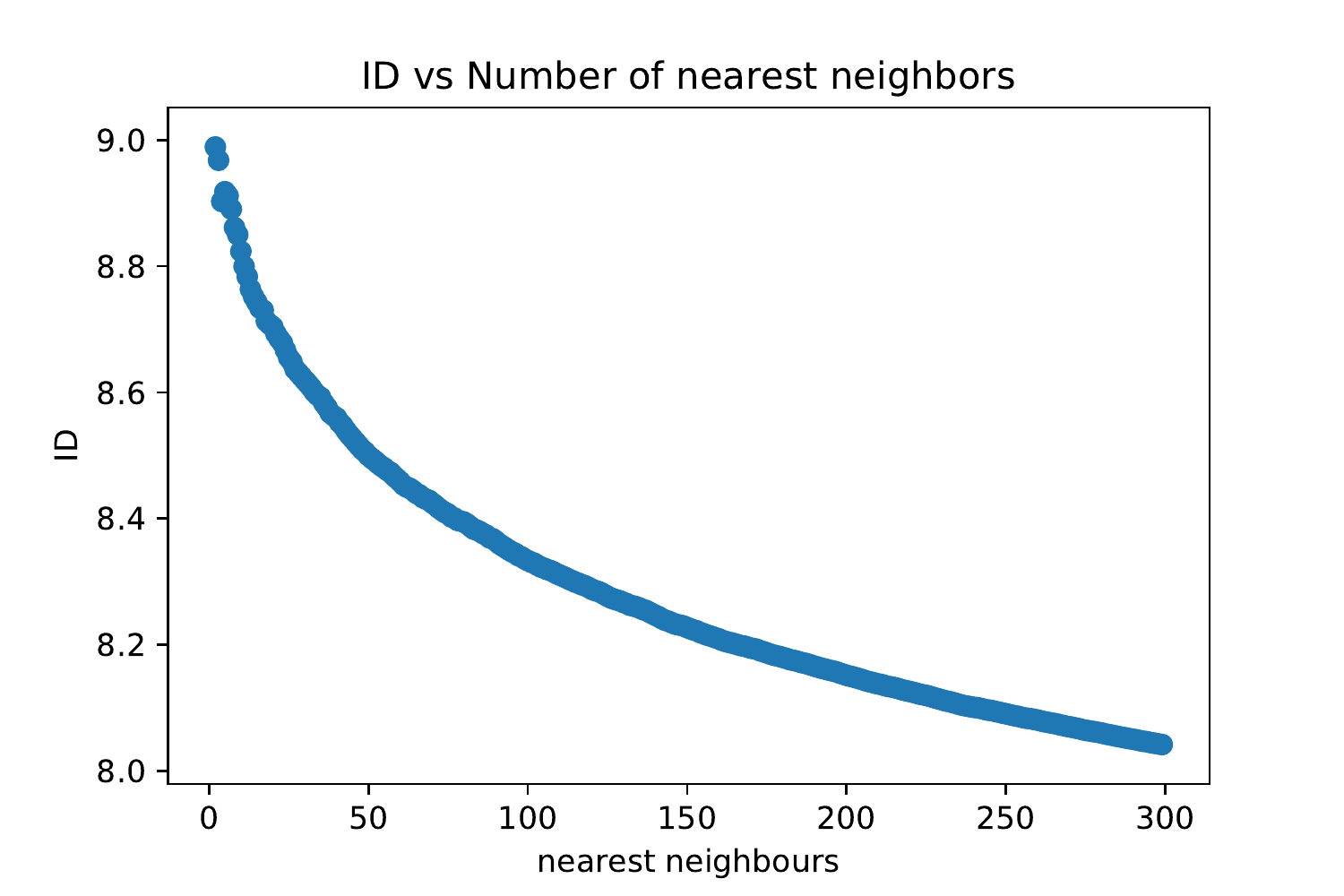} 
	\includegraphics[scale=0.48]{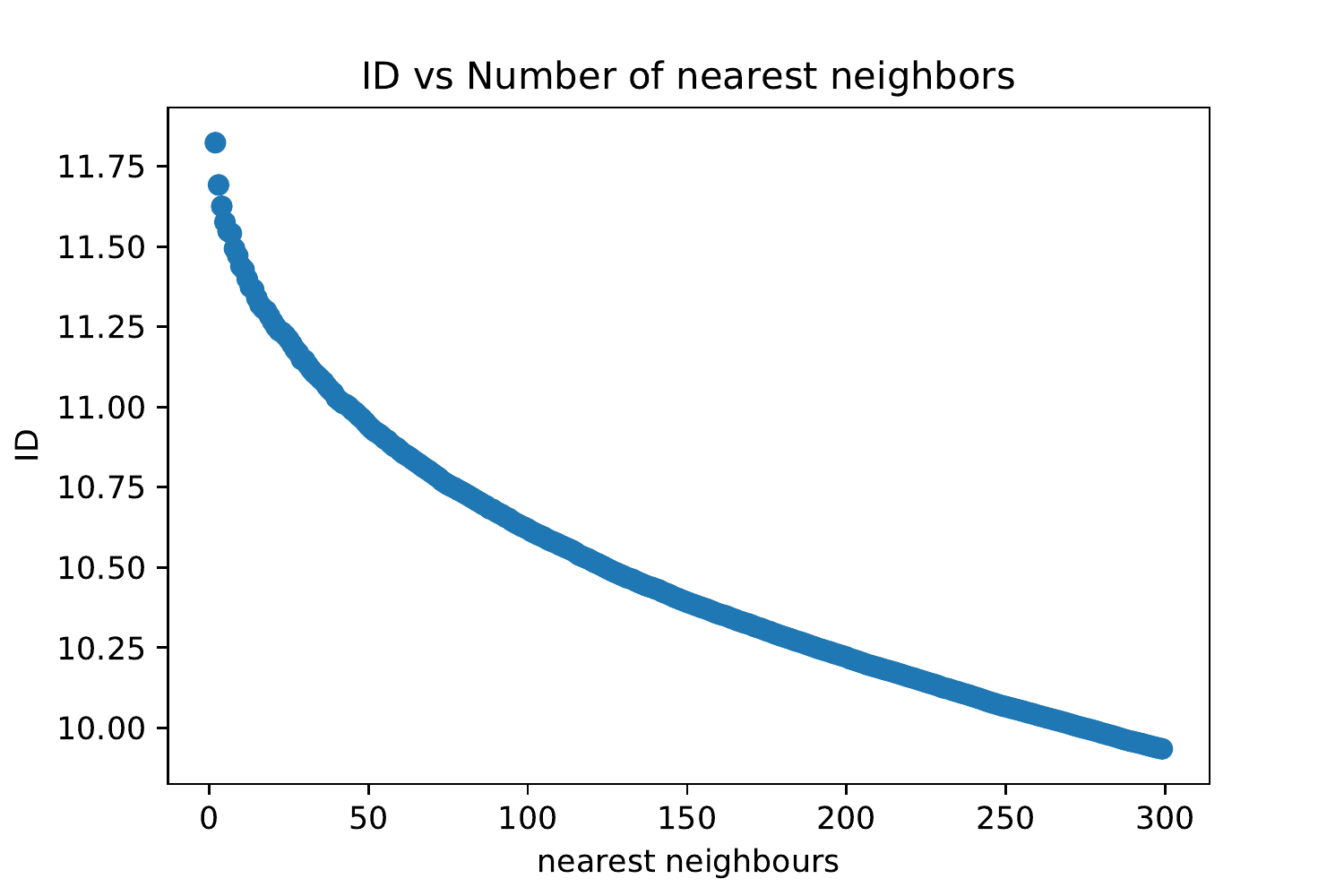} 
	\caption{Variation of Intrinsic Dimension (ID) with number of neighbors used in the algorithm. The figure on the left shows a student of size $[20,25,25,2]$ trained on a teacher with 10 features, while the one on the right has student shape   $[15,28,28,2]$ trained on teacher with $15$ features.  \label{fig:NumberNNUsedID}}
\end{figure}

In all cases we measure ID from fully trained networks, and we always use students (not teachers) in that context.  There are a large variety of potential statistical and systematic errors associated with these measurements:
\begin{itemize}
\item Variation among IDs measured from students of the same size and trained with the same teacher network (or dataset), but with different  initialization (see figure \ref{fig:VariationAmongStudentsofID}).
\item Variation of ID measurements among random groups of points sampled from the same data manifold 
\item Dependence of ID on the number of points used (and so the overall density) from the data manifold.  More points samples shorter distance scales on the manifold.  See figure \ref{fig:NumberVctUsedID}.
\item Dependence of ID on how many nearest neighbor points are used, either for NN (see figure \ref{fig:NumberNNUsedID}) or MLE type estimation.
\item Variation of ID from among points in different locations on the data data manifold (we show a histogram of results from MLE in figure \ref{fig:CIFARMLE})
\item Dataset specific distinctions, eg from the same or different classes in an image classifier, or from the same or different text sequences in a language model (discussed in section \ref{sec:Language})
\item Dependence of ID measurements on the layer studied (see figures \ref{fig:LanguageID} and \ref{fig:LayerVariationIDTS})
\end{itemize}
We  provide some brief information about many of these sources of variation in the referenced plots.  In most cases we find that the variation of the ID is small as long as it is measured with sufficiently many vectors.  It would be interesting obtain a more precise theoretical and experimental characterization of these methods in the future.

But as evidenced by the synthetic examples in figure \ref{fig:SyntheticData}, this does not lead us to believe that the IDs are fully trustworthy, especially when they are measured to be large.  Though the apparent statistical errors in ID measurements may seem small, there may be systematic errors that are more difficult to observe. 

It's conceivable that deficiencies in ID measurement actually work to the advantage of the theory relating $d$ and $4/\alpha$.  For example, $d$ tends to be underestimated when the data manifold has a boundary (or simply less support in some region), but this may also correlate with regions of the manifold where there really is less data, and these regions do not need to be modeled as precisely to achieve a good test loss.  But we leave a more thorough investigation of such subtleties to future work.

\begin{figure}
	\centering
	\includegraphics[scale=0.32]{plots/GPT_LastLyr_a_lastToken_mle_plot} 
	\includegraphics[scale=0.32]{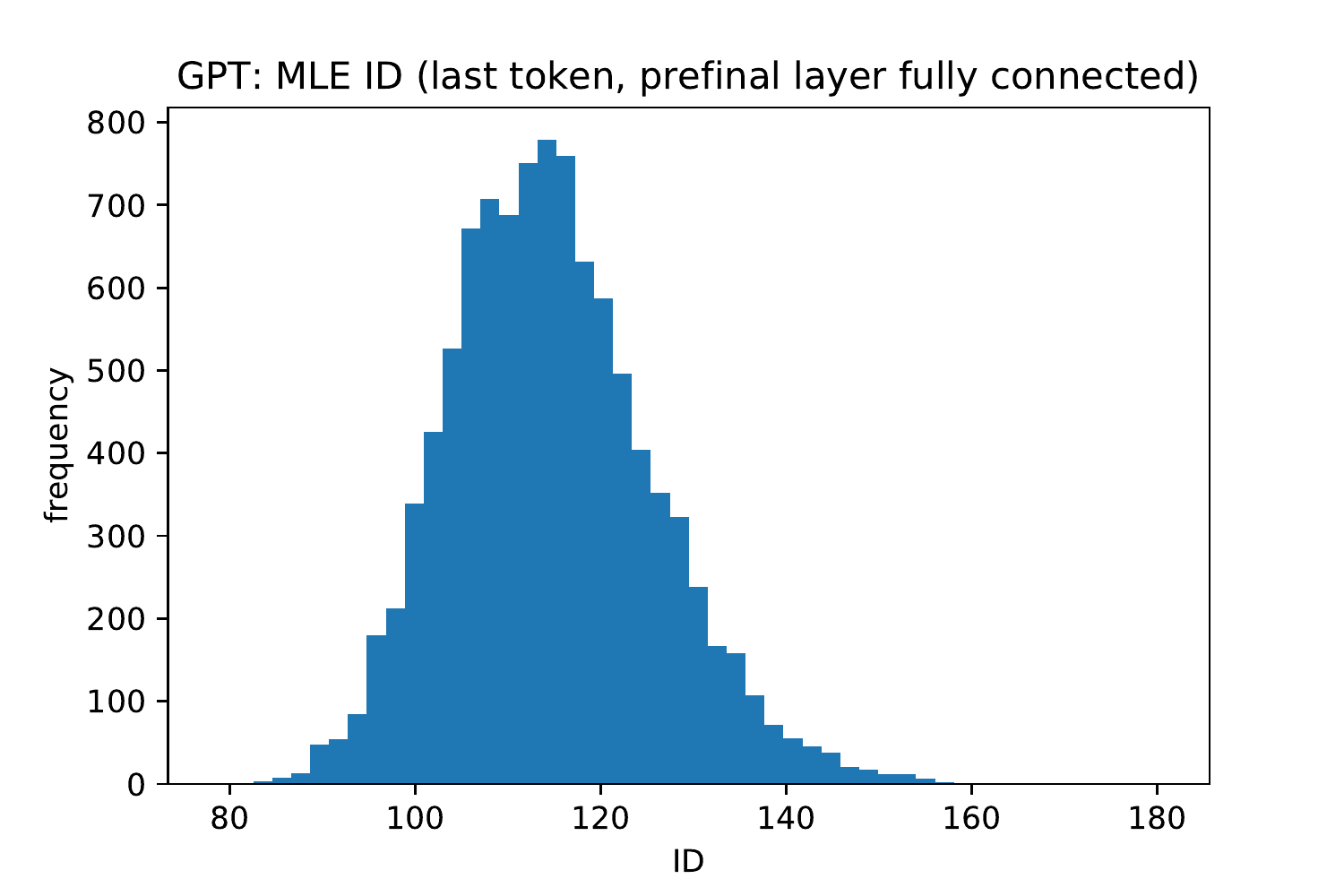} 
	\includegraphics[scale=0.32]{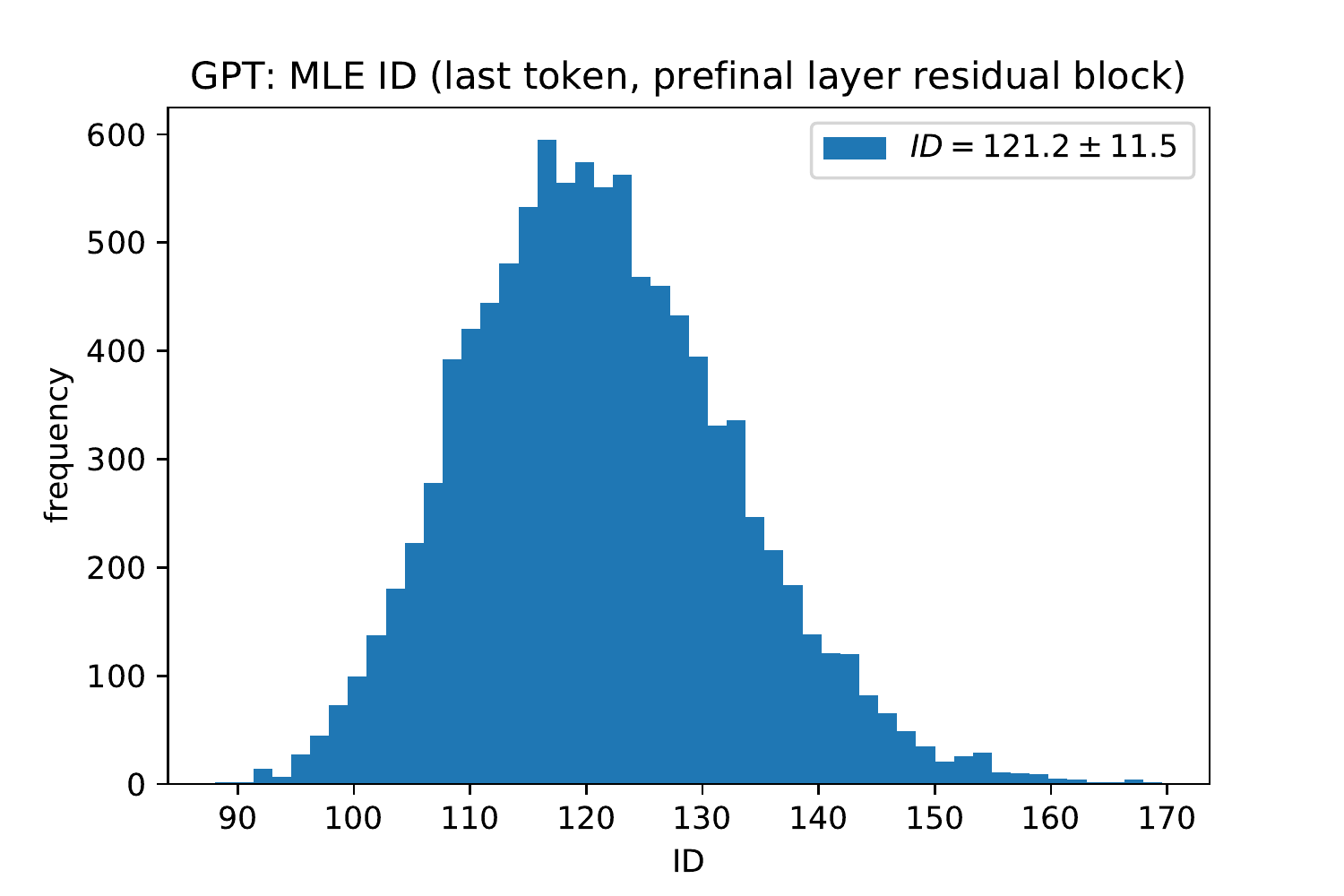} 
	\caption{These figures are histograms of the GPT MLE estimates using the last token of the prefinal layer (using $n_{neighbor}=100$).  Counts include the number of points in the data manifold that produce a given maximum-likelihood ID.  These are computed using all available text sequences, ie  test+validation (10k pts) \label{fig:GPTIDvsNumvct}}
\end{figure}

\bibliographystyle{halpha}
\bibliography{bibliography}

\end{document}